\documentclass{article}

\usepackage{microtype}
\usepackage{graphicx}
\usepackage{subcaption}
\usepackage{booktabs} 
\usepackage{makecell}
\usepackage{hyperref}

\usepackage[preprint]{icml2026}

\usepackage{amsmath}
\usepackage{amssymb}
\usepackage{mathtools}
\usepackage{amsthm}

\usepackage{xcolor}
\usepackage{soul}

\usepackage[capitalize,noabbrev]{cleveref}

\theoremstyle{plain}

\theoremstyle{definition}

\theoremstyle{remark}

\usepackage{siunitx}
\usepackage{multirow}
\usepackage[textsize=tiny]{todonotes}

\usepackage{listings}
\usepackage[T1]{fontenc}
\usepackage{tcolorbox}
\tcbuselibrary{skins,breakable}
\definecolor{LGMaroon}{RGB}{200,0,0}
\definecolor{LGGray}{RGB}{200,200,200}
\newtcolorbox{analysisbox}[1][]{
    enhanced jigsaw,
    colback=white,
    colframe=LGMaroon!75!black,
    fonttitle=\bfseries,
    boxsep=5pt,
    left=5pt,
    right=5pt,
    top=5pt,
    bottom=5pt,
    title=#1
}

\usepackage{longtable}
\usepackage[table]{xcolor}
\usepackage{arydshln}

\definecolor{proprietary}{HTML}{E8FFE8} 

\usepackage{pifont}
\newcommand{\cmark}{\color{green}{\ding{51}}}%
\newcommand{\xmark}{\color{red}{\ding{55}}}%
\newcommand{\up}[1]{\textcolor{green!60!black}{$\uparrow$\,#1}}
\newcommand{\down}[1]{\textcolor{red!70!black}{$\downarrow$\,#1}}
\newcommand{\minus}{\scalebox{0.75}[1.0]{$-$}}

\usepackage{tcolorbox}

\icmltitlerunning{PersonalHomeBench: Evaluating Agents in Personalized Smart Homes}

\begin{document}

\twocolumn[
  \icmltitle{PersonalHomeBench: Evaluating Agents in Personalized Smart Homes}

  \icmlsetsymbol{equal}{*}

  \begin{icmlauthorlist}
    \icmlauthor{Manasa Bharadwaj}{exlg}
    \icmlauthor{Yolanda Liu}{equal,exlg}
    \icmlauthor{InJung Yang}{equal,lg}
    \icmlauthor{Sungil Kim}{lg}
    \icmlauthor{Nikhil Verma}{exlg}
    \icmlauthor{Ko Keun Kim}{lg}
    \icmlauthor{Kevin Ferreira}{exlg}
    \icmlauthor{Youngjoon Kim}{lg}
  \end{icmlauthorlist}

  \icmlaffiliation{lg}{LG Electronics, Korea}
  \icmlaffiliation{exlg}{Work done while at LG Toronto AI Lab.}

  \icmlcorrespondingauthor{InJung Yang}{kayla.yang@lge.com}

  \icmlkeywords{Machine Learning, ICML}

  \vskip 0.3in
]

\printAffiliationsAndNotice{}  

\begin{abstract}

Agentic AI systems are rapidly advancing toward real-world applications, yet their readiness in complex and personalized environments remains insufficiently characterized. 
To address this gap, we introduce \textbf{PersonalHomeBench}, a benchmark for evaluating foundation models as agentic assistants in personalized smart home environments. 
The benchmark is constructed through an iterative process that progressively builds rich household states, which are then used to generate personalized, context-dependent tasks. 
To support realistic agent–environment interaction, we provide \textbf{PersonalHomeTools}, a comprehensive toolbox enabling household information retrieval, appliance control, and situational understanding.
PersonalHomeBench evaluates both reactive and proactive agentic abilities under unimodal and multimodal observations.
Thorough experimentation reveals a systematic performance reduction as task complexity increases, with pronounced failures in counterfactual reasoning and under partial observability, where effective tool-based information gathering is required. 
These results position PersonalHomeBench as a rigorous evaluation platform for analyzing the robustness and limitations of personalized agentic reasoning and planning.
\end{abstract}

\begin{figure}[ht]
  \begin{center}
    \centerline{\includegraphics[width=\linewidth]{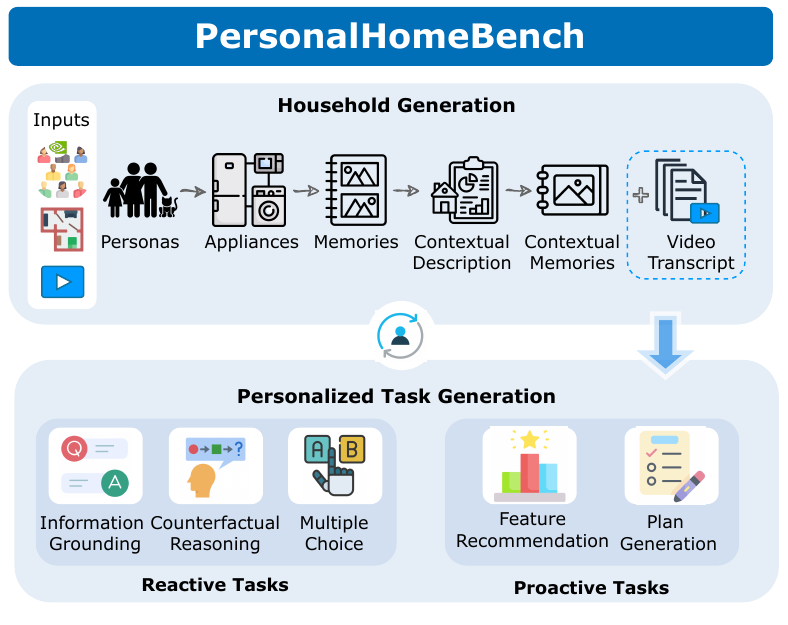}}
    \caption{
    \textbf{Overview of the PersonalHomeBench data generation pipeline.}
      Detailed household environments are first assembled from personas, devices, memories, and contextual descriptions, with optional video grounding. Five categories of personalized tasks are then generated from these environments, spanning reactive question answering and proactive assistance, enabling evaluation of agent behavior in personalized smart home settings.
    }
    \label{fig:vertical_pipeline}
  \end{center}
\end{figure}

\section{Introduction}
\label{sec:intro}
Agentic AI systems aim to move beyond passive prediction toward autonomous assistance in everyday environments, where robust performance depends on contextual reasoning, uncertainty handling, and personalization. 
Although a growing body of benchmarks have been proposed to evaluate agentic systems, most existing evaluations focus on generic tasks in simplified or simulated settings \cite{chang2024agentboard,bonatti2024windows,xi2025agentgym}. 
As a result, they offer limited insight into model behavior under evolving individualized preferences, long term context, and partial observability, yielding an incomplete assessment of readiness for personalized real world deployment.

Smart home environments provide a natural and challenging testbed for exposing these limitations. 
Effective operation in this domain requires reasoning over heterogeneous devices, maintaining persistent and temporally grounded context, and adapting to evolving user preferences while accounting for safety, efficiency, and operational constraints. 
Prior smart home benchmarks have largely relied on simplified abstractions, such as text only environments with limited device coverage and primitive actions, or narrowly scoped tasks focused on isolated device control \cite{li2025homebench, bartkowiak2025edgewisepersona}. 
Viewing smart homes as a realistic setting for personalized agentic intelligence, we introduce \textbf{PersonalHomeBench}, a framework for evaluating foundation models in household environments that demand contextual reasoning, personalization, and proactive decision making beyond reactive task execution.

\textit{PersonalHomeBench} adopts a staged environment construction process that reflects real household evolution, sequentially instantiating resident personas, assigning device ecosystems, and accumulating contextual signals over time to synthesize coherent household worlds, with optional multimodal grounding. 
As illustrated in Fig.~\ref{fig:vertical_pipeline}, five categories of personalized tasks are derived from these environments to support unified evaluation of reactive reasoning and proactive planning, using a procedure inspired by \cite{singh-etal-2024-personal}.
These include three reactive question answering tasks, \textsc{Information Grounding}, \textsc{Counterfactual Reasoning}, and \textsc{Multiple Choice}, and two proactive tasks, \textsc{Feature Recommendation} and \textsc{Plan Generation}, which target appliance- and routine-level orchestration for timely, preference-aware home assistance.

To support faithful evaluation of agentic behavior, we introduce \textbf{PersonalHomeTools}, a smart home toolbox that provides a grounded interaction layer between agents and realistic household environments. 
It exposes structured household information retrieval, persistent memory and event access, environmental perception, and realistic appliance control through domain-specific APIs. 
Together, these capabilities enable systematic study of agents that unify perception, memory, and tool use for personalized decision making.

We evaluate a wide spectrum of foundation models, spanning compact four billion parameter models suitable for privacy-preserving edge deployment to cloud-based large-scale models. 
Beyond task-specific metrics, we employ a Role-Playing LLM \cite{wang2024rolellm} as a Judge to estimate projected satisfaction for individual household members. 
Across model scales, performance degrades sharply on tasks involving tool use, long-horizon planning, and advanced reasoning, often accompanied by increased hallucination. 
These findings reveal fundamental limitations of current systems and highlight the difficulty of achieving robust, personalized, and proactive behavior in realistic smart home settings, establishing PersonalHomeBench as a practical resource for diagnosing failures and guiding the development of more reliable personalized agents.

Our contributions are threefold:
\begin{itemize}
\itemsep0em
\item We introduce \textit{PersonalHomeBench}\footnote{\href{https://huggingface.co/datasets/PHBM-lab/PHBM-dataset}{Dataset}}, a benchmark for studying personalized and proactive agent behavior in contextually grounded smart home environments.
\item We present \textit{PersonalHomeTools}, a comprehensive smart home toolkit for realistic agent evaluation.
\item We conduct systematic evaluation across model scales and reasoning levels, and leverage a Role Playing LLM-as-a-Judge for assessing personalized satisfaction.
\end{itemize}

\section{Related Works}
\label{sec:related_works}
\begin{table}[ht]
\caption{Comparison of benchmarks along five dimensions: \textsc{home type}, personalization (\textsc{PERS.}), proactiveness (\textsc{PROACT.}), agentic interaction with tools (\textsc{AGENTIC}), and multimodal support (\textsc{MM}). SIM = simulated homes, REAL = real-home data, and NONE = non–home-based settings. * marks settings where multimodal content is represented as text for inference.
}
\label{tab:related_works_compact}
\begin{sc}
\centering
\resizebox{\linewidth}{!}{%
\begin{tabular}{lccccc}
\toprule
Method & Home-Type & Pers. & Proact. & Agentic & MM \\
\midrule
Embodied Benchmarks & SIM & \xmark & \xmark & \cmark & \cmark \\
Toyota Smarthome & REAL & \xmark & \xmark & \xmark & \cmark \\
PersonalTravelPlanner & NONE & \cmark & \xmark & \cmark & \xmark \\
HomeBench & SIM & \xmark & \xmark & \cmark & \xmark \\
EdgeWisePersona & SIM & \cmark & \cmark & \cmark & \xmark \\
ContextAgent & NONE & \cmark & \cmark & \cmark & \cmark* \\
\midrule
\textbf{PersonalHomeBench (Ours)} & SIM+REAL & \cmark & \cmark & \cmark & \cmark \\
\bottomrule
\end{tabular}
}
\end{sc}
\end{table}
As summarized in Table \ref{tab:related_works_compact}, embodied household benchmarks \cite{Puig2018VirtualHomeSH,Shridhar2019ALFREDAB,li2024behavior1k,chang2024partnr, yang2025embodiedbench} focus on navigation and atomic action execution in simulated 3D environments, but do not model real smart-device ecosystems or personalization. 
Toyota Smarthome \cite{Das_2019_ICCV} offers real-world home video data, but is limited to narrow demographics and fixed environments, and focuses on activity recognition rather than personalized, proactive, or interactive agentic assistance.

Recent benchmarks incorporate aspects of tool usage and personalization. PersonalTravelPlanner \cite{singh-etal-2024-personal} studies personalized planning with simple tools outside the home domain. HomeBench \cite{li2025homebench} evaluates reactive smart-home command validation with limited appliance coverage and restricted operations, while EdgeWisePersona \cite{bartkowiak2025edgewisepersona} focuses on profile reconstruction and lightweight routine inference over a small device set, with only minimal personalization via short persona descriptions. ContextAgentBench \cite{yang2025contextagent} introduces proactive agents with contextual information and tools, but relies on ego-centric traces, generic APIs, and language-only inference with limited personalization. 

\textit{PersonalHomeBench} focuses on agent behavior in realistic home settings where personalization, long-term context, perception, and smart-device interaction jointly influence decisions. By integrating with \textit{PersonaHomeTools}, it evaluates whether agents can retrieve relevant household information, reason over persistent state, and coordinate appliance actions to provide proactive, personalized assistance. As summarized in Table~\ref{sec:related_works}, no existing benchmark simultaneously covers real home data, personalization, proactiveness, domain-specific tools, and multimodality, making \textit{PersonalHomeBench} a unified setting for studying personalized agentic behavior in smart home domain.
%
Additional discussion of related benchmarks is provided in Appendix~\ref{app:related_works}.
\section{PersonalHomeBench}
\label{sec:personal_home_bench}
An overview of the PersonalHomeBench generation pipeline is shown in Figure \ref{fig:vertical_pipeline}. 
The individual components are detailed in the subsequent sections.

\begin{figure*}[ht]
  \begin{center}
    \centerline{\includegraphics[width=\linewidth]{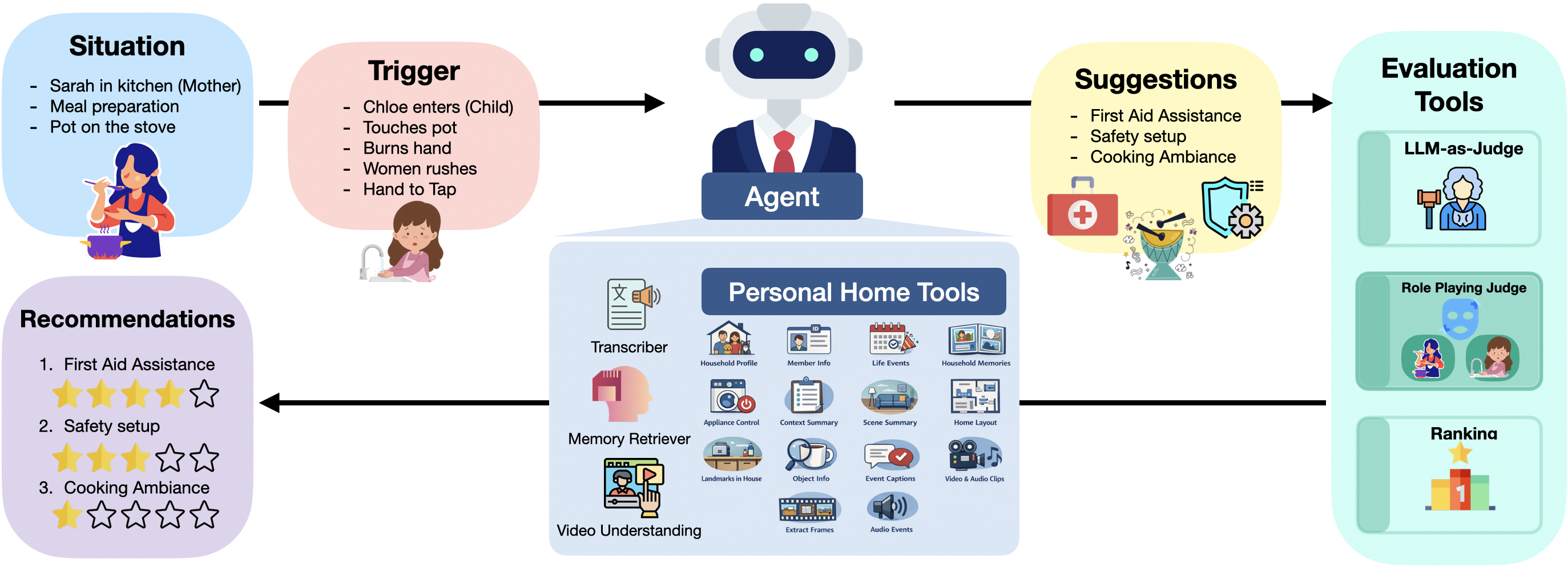}}
    \caption{
An agent interprets real-world household situations and triggers, using PersonalHomeTools to generate proactive plans (e.g., safety, first aid, ambience control), which are evaluated by the Role Playing Judge to produce personalized recommendations.
    }
    \label{fig:inference_flow}
  \end{center}
\end{figure*}

\subsection{Household Generation}
\label{sec:hh_gen}

As illustrated in the \emph{Household Generation} module of Fig.~\ref{fig:vertical_pipeline}, generation begins from a minimal scaffold specifying household type, layout, and coarse demographics, and is seeded with personas from Nemotron Personas \cite{nvidia-Nemotron-Personas-USA} to instantiate realistic occupants. These personas, together with household metadata, are used by a data generation model to produce complete occupant profiles that conform to a predefined schema (Appendix~\ref{app:schemas}).
Household context is then progressively enriched by assigning appliances appropriate to the environment, synthesizing historical memories that capture both device usage and personal events, and updating occupant attributes and appliance states to reflect the current situation. Finally, a compact cache of situation-relevant memories is constructed, providing a coherent snapshot of the household that serves as the foundation for personalized task generation.

For the multimodal setting, we further ground household context using exocentric videos captured by cameras placed across rooms. 
We produce structured video transcripts containing entity references, object mentions, time-stamped events, and aligned audio captions. 
To preserve fidelity to visual evidence while supporting reasoning beyond direct observation, we maintain separate representations for video-grounded and inferred information.
The detailed subject instructions and the video transcript schema are provided in Appendices \ref{app:video_instructions} and \ref{app:schemas}, respectively.

All generated data undergo manual review to ensure realism, internal consistency, and schema compliance, yielding high-quality household environments used for downstream personalized task generation.

\subsection{Personalized Task Generation}
\label{sec:task_gen}
Using the constructed household environments and associated videos and transcripts when available, we derive five categories of personalized tasks that probe complementary aspects of reasoning and planning in smart home settings. 
Task generation assumes full observability, providing models with the entire household context.

\noindent\textbf{Reactive} tasks are framed as closed-form question answering and include \textsc{Information Grounding (IG)}, \textsc{Counterfactual Reasoning (CF)}, and \textsc{Multiple Choice (MC)}. 
These tasks assess the ability of an agent to extract and integrate household information across varying levels of difficulty, ranging from direct factual lookup to multi-hop reasoning over numerous entities and attributes. 
\textsc{IG} measures generative factual grounding, while \textsc{MC} evaluates the same capability under constrained answer choices. 
\textsc{CF} extends \textsc{IG} by introducing paired counterfactual variants, one that alters the ground-truth answer and one that leaves it unchanged, enabling targeted analysis of robustness to hypothetical perturbations. 
For instance, modifying temporal context may change a resident’s age, whereas altering an unrelated attribute such as attire should not.

\noindent\textbf{Proactive} tasks evaluate open-ended, subjective decision making for personalized assistance, testing whether agents can anticipate user needs and initiate appropriate actions without explicit prompts. 
We define two such tasks, \textsc{Feature Recommendation (FR)} and \textsc{Plan Generation (PG)}, centered on appliance and routine orchestration.

In \textsc{FR}, the agent ranks appliance features by contextual relevance using provided API-level inventories, reflecting real-world feature discovery and configuration.
In \textsc{PG} setting, agent generates structured multi-step plans that coordinate heterogeneous devices toward a shared goal.
For example, safety-critical situations can prompt a coordinated response that initiates first-aid assistance, notifies caregivers, and shuts down hazardous appliances (Fig.~\ref{fig:inference_flow}).

\subsection{Dataset Statistics}
\begin{figure*}[t]
    \centering
    \begin{subfigure}{0.24\linewidth}
        \centering
        \includegraphics[width=\linewidth]{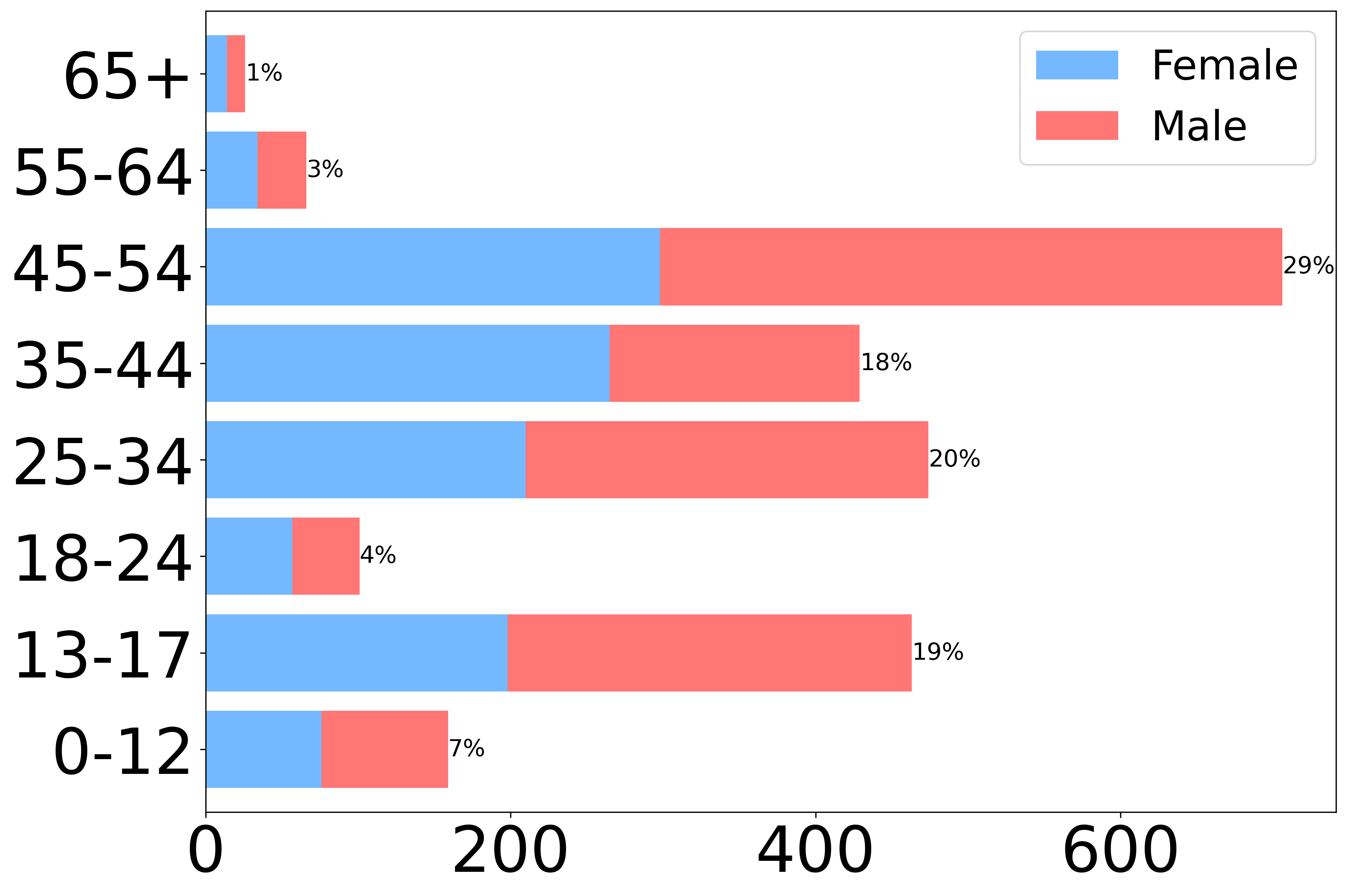}
        \caption{Demographics}
        \label{fig:age_gender}
    \end{subfigure}
    \hfill
    \begin{subfigure}{0.24\linewidth}
        \centering
        \includegraphics[width=\linewidth]{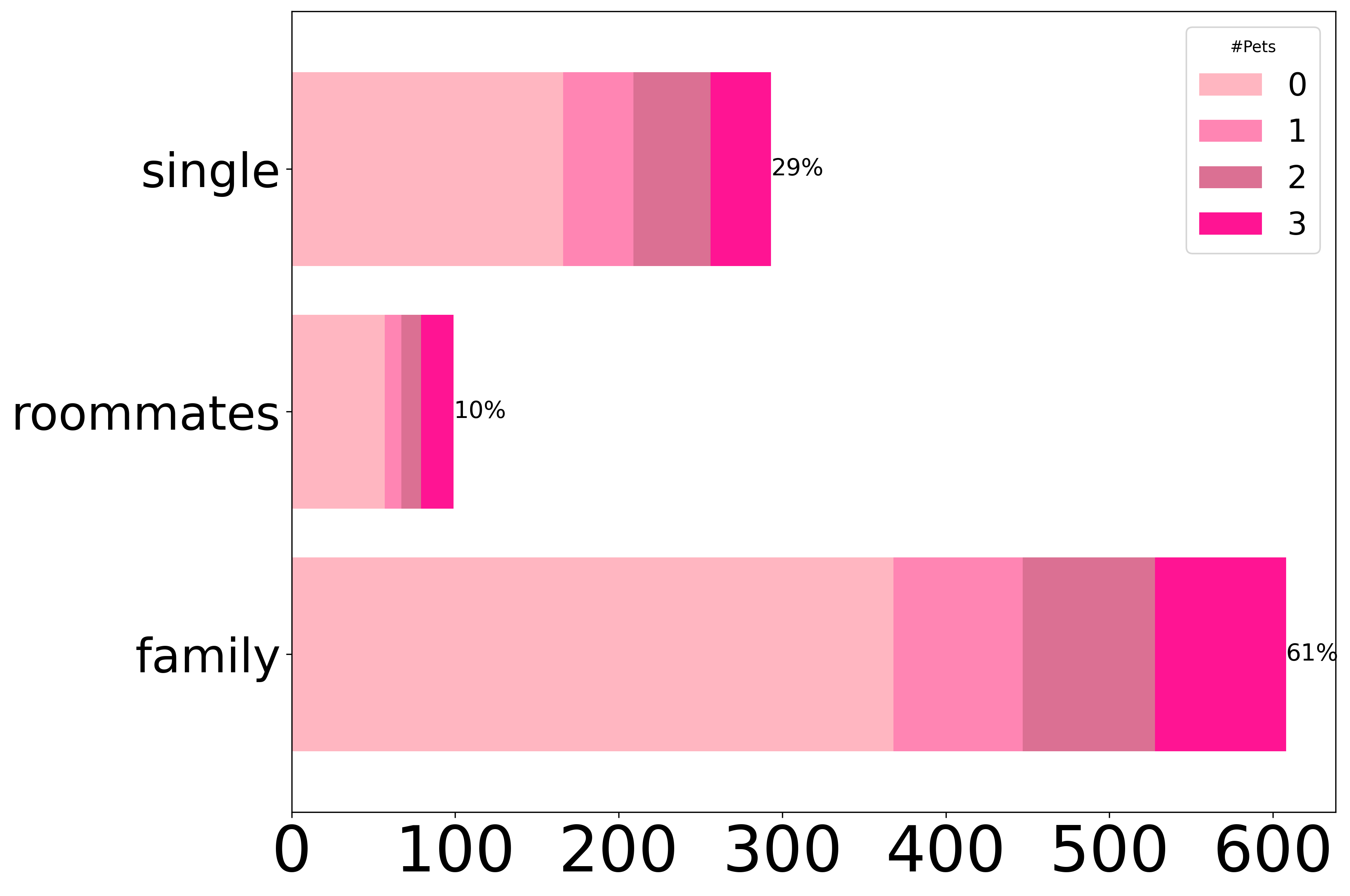}
        \caption{Pets}
        \label{fig:pets_dist}
    \end{subfigure}
    \hfill
    \begin{subfigure}{0.24\linewidth}
        \centering
        \includegraphics[width=\linewidth]{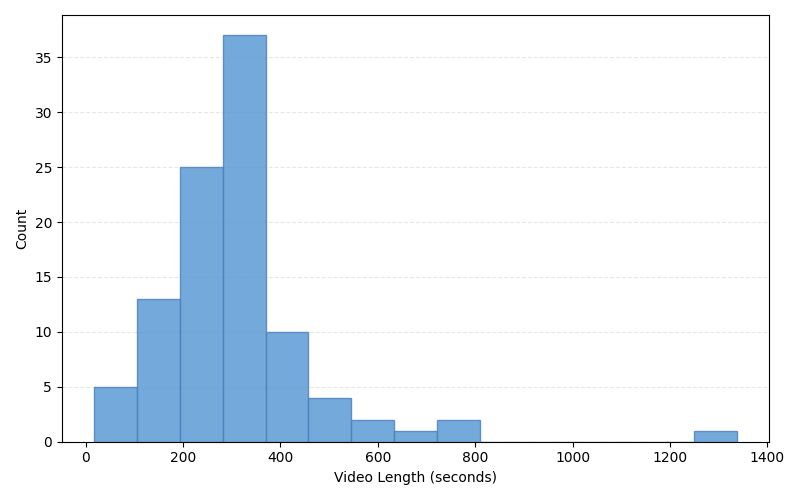}
        \caption{Video Lengths}
        \label{fig:video_lens_dist}
    \end{subfigure}
    \hfill
    \begin{subfigure}{0.24\linewidth}
        \centering
        \includegraphics[width=\linewidth]{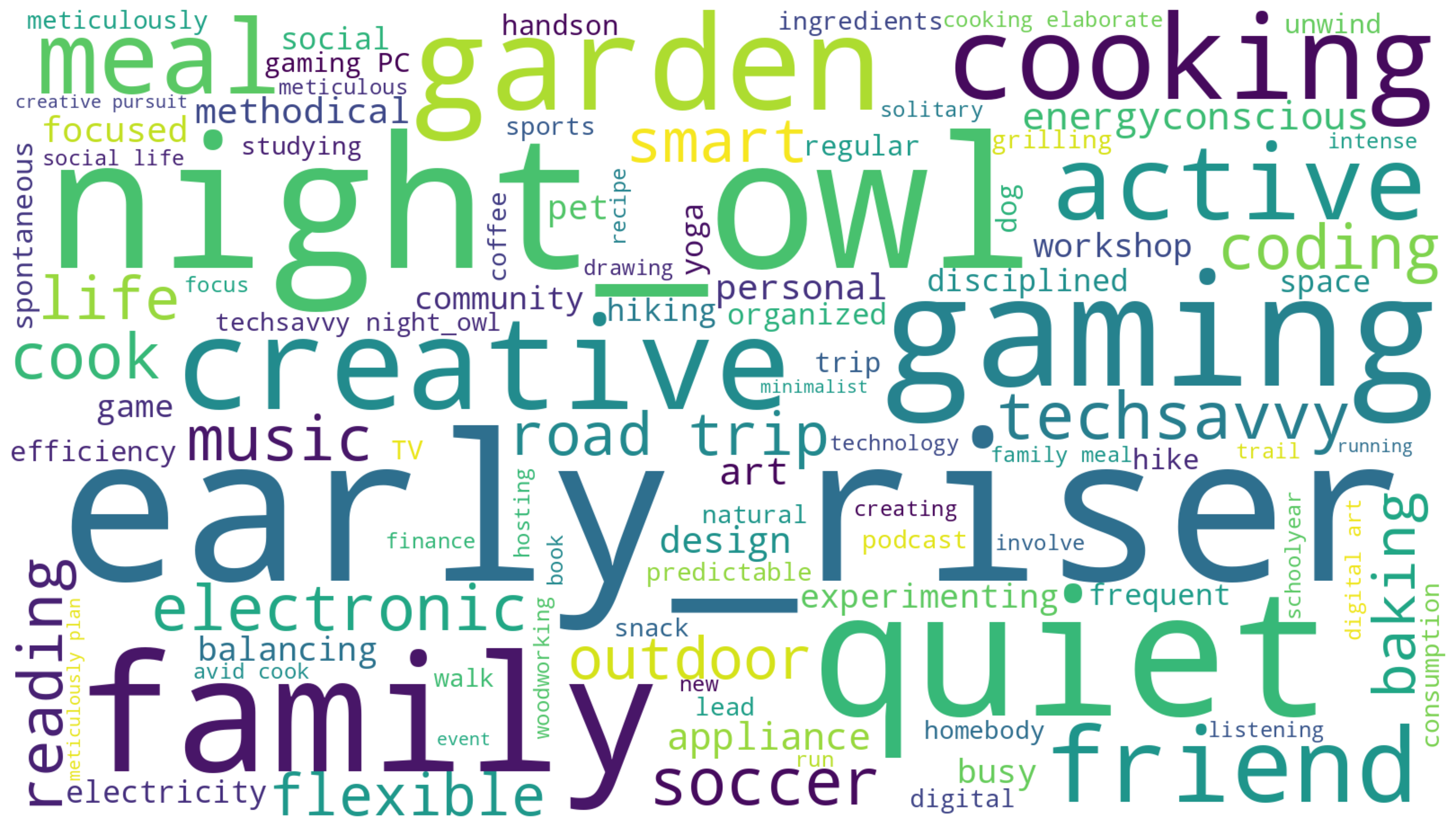}
        \caption{Attributes of Personas}
        \label{fig:persona_cloud}
    \end{subfigure}
    \caption{Statistics of generated household and persona diversity in PersonalHomeBench. (a) Age and gender distribution of household members. (b) Distribution of the number of pets per household type. (c) Distribution of video lengths (sec). (d) Word cloud of common persona attributes, illustrating diversity in interests, habits, and lifestyle characteristics.}
    \label{fig:text_data_hh_stats}
\end{figure*}

\textbf{Household Data.}
\textit{PersonalHomeBench} includes 1,100 households generated using the procedure in Sec.~\ref{sec:hh_gen}, each adhering to the schema in Appendix~\ref{app:schemas}, and comprising over 2,000 unique household member profiles. 
The households exhibit broad demographic and structural diversity, with balanced age and gender distributions (Fig.~\ref{fig:age_gender}).
Most households do not own pets, while pet ownership is most popular among families (Fig.~\ref{fig:pets_dist}).
Every household contains at least five mandatory appliances, and the majority include 9–10 appliances (Appendix Fig.~\ref{fig:hh_by_appliance}). 
Household members are further annotated with rich persona attributes, including hobbies, lifestyles, and preferences, supporting fine-grained personalization; frequently occurring attributes are summarized in Fig.~\ref{fig:persona_cloud}. 
Furthermore, we ground 100 households in real-home videos spanning \textit{Health}, \textit{Safety}, and \textit{Daily Care} scenarios, with the video length distribution shown in Fig.~\ref{fig:video_lens_dist}. 
Additional analyses appear in Appendix~\ref{app:additional_hh_stats}.

\textbf{Personalized Task Data.}
\textit{PersonalHomeBench} contains a total of \textbf{9,168} task instances, comprising 6,000 text-only examples and 3,168 multimodal examples. 
The text-only split contains 1,000 instances each for \textsc{IG}, \textsc{MC}, \textsc{FR}, and \textsc{PG}, and 2,000 instances for \textsc{CF}.
For multimodal grounding, the dataset contains 742 instances each of \textsc{IG} and \textsc{MC}, 1,484 \textsc{CF} instances, and 100 instances each of \textsc{FR} and \textsc{PG}.
Reactive tasks (\textsc{IG}, \textsc{CF}, \textsc{MC}) are further annotated with three difficulty levels (easy, medium, hard). 
Compared to existing benchmarks on personalized and proactive agentic systems \cite{xie2024travelplanner,singh-etal-2024-personal,yang2025contextagent,bartkowiak2025edgewisepersona}, PersonalHomeBench offers broader coverage in scale, task diversity, and supported input modalities, enabling a more comprehensive evaluation of agentic behavior in realistic settings.

\section{PersonalHomeTools}
\label{sec:toolbox}
As depicted in Fig. \ref{fig:inference_flow}, PersonalHomeTools serves as a unified interaction layer that enables agents to perceive, reason about, and act within household environments, constituting a core component of our framework.
The toolbox provides capabilities for household information retrieval, appliance control, and contextual understanding.

\textbf{Household information retrieval} provides access to relatively static metadata about occupants, pets, appliances, and other household entities.
Since this information changes infrequently, these functions are implemented as non-agentic tools that return deterministic structured outputs.

\textbf{Appliance control} tools enable agents to influence the state of smart devices within a household. 
We construct a comprehensive control sandbox by implementing feature-level interfaces for more than 40 smart appliances, derived from public documentation of major smart home providers\footnote{\href{https://www.lg.com/}{LG
 Electronics} and \href{https://www.samsung.com/}{Samsung}.}.
Appliance interactions may be deterministic, such as power control, or mediated by appliance-specific agents, such as a personalized recipe recommender for a refrigerator.

The majority of agentic modules are designed to support deep \textbf{contextual understanding} by enabling hybrid reasoning via natural-language queries with optional attribute-based filtering, while reusing cached results until the underlying state changes.
Core agentic components consist of three specialized agents: a \textit{Transcriber}, a \textit{Memory Retriever}, and a \textit{Video Understanding Agent}.
The Transcriber incrementally constructs and maintains the household state by extracting and updating memories, events, and contextual attributes from observations.
The Memory Retriever can surface situation-aware memories tied to the overall household and to specific occupants as needed.
The \textit{Video Understanding Agent} retrieves relevant clips, frames, or audio cues when multimodal grounding is available, enabling visual and auditory context to be incorporated into inference.

Together, these components provide a coherent interface through which agents gather context, reason over persistent memory, and execute personalized actions, enabling rigorous evaluation of tool-mediated reasoning and proactive assistance.
A complete list of supported appliances, features, tools, and agent interfaces is provided in Appendix \ref{app:toolbox}.


\section{Experimental Setup}
\label{sec:experiments}
This section presents the evaluation protocol for assessing foundation models on \textit{PersonalHomeBench}, encompassing the evaluated models, experimental settings, and metrics.

\subsection{Models and Hyperparameters}

\textbf{Models.}
We benchmark eleven foundation models spanning a wide range of size tiers, architectures, and reasoning configurations from the \textsc{Qwen3} \cite{yang2025qwen3}, \textsc{GPT}, \textsc{Nemotron}, and \textsc{Gemini} model families. The evaluated models are grouped into three categories:

(1) \emph{Large proprietary models}: \textsc{GPT-4} \cite{achiam2023gpt}, \textsc{GPT-4o} \cite{hurst2024gpt}, and \textsc{Gemini 2.5 Pro} \cite{comanici2025gemini}, 
(2) \emph{Medium-size open models}: \textsc{Qwen3-30B-Instruct}, \textsc{Qwen3-30B-Thinking}, \textsc{Nemotron3-Nano-30B-A3B} \cite{blakeman2025nemotron}, and \textsc{GPT-OSS-20B} \cite{agarwal2025gpt}.
(3) \emph{Small-size open models}: \textsc{Qwen3-4B} and \textsc{Qwen3-4B-Thinking}.

For models that support explicit reasoning modes, we evaluate multiple reasoning configurations. 
In particular, \textsc{GPT-OSS-20B} is tested at three reasoning levels (low, medium, high). 
For clarity, we append a single-letter tag to model names to denote reasoning mode, e.g., \textsc{Qwen3-4B [T]}, \textsc{Qwen3-30B [T]}, and \textsc{GPT-OSS-20B [L/M/H]}. 

We use \textsc{Gemini 2.5 Pro} as the primary model for dataset generation and as the evaluation backbone for all LLM-as-a-Judge protocols, including semantic exact matching and the Role Playing Judge ($RPJ$). 
To reduce potential evaluation bias, we report the performance of \textsc{Gemini 2.5 Pro} separately from that of the other evaluated models.

\textbf{Hyperparameters.}
We use temperature $0.6$ for dataset generation to encourage diversity, and temperature $0.0$ with top-$p=1.0$ during evaluation for deterministic and reproducible outputs. 
Each task is allotted up to 15 turns to finish.


In addition to the agentic inference, we evaluate models in a \textit{Sole-Reasoning} setting, where the full household context is directly provided to the model without requiring any tool invocation.
This configuration serves as an upper bound as relevant information is explicitly available in the prompt.


A complete description of all model configurations and inference settings is provided in Appendix \ref{app:models}.

\subsection{Evaluation Metrics}
\label{sec:eval_metrics}
This subsection describes the evaluation metrics used across tasks, followed by a detailed description of the \textbf{Role Playing Judge ($RPJ$)} framework used for subjective assessment.

\noindent\textbf{Reactive Tasks.}
For these deterministic question answering tasks, we report Accuracy (\textsc{ACC}) (Appendix Eq. \ref{eq:ig_cf_correct} \& \ref{eq:mc_acc}).


For \textsc{IG} and \textsc{CF}, correctness is first assessed using string-level exact match. When an exact match fails, we employ \textsc{Gemini 2.5 Pro} as a semantic equivalence judge to determine whether the predicted answer is meaningfully consistent with the ground truth. This hybrid approach is necessary because valid responses may vary in surface form, ranging from short phrases to full sentences, while expressing the same underlying meaning.
For \textsc{CF}, Table~\ref{tab:main_results_table} reports the macro-average over \textsc{$CF_{unchanged}$} and \textsc{$CF_{changed}$}.

For \textsc{MC}, accuracy is computed as an exact match between the selected option (or its text) and the ground truth choice. Since all valid answers are explicitly enumerated, we do not apply semantic judging for this task.

\noindent\textbf{Proactive Tasks.}
The \textsc{Feature Recommendation} task is formulated as a ranking problem and evaluated using Mean Average Precision at rank one (\textsc{MAP@1}) \cite{zhu2004map}. 
For \textsc{Plan Generation}, we report both Role Playing Judge score to assess preference-aligned plan quality and a \emph{validity} metric, defined as the proportion of sub-program steps that are executable with correct tool or feature names and well-formed arguments (Appendix Eq. \ref{eq:plan_validity}).

\noindent\textbf{Role Playing Judge.}
Evaluating \textsc{PG} requires capturing subjective satisfaction under personalized household preferences. We therefore leverage a Role Playing LLM-as-a-Judge framework, \textbf{$RPJ$}, in which the judge model evaluates plans from the perspectives of individual household members and the household collectively.
As formalized in Algorithm \ref{alg:role_playing_judge}, $RPJ$ operates in two modes: \textsc{abs} and \textsc{rel}.

In \textsc{abs} mode, $RPJ$ assigns a scalar score and textual feedback to each sub-program within a plan. 
Since plans may include both strong and weak sub-programs, this mode enables fine-grained analysis and supports surfacing only high-quality components. 
We employ $RPJ_{ABS}$ in multimodal evaluation and report overall average scores.

In \textsc{rel} mode, $RPJ$ ranks complete plans across models according to household preferences, framing evaluation as a preference comparison that aligns naturally with RLHF-style training objectives \cite{comanici2025gemini}.
$RPJ_{\textsc{REL}}$ is particularly efficient when comparing many models. For example, evaluating eleven models requires $11x$ calls under $RPJ_{\textsc{ABS}}$ but only $x$ calls under $RPJ_{\textsc{REL}}$.

For Table \ref{tab:main_results_table} $RPJ_{\textsc{REL}}$ ranks the 11 candidate plans per instance and we report win rate, i.e., the fraction of times a model is ranked first; since win rates sum to 100\%\footnote{For top-$k$ rankings with $k>1$, win rates do not sum to 100\%.}, this metric reflects a distribution of preference mass rather than absolute quality.
Details on metric definitions and mathematical formulations are provided in the Appendix \ref{app:additional_metrics_definitions}.

\begin{small}    
\begin{algorithm}[htb]
\caption{Role-Playing Judge ($RPJ$)}
\label{alg:role_playing_judge}
\begin{algorithmic}[1]
\REQUIRE Household personas $\mathcal{H}$, situation description $\mathcal{S}$, models $\{\mathcal{M}_i\}_{i=1}^N$ with generated plans $\{\mathcal{P}_i\}_{i=1}^N$, evaluation mode $mode \in \{\textsc{abs}, \textsc{rel}\}$, judge model $\mathcal{M}_\theta$
\ENSURE Evaluation results $R$ containing overall and per-persona outcomes

\STATE Initialize judge model $\mathcal{M}_\theta$ with $RPJ$ prompt conditioned on $\mathcal{H}$, $\mathcal{S}$, and $mode$

\IF{$mode\ is\ \textsc{abs}$}
    \STATE \textbf{Scoring:} All scores are integers in $\{1,\dots,5\}$, where $5$ denotes the highest quality.
    \FOR{each plan $\mathcal{P}_i$}
        \STATE Query $\mathcal{M}_\theta$ to assign an overall $score_o$ to $\mathcal{P}_i$
        \FOR{each household persona $h \in \mathcal{H}$}
            \STATE Query $\mathcal{M}_\theta$ to assign a per-persona $score_p$ to $\mathcal{P}_i$
        \ENDFOR
        \STATE Store overall and per-persona scores in $R$
    \ENDFOR
\ELSE
    \STATE Query $\mathcal{M}_\theta$ to rank all models $\{\mathcal{M}_i\}_{i=1}^N$ by overall usefulness of their plans
    \FOR{each household persona $h \in \mathcal{H}$}
        \STATE Query $\mathcal{M}_\theta$ to rank all models $\{\mathcal{M}_i\}_{i=1}^N$ from the perspective of $h$
    \ENDFOR
    \STATE Store overall and per-persona rankings in $R$
\ENDIF

\STATE \textbf{Return} $R$
\end{algorithmic}
\end{algorithm}
\end{small}

\section{Results and Discussion}
\label{sec:results}
\begin{table*}[t]
\caption{Model performance sorted by size. Proprietary models are \colorbox{proprietary}{shaded}. Bold and underline mark the best and second-best non--Gemini 2.5 Pro results. Confidence intervals for non-percentage metrics are reported in Appendix~\ref{app:proactive_additional_metrics}.}
\begin{center}
\resizebox{\linewidth}{!}{%
\begin{tabular}{lcccccccccccccc}
\hline
 &  & 
 \multicolumn{2}{c}{\textbf{\makecell{Information\\Grounding (ACC)}}} 
 & \multicolumn{2}{c}{\textbf{\makecell{Counterfactual\\Reasoning (ACC)}}}
 & \multicolumn{2}{c}{\textbf{\makecell{Multiple\\Choice (ACC)}}}
 & \multicolumn{2}{c}{\textbf{\makecell{Feature Reco.\\(MAP@1)}}}
 & \multicolumn{2}{c}{\textbf{\makecell{Plan Gen.\\($RPJ_{REL}$)}}}
 & \multicolumn{2}{c}{\textbf{\makecell{Plan Gen.\\(Validity)}}} \\

\cline{3-14}
\textbf{Model} & \textbf{\makecell{Input\\Modality}}
& \makecell{No\\ Tools} & \makecell{With\\ Tools}
& \makecell{No\\ Tools} & \makecell{With\\ Tools}
& \makecell{No\\ Tools} & \makecell{With\\ Tools}
& \makecell{No\\ Tools} & \makecell{With\\ Tools}
& \makecell{No\\ Tools} & \makecell{With\\ Tools}
& \makecell{No\\ Tools} & \makecell{With\\ Tools} \\
\hline

\rowcolor{orange!15}
\multicolumn{14}{l}{\textbf{Text-Based Data}} \\
\hline

Qwen3-4B
& Text
& 75.00 & 56.40
& 67.50 & 45.45
& 86.70 & 65.10
& 0.258 & 0.216
& 8.100 & 2.000
& 0.664 & 0.001 \\

Qwen3-4B [T]
& Text
& 75.70 & 32.60
& 71.00 & 36.35
& \underline{90.30} & 57.00
& 0.257 & 0.229
& 6.900 & 5.500
& 0.773 & 0.006 \\

GPT-OSS-20B [L]
& Text
& \underline{80.00} & 67.80
& 72.80 & 63.10
& 88.70 & \textbf{79.60}
& 0.242 & 0.211
& 9.700 & \underline{7.400}
& 0.753 & \underline{0.343} \\

GPT-OSS-20B [M]
& Text
& 79.94 & 68.90
& 72.81 & \textbf{63.65}
& 88.70 & 77.20
& 0.242 & 0.217
& \underline{10.000} & 5.500
& 0.725 & \textbf{0.415} \\

GPT-OSS-20B [H]
& Text
& \textbf{81.20} & 69.60
& \underline{74.00} & \underline{63.25}
& 89.20 & 75.20
& 0.247 & 0.229
& 8.500 & 4.400
& 0.698 & \textbf{0.413} \\

Qwen3-30B
& Text
& 73.50 & 64.14
& 63.55 & 52.35
& 83.20 & 61.70
& 0.260 & 0.229
& 5.800 & 1.900
& 0.645 & 0.002 \\

Qwen3-30B [T]
& Text
& 78.82 & 67.50
& \textbf{74.53} & 58.75
& \textbf{91.60} & 77.60
& 0.254 & \underline{0.244}
& 6.900 & 4.800
& 0.665 & 0.010 \\

Nemotron3-Nano-30B-A3B
& Text
& 75.62 & 64.90
& 70.83 & 57.95
& 86.20 & 73.80
& 0.259 & 0.239
& 9.400 & 5.700
& 0.825 & 0.010 \\

\colorbox{proprietary}{GPT-4o}
& Text
& \underline{80.00} & \underline{71.03}
& 71.30 & 60.94
& 86.30 & 75.00
& \textbf{0.277} & 0.227
& 6.100 & 5.100
& \underline{0.836} & 0.023 \\

\colorbox{proprietary}{GPT-4}
& Text
& 77.90 & \textbf{73.80}
& 73.45 & 62.90
& 86.30 & \underline{79.20}
& \underline{0.270} & \textbf{0.247}
& \textbf{19.300} & \textbf{22.300}
& \textbf{0.869 }& 0.022 \\

\hdashline
\colorbox{proprietary}{Gemini 2.5 Pro}
& Text
& 86.60 & 77.50
& 80.70 & 72.90
& 93.80 & 85.80
& 0.297 & 0.287
& 12.000 & 35.400
& 0.795 & 0.030 \\

\hline
\rowcolor{blue!10}
\multicolumn{10}{l}{\textbf{Multimodal Data}} & \multicolumn{2}{c}{\textbf{$RPJ_{ABS}$}} & & \\
\hline

\colorbox{proprietary}{Gemini 2.5 Pro}
& Text
& 67.74 & 63.31
& 67.74 & 68.82
& 85.18 & 80.59
& 0.209 & 0.205
& 3.660 & 3.130
& 0.753 & 0.508 \\

\colorbox{proprietary}{Gemini 2.5 Pro}
& Multimodal
& 77.49 & 70.03
& 76.08 & 77.16
& 88.95 & 84.50
& 0.197 & 0.163
& 2.850 & 3.250
& 0.697 & 0.405 \\

\hline
\end{tabular}
}
\end{center}
\label{tab:main_results_table}
\end{table*}

In this section, we summarize key findings from our experiments and analyze the impact of individual factors.

\subsection{Overall Results}
\label{sec:main_results}
Table~\ref{tab:main_results_table} reveals a consistent pattern across models and tasks: performance drops substantially when moving from \textit{Sole Reasoning} to agentic execution with tools. 
This degradation affects both reactive and proactive tasks, indicating that effective tool interaction remains a primary bottleneck. 
While larger models achieve higher absolute performance, scale alone does not close this gap, suggesting that dominant failures arise from interaction-level reasoning rather than insufficient model capacity. 
Among evaluated models, \textsc{GPT-4} achieves the strongest results in most settings; beyond this, no single model consistently dominates across tasks, modalities, and execution regimes.

\textbf{Reactive Tasks.}
Under \textit{Sole Reasoning}, models perform strongly on \textsc{IG} and \textsc{MC}, with many exceeding 85\% accuracy. Introducing tools causes sharp declines, frequently on the order of tens of points, even for large models. 
The most severe degradations occur for \textsc{IG} and \textsc{CF}, while \textsc{MC} shows slightly smaller but still substantial drops, likely due to its constrained answer space. These trends indicate that failures in retrieval and integration of relevant household information dominate error patterns in agentic settings.

\textbf{Proactive Tasks.}
\textsc{FR} exhibits moderate performance under \textit{Sole Reasoning} and shows the smallest relative drop when tools are introduced, likely because it is formulated as a predictive ranking problem rather than free-form generation. Nevertheless, even large models struggle to maintain strong ranking quality with tools, indicating persistent difficulty in mapping household context to relevant appliance features.

For \textsc{PG}, $RPJ_{REL}$ evaluation shows that although \textsc{GPT-4} captures a large share of wins, many other models obtain non-trivial win rates, producing a dispersed preference distribution. 
This dispersion suggests substantial variability in plan quality and indicates that models often generate plans that are only partially aligned with household preferences.

Plan \textit{validity} paints a more severe picture. 
Under \textit{Sole Reasoning}, most models generate highly valid structured plans. In agentic settings, validity collapses for nearly all text-based models, demonstrating that reliably producing runnable tool calls is a major issue even when high-level plans appear reasonable. 
A small number of models retain comparatively higher validity, but multimodal inputs do not improve executability and slightly reduce it.

\textbf{Reasoning Modes.}
Explicit reasoning variants (e.g., \textsc{[T/L/M/H]}) consistently improve performance under \textit{Sole Reasoning}, especially for reactive tasks, but these gains largely vanish in agentic settings and for proactive tasks. 
In several cases, higher reasoning levels fail to improve and can even degrade tool-mediated performance, indicating that additional internal thinking does not translate into better tool coordination or state tracking.

\textbf{ Multimodality.}
Multimodal inputs improve reactive performance under \textit{Sole Reasoning}, suggesting that visual grounding aids recovery of situational details. 
However, both proactive tasks degrade with multimodality, and once tools are introduced, differences between text-only and multimodal largely disappear. 
This indicates that perception alone does not address the dominant challenges in tool use and long-horizon planning.

\paragraph{Summary.}
Overall, the results expose three persistent weaknesses: brittle tool usage, difficulty reasoning over structured household context under partial observability, and limited ability to generate coherent, executable, and preference-aligned plans. These findings highlight that personalized agentic intelligence in realistic smart home environments remains far from solved.

\subsection{Additional Discussion}
\begin{figure}[ht]
  \begin{center}
    \centerline{\includegraphics[width=\linewidth]{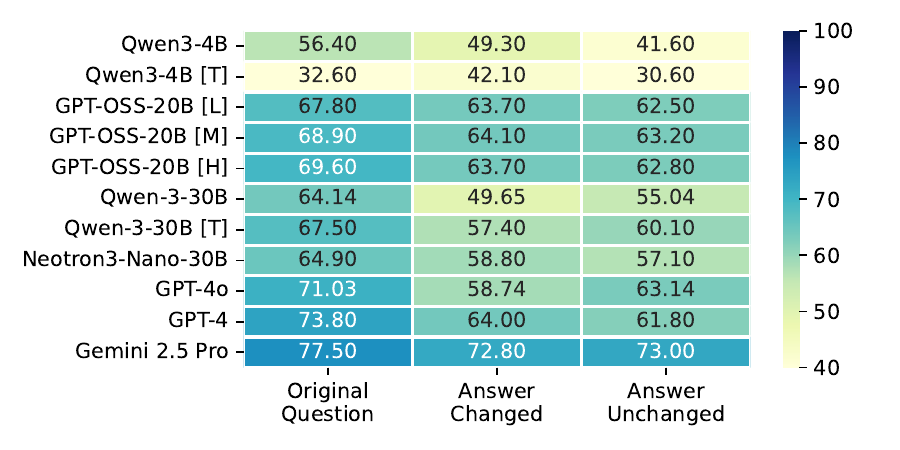}}
    \caption{
     Agentic Counterfactual Reasoning (\textsc{CF}) performance across models. The heatmap reports accuracy on original questions, on counterfactual cases with changed and unchanged answers.
    }
    \vspace{-0.5em}
    \label{fig:cf_heatmaps_with_ig_w_tools}
  \end{center}
\end{figure}

\textbf{Counterfactual Reasoning.}
Fig.~\ref{fig:cf_heatmaps_with_ig_w_tools} shows that for most models the degradation is larger in \textsc{$CF_{unchanged}$} than in \textsc{$CF_{changed}$}, indicating difficulty in preserving correct conclusions under hypothetical perturbations; \textsc{GPT-4o}, \textsc{Gemini 2.5 Pro}, and \textsc{Qwen3-30B} (with and without reasoning) are partial exceptions. The most severe failures occur for \textsc{Qwen3-4B [T]}, suggesting that smaller models, especially with explicit reasoning modes, can be particularly brittle. Interestingly, \textsc{Qwen3-4B [T]} is also the only model showing higher accuracy in \textsc{$CF_{changed}$} than on original questions, reflecting unstable counterfactual behavior rather than robust generalization. 
See Appendix Fig.~\ref{fig:cf_heatmaps} for \textit{Sole Reasoning}.

\textbf{Impact of Personalization and PersonalHomeTools.}
Table \ref{tab:pers_delta_tools} analyzes the effect of personalization by comparing performance differences between personalized and generic settings, where user-specific attributes are replaced with anonymized identifiers, reducing the need for individualized reasoning. 
Under \textit{Sole Reasoning}, personalization consistently degrades performance for both \textsc{IG} and \textsc{CF} across all models, with large drops indicating that reasoning over personalized household details introduces substantial additional complexity. In contrast, when using \emph{PersonalHomeTools} (i.e. With Tools), these drops are dramatically reduced and, in several cases, reversed into modest gains. 
This pattern validates the core design of our \emph{PersonalHomeTools}, demonstrating that its structured, tool-mediated access to household information enables models to adapt effectively to personalized contexts by retrieving user-specific details on demand rather than relying on brittle implicit inference.
Additional results are in the Appendix Tables \ref{tab:no_pers_w_tools} and \ref{tab:no_pers_no_tools}.

\begin{table}[ht]
\caption{\textbf{Effect of personalization}, measured as accuracy difference ($\Delta = Personalized \minus Generic$)}
\begin{center}
\resizebox{0.95\linewidth}{!}{%
\begin{tabular}{lcccc}
\hline
 & \multicolumn{2}{c}{\textbf{IG}} 
 & \multicolumn{2}{c}{\textbf{CF}} \\
\cline{2-5}
\textbf{Model}
& No Tools & With Tools
& No Tools & With Tools \\
\hline

GPT-OSS-20B [H]
& \down{18.34} & \up{3.83}
& \down{22.54} & \up{1.42}\\

GPT-4
& \down{21.88} & \down{0.53}
& \down{26.33} & \down{0.64}\\

Gemini 2.5 Pro
& \down{13.20} & \up{0.60}
& \down{19.10} & \up{1.10}\\

\hline
\end{tabular}%
}
\end{center}
\label{tab:pers_delta_tools}
\end{table}

\begin{figure}[ht]
  \begin{center}
    \centerline{\includegraphics[width=0.9\linewidth]{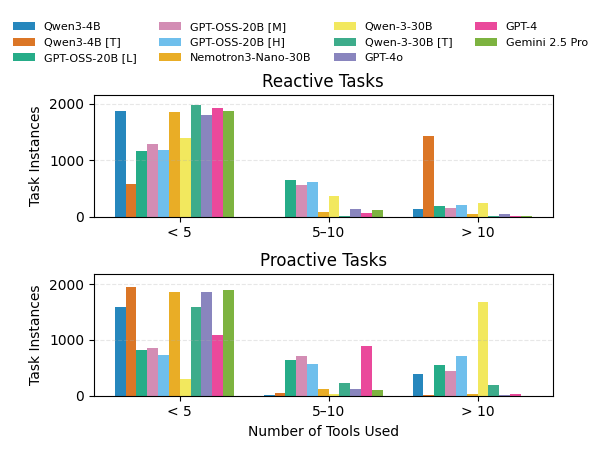}}
    \caption{Tool usage distributions across models for reactive (top) and proactive (bottom) tasks, grouped by low ($<5$), medium ($5$–$10$), and high ($>10$) numbers of calls per task instance.
    }
    \label{fig:num_tools_dist}
  \end{center}
\end{figure}

\textbf{Tool Usage Patterns.}
Fig. \ref{fig:num_tools_dist} shows that proactive tasks generally trigger more tool calls than reactive tasks, reflecting their multi-step nature, while larger and reasoning-enabled models tend to invoke tools more frequently than smaller models. However, higher tool usage does not consistently translate into better performance, underscoring that effective coordination and timing of tool use, rather than sheer frequency, is the primary challenge.





\textbf{Impact of Reflection.}
We repurpose our Role Playing Judge (RPJ) as a self-reflection signal, enabling models to iteratively critique and revise generated plans in the spirit of Reflexion \cite{shinn2023reflexion}. $RPJ$ feedback highlights preference mismatches and tool-level errors, guiding targeted plan refinement in multimodal settings.
As shown in Table~\ref{tab:reflection_ablation}, reflection consistently improve both RPJ$_{ABS}$ scores and plan validity, indicating more executable and better-aligned plans for both Text-Only and Multimodal settings. 

\begin{table}[ht]
\caption{Reflection-based ablations for Agentic \textsc{Gemini 2.5 Pro}. We report average $RPJ_{\textsc{ABS}}$ and Validity for Plan Generation.}
\label{tab:reflection_ablation}
\centering
\resizebox{0.9\linewidth}{!}{%
\begin{tabular}{lcccc}
\hline
\textbf{Setting} 
& \multicolumn{2}{c}{\textbf{Text-only}} 
& \multicolumn{2}{c}{\textbf{Multimodal}} \\
\cline{2-5}
& \textbf{$RPJ_{ABS}$} & Validity 
& \textbf{$RPJ_{ABS}$} & Validity \\
\hline
Base            & 3.380 & 0.030 & 3.250  & 0.405   \\
w/ Reflection   & 3.520 & 0.428 & 4.240  & 0.496   \\
\hline
\end{tabular}

}
\end{table}

\textbf{Human Analysis.}
LLM-as-a-Judge is known to be imperfect \cite{ICLR2025_4264ee43}, and is commonly validated with human assessment for subjective tasks. Accordingly, we asked four human annotators to evaluate $RPJ$ outputs, marking disagreement, partial agreement, or full agreement with both overall and persona-level scores. We observe strong alignment, with no cases of complete disagreement and over 65\% inter-annotator agreement (Appendix Fig. ~\ref{fig:human_assess}). This indicates that $RPJ$ provides a reliable and meaningful signal for assessing plan quality and personalization.

\begin{figure}[h]
    \centering
    \begin{subfigure}[b]{\linewidth}
        \centering
        \includegraphics[width=\linewidth]{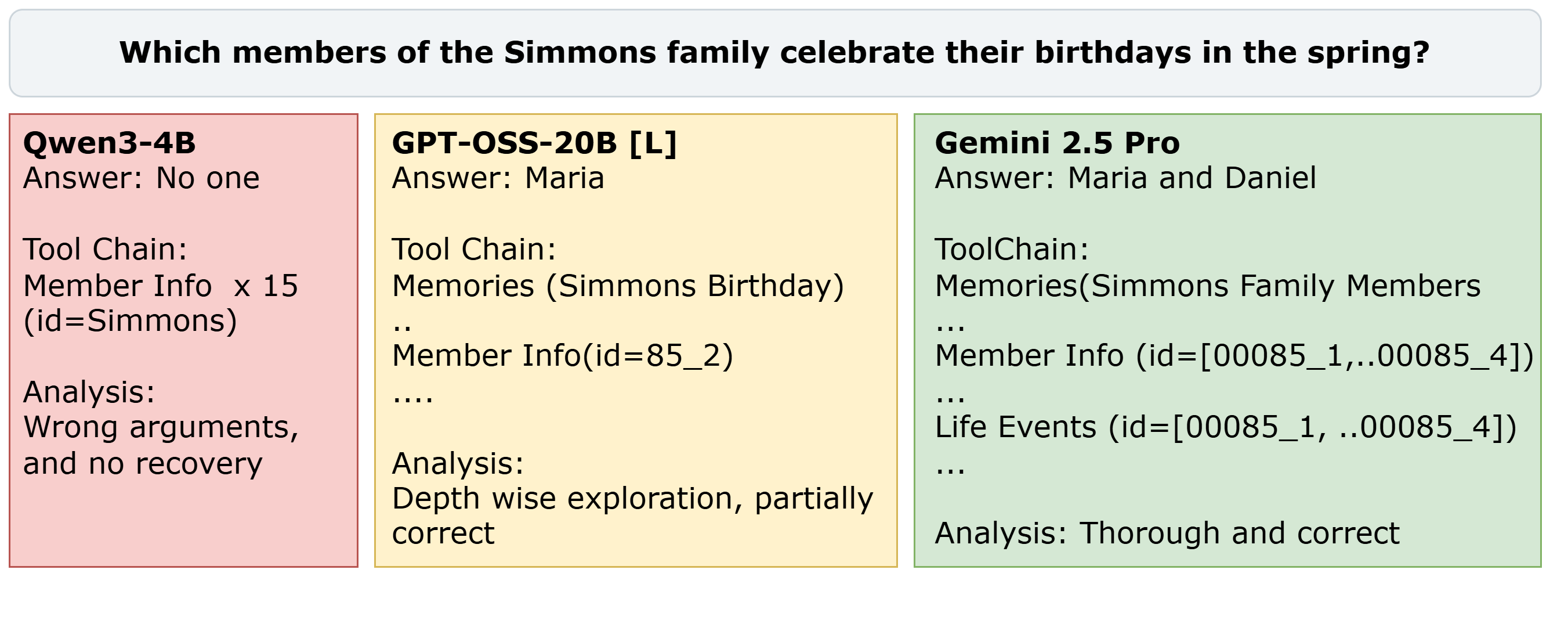}
        \caption{Reactive Task (IG) Example}
        \label{fig:reactive_qual}
    \end{subfigure}
    \begin{subfigure}[b]{\linewidth}
        \centering
        \includegraphics[width=\linewidth]{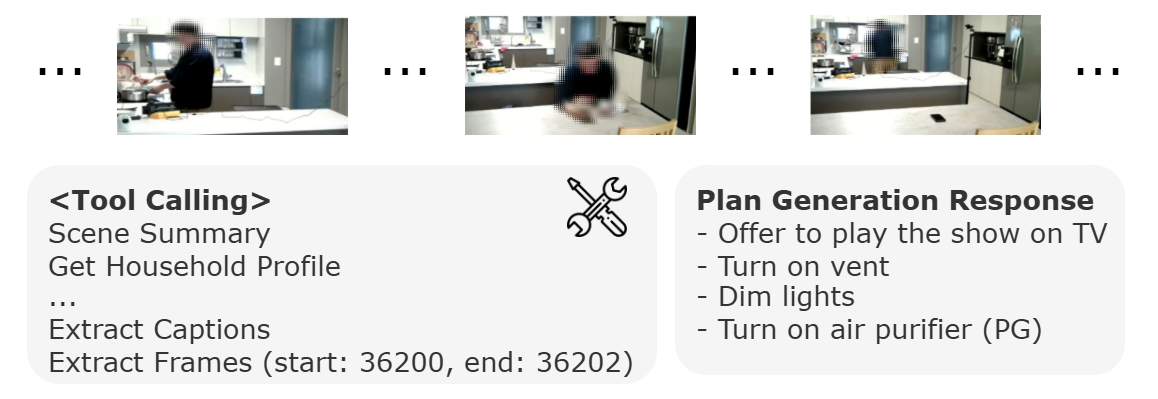}
        \caption{Proactive Task (PG) Example}
        \label{fig:proactive_qual}
    \end{subfigure}%
    \caption{Qualitative examples illustrating personalized agent behavior in PersonalHomeBench, covering reactive question answering (top) and multimodal proactive planning (bottom).}
    \label{fig:qualitative_examples}
\end{figure}

\textbf{Qualitative Analysis.}
Fig. \ref{fig:qualitative_examples} illustrates representative success and failure modes for both reactive and proactive settings. 
In Fig. \ref{fig:reactive_qual}, weaker models make redundant or misdirected tool calls and fail to recover from early errors, leading to incorrect or partial answers, whereas stronger models issue targeted queries over memories and member information and assemble coherent evidence before answering. 
Fig.  \ref{fig:proactive_qual} shows that effective proactive behavior begins with acquiring situational context, followed by synthesizing multi-step appliance-control plans.
Even when plans appear sensible, models frequently fail to generate fully correct tool invocations, and these examples (c.f. Appendix~\ref{app:qual_analysis}) mirror the quantitative results, underscoring that reliable tool usage and state-aware coordination remain dominant limitations.

Refer to appendix for tool usage failure analysis (\ref{appsec:tool_usage_failures}), limitations (\ref{sec:limitations}), and future directions.


\section*{Conclusion}
We presented PersonalHomeBench, a benchmark for evaluating personalized agentic behavior in realistic smart home environments, together with PersonalHomeTools, a domain-specific toolbox for grounded interaction with household context, memory, perception, and device control. 
Extensive evaluation across foundation models demonstrates that strong performance under full observability does not translate to reliable tool usage or proactive behavior, and that counterfactual reasoning and personalized plan generation remain challenging. 
Improvements from scaling, explicit reasoning modes, and multimodal inputs are limited and inconsistent.
These results suggest that key challenges remain in combining reasoning, tool usage, and personalization, and we hope that PersonalHomeBench will facilitate more systematic study of these issues and support future advances in robust personalized agentic systems.

\clearpage 

\section*{Impact Statement}
This work aims to advance research on personalized and proactive agentic systems, with a focus on smart home environments, and carries both potential benefits and risks. 
On the positive side, improved personalization can enhance accessibility, safety, child supervision, and overall quality of life, particularly for elderly users and individuals with special needs.
At the same time, agentic technologies operating in personal spaces raise important issues around privacy, data security, and misuse.
Our inclusion of compact 4B-parameter models also sheds light on the feasibility of privacy-preserving, on-device deployment, helping mitigate security and data exposure concerns.  
By releasing a benchmark and toolbox designed for systematic evaluation, we aim to support responsible development grounded in rigorous analysis. We encourage future work to prioritize safeguards, transparency, and informed user consent. 
Overall, we view the primary impact of this work as enabling more careful, accountable, and principled progress in personalized and proactive agentic systems.

\bibliography{home_aware_icml/personalhomebench}
\bibliographystyle{icml2026}

\newpage
\appendix
\onecolumn

\newcommand{\mainlinks}{%
\begin{tcolorbox}[
  breakable,
  colback=gray!5,
  colframe=green!20,
  coltitle=green!20!black,
  title=Links to the main draft sections
]
Introduction (\ref{sec:intro}), 
Related Works (\ref{sec:related_works}), 
PersonalHomeBench (\ref{sec:personal_home_bench}), 
PersonalHomeTools (\ref{sec:toolbox}), 
Experimental Setup (\ref{sec:experiments}), 
and Results (\ref{sec:results}).
\end{tcolorbox}}

\section{Outline}
The appendix provides supplementary material that supports and extends the main paper, including additional analyses, experimental details, and resources omitted from the main text due to space constraints. Specifically, we include:

\begin{itemize}
\item Sec~\ref{app:contributions}: Presents the distribution of technical contributions
\item Sec~\ref{sec:limitations}: Discusses key limitations of our work and outlines promising directions for future research.
\item Sec.~\ref{app:related_works}: Additional related work and extended comparisons.
\item Sec.~\ref{app:additional_hh_stats}: Further statistics and analysis of households in PersonalHomeBench.
\item Sec.~\ref{app:additional_metrics}: Extended experimental results and auxiliary evaluation metrics.
\item Sec.~\ref{app:models}: Model descriptions and hyperparameter settings.
\item Sec.~\ref{app:schemas}: Data schemas used in the benchmark.
\item Sec.~\ref{app:prompts}: Prompts used for simulated and semi-simulated data generation.
\item Sec.~\ref{app:human}: Instructions provided to human annotators for RPJ evaluation and for video collection.
\item Sec.~\ref{app:qual_analysis}: Additional qualitative examples.
\item Sec.~\ref{app:toolbox}: Detailed documentation of all tools in PersonalHomeTools.
\end{itemize}

\section{Contributions}
\label{app:contributions}
\paragraph{Technical Contributions.}
A detailed breakdown of individual contributions is provided in Table~\ref{tab:app_contributions}. 
In summary, the work spans dataset development, experimental design and execution, human annotation, 
and manuscript preparation, with contributions distributed across all co-authors.

\paragraph{Acknowledgments.}
We would like to acknowledge several individuals for their guidance and support. 
\textbf{Ko Keun Kim} provided support and ideation for the prior version of the dataset and served as a primary leader on the then LG AI Research Korea Team. 
We thank \textbf{Kevin Ferreira} for his support as the then lab leader of the Toronto AI Lab, including guidance, paper review, and ideation. 
We also acknowledge \textbf{Youngjoon Kim} for leadership and strategic guidance throughout the project.

\paragraph{Gratitude.}
We sincerely thank all collaborators and contributors who supported this work, including those involved in earlier iterations, discussions, and feedback that helped shape the final outcome.
\begin{table}[ht]
\caption{Distribution of contributions by the researchers.}
\label{tab:app_contributions}
\small
\begin{tabular}{l l}
\toprule
\textbf{Contributor} & \textbf{Core Contributions (Not an exhaustive list)}  \\
\midrule
Manasa Bharadwaj & Lead Researcher and developer, paper drafting, reciprocal reviewing, rebuttal, and arxiv support\\
Yolanda Liu* & Lead Research Engineer for LLM experiments, support for paper drafting,  quality assurance\\
Injung (Kayla) Yang* & Lead Researcher for multimodal experimentation, public data and code maintenance, arxiv support\\
Sungil Kim & Support for various experimentation, human annotation\\
Nikhil Verma & Delivered figure 2, human annotation, rebuttal and arxiv support\\
\bottomrule
\end{tabular}
\end{table}

\section{Limitations and Future Work}
\label{sec:limitations}
We restrict multimodal benchmarking to \textsc{Gemini 2.5 Pro}, as most currently available multimodal foundation models either lack native video input support or exhibit very weak reasoning and planning capabilities, making them unsuitable for long-horizon agentic evaluation \cite{singh2025gpt5, tang2025videosalmonn, xu2025qwen2omni}. At present, \textsc{Gemini 2.5 Pro} is the only model that demonstrates competitive performance across perception, reasoning, and planning in our setting. As multimodal models continue to mature, future iterations of PersonalHomeBench will incorporate a broader set of video-capable agents and enable more comprehensive multimodal comparisons. PersonalHomeBench also focuses on non-embodied smart home assistance rather than physical navigation or low-level manipulation; extending the benchmark to hybrid embodied–device settings is an important direction for future work.

In addition, our Role Playing Judge enables scalable evaluation of subjective plan quality and personalized satisfaction, but remains an LLM-based proxy. Although human agreement analysis indicates strong alignment, future work will expand direct human evaluation, explore preference learning from human feedback, and investigate training-time integration of RPJ-style supervision. Together, these directions outline a path toward strengthening both the scope and the fidelity of personalized agent evaluation.
\section{Additional Related Works}
\mainlinks
\label{app:related_works}
\noindent\textbf{Reactive Agentic Benchmarks.}
A large body of work has established benchmarks for evaluating agentic behavior in web-based interaction \cite{NEURIPS2022_82ad13ec}, UI navigation \cite{NEURIPS2024_5d413e48}, game-like environments, and simulated worlds \cite{chang2024agentboard}. These benchmarks have standardized evaluation of multi-step instruction following and reasoning, but typically emphasize reactive task completion in environments with limited long-term context and minimal personalization. More recent tool-centric benchmarks evaluate whether foundation models can select and invoke general-purpose APIs, providing insight into basic tool usage \cite{qin2024toolllm}. However, the tools are largely domain-agnostic, and tasks rarely require sustained user modeling, household continuity, or proactive assistance. 

\noindent\textbf{Embodied Home Agentic Benchmarks.}
Embodied household benchmarks such as VirtualHome \cite{Puig2018VirtualHomeSH}, ALFRED \cite{Shridhar2019ALFREDAB}, BEHAVIOR \cite{li2024behavior1k}, and PARTNR \cite{chang2024partnr} evaluate agents in simulated home environments with navigation and atomic action execution. 
In contrast, PersonalHomeBench focuses on non-embodied, real-world home environments, using long-horizon videos captured in diverse households to reflect natural daily routines, health, and safety scenarios. 

\noindent\textbf{Proactive Agentic Benchmarks.}
Current proactive LLM agents cover specialized frameworks for clarification \cite{Zhang2024AskbeforePlanPL}, multi-agent prediction \cite{Lu2024ProactiveAS}, and UI-monitored task assistance \cite{Lu2024ProactiveAS, Zhao2025CodingGenieAP}.
ContextAgentBench \cite{yang2025contextagent} takes an important step toward evaluating proactive agents with contextual information and tool usage, but focuses primarily on ego-centric daily traces, relies on generic tools, and operates in language-only inference settings. In contrast, our benchmark targets shared household environments, uses domain-specific smart home tools, and evaluates both reactive and proactive behaviors grounded in persistent, personalized context with optional multimodal inputs.

\noindent\textbf{Personalization in Interactive Systems.}
Personalization has been studied through persona consistency benchmarks, long-term memory datasets, and user-specific question answering tasks. These benchmarks primarily assess whether models can recall or remain consistent with user attributes in text-based interactions \cite{salemi-etal-2024-lamp, maharana-etal-2024-locomo}. Parallel work explores personalization for narrow planning problems, such as personalized travel assistants with limited tool support \cite{singh-etal-2024-personal}. While valuable, these settings treat personalization largely as a property of dialogue or retrieval. PersonalHomeBench treats personalization as a central driver of interactive decision making, assessing how user preferences influence reasoning, tool usage, and long-horizon planning in dynamic environments, and how this information is leveraged when evaluating generated plans.

\noindent\textbf{Home-Centric Benchmarks.}

Unlike datasets such as Toyota Smarthome \cite{Das_2019_ICCV}, which are staged, demographically narrow, and not designed around smart-device interaction, our data spans diverse home layouts, age groups, relationships, and unprompted activities. 
Videos range from seconds to tens of minutes and may cover multiple activities (e.g., cooking, leisure, sleep), enabling evaluation of persistent context tracking and proactive assistance over extended horizons.

Recent language-model-based home assistant benchmarks, such as HomeBench \cite{li2025homebench} and EdgeWisePersona \cite{bartkowiak2025edgewisepersona}, move closer to smart home scenarios but remain limited. HomeBench focuses on reactive validation and execution of smart home commands without modeling personalization, while EdgeWisePersona emphasizes profile reconstruction and lightweight proactive routine inference with a small set of devices and simple state transitions. 

\section{Additional Household Statistics}
\mainlinks
\label{app:additional_hh_stats}

\subsection{Text-based Data Statistics}
\begin{figure*}[t]
    \centering
    \begin{subfigure}{0.33\linewidth}
        \centering
        \includegraphics[width=\linewidth]{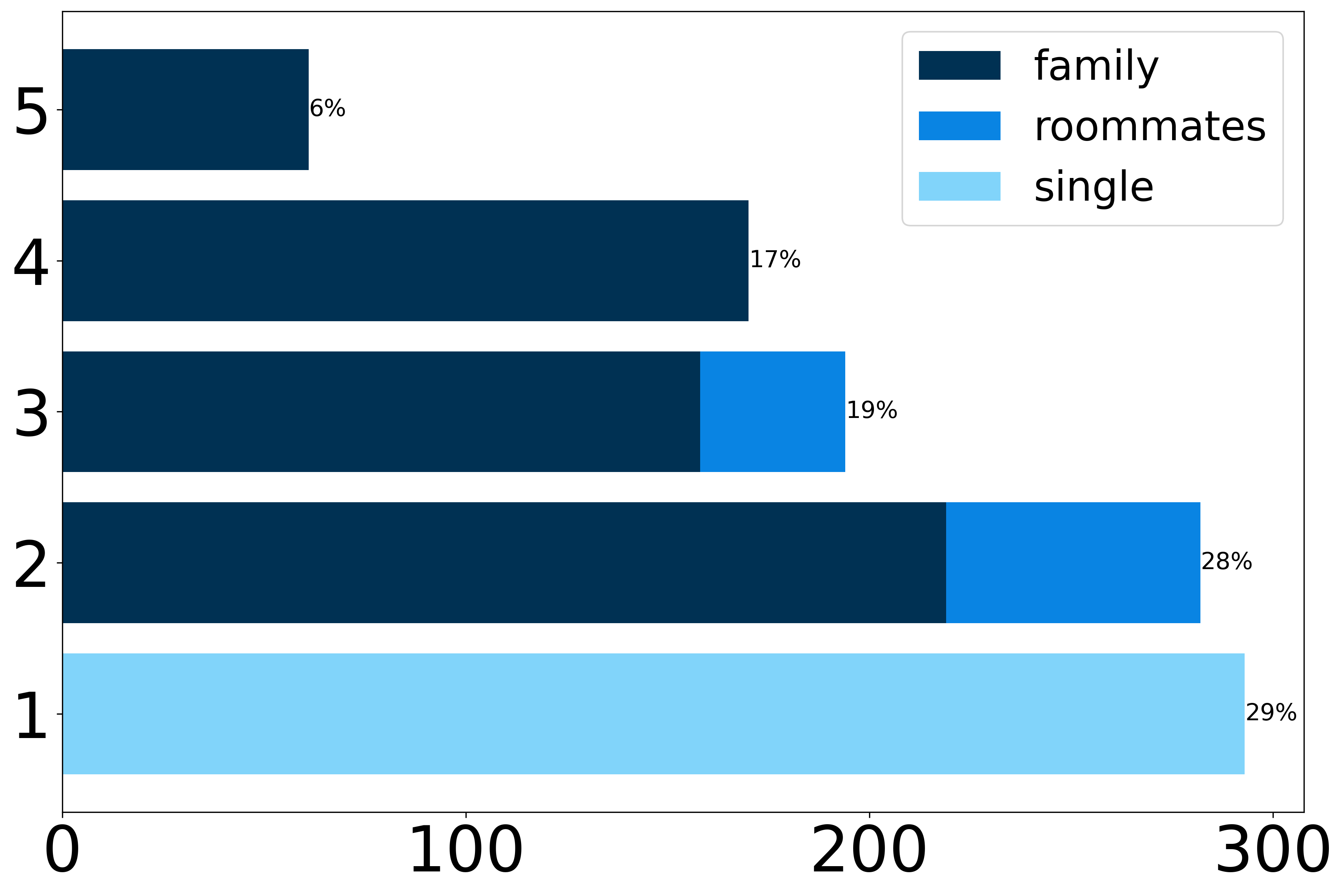}
        \caption{Household Structure}
        \label{fig:hh_by_size}
    \end{subfigure}
    \hfill
    \begin{subfigure}{0.33\linewidth}
        \centering
        \includegraphics[width=\linewidth]{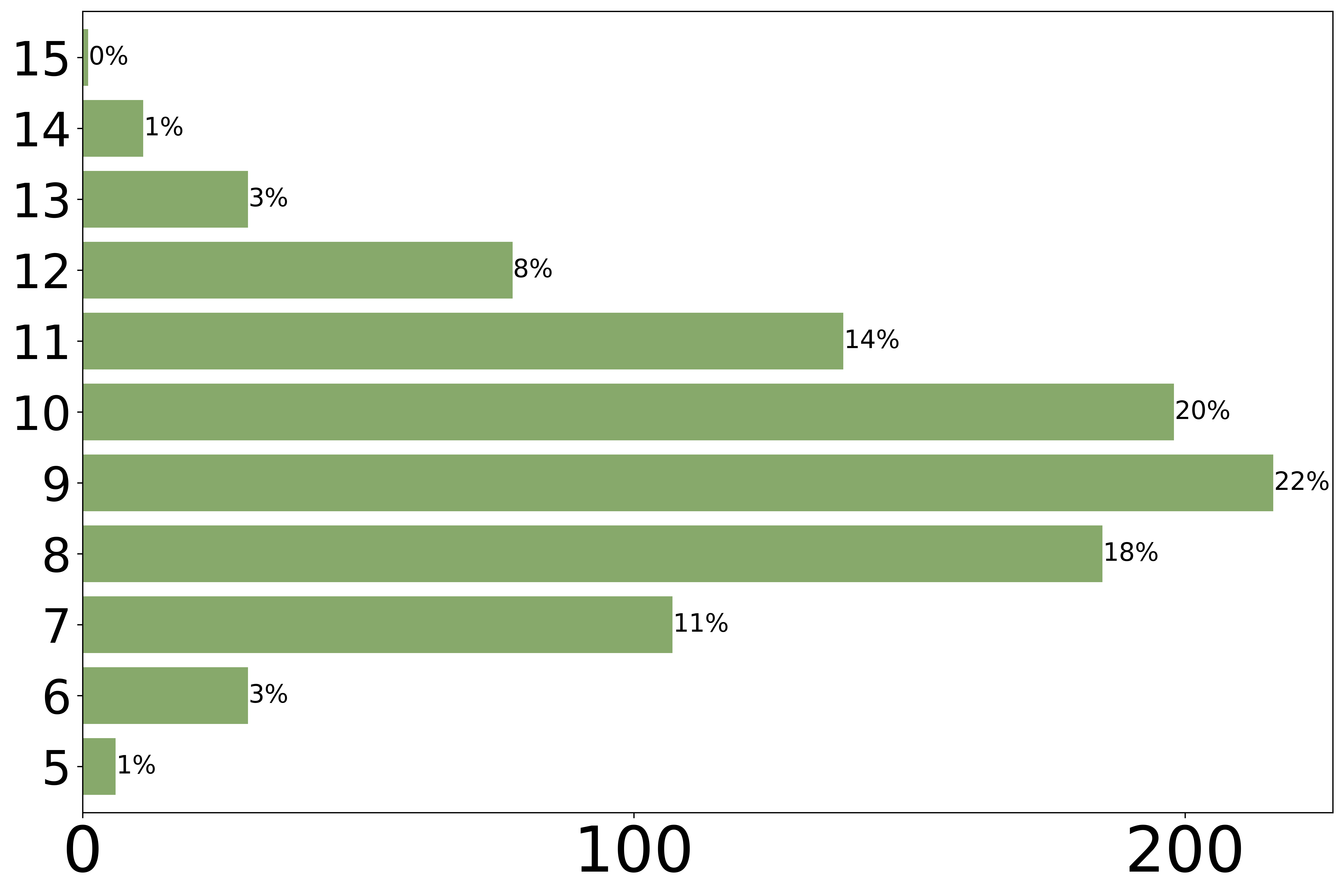}
        \caption{\#Appliances Owned}
        \label{fig:hh_by_appliance}
    \end{subfigure}
    \hfill
    \begin{subfigure}{0.33\linewidth}
        \centering
        \includegraphics[width=\linewidth]{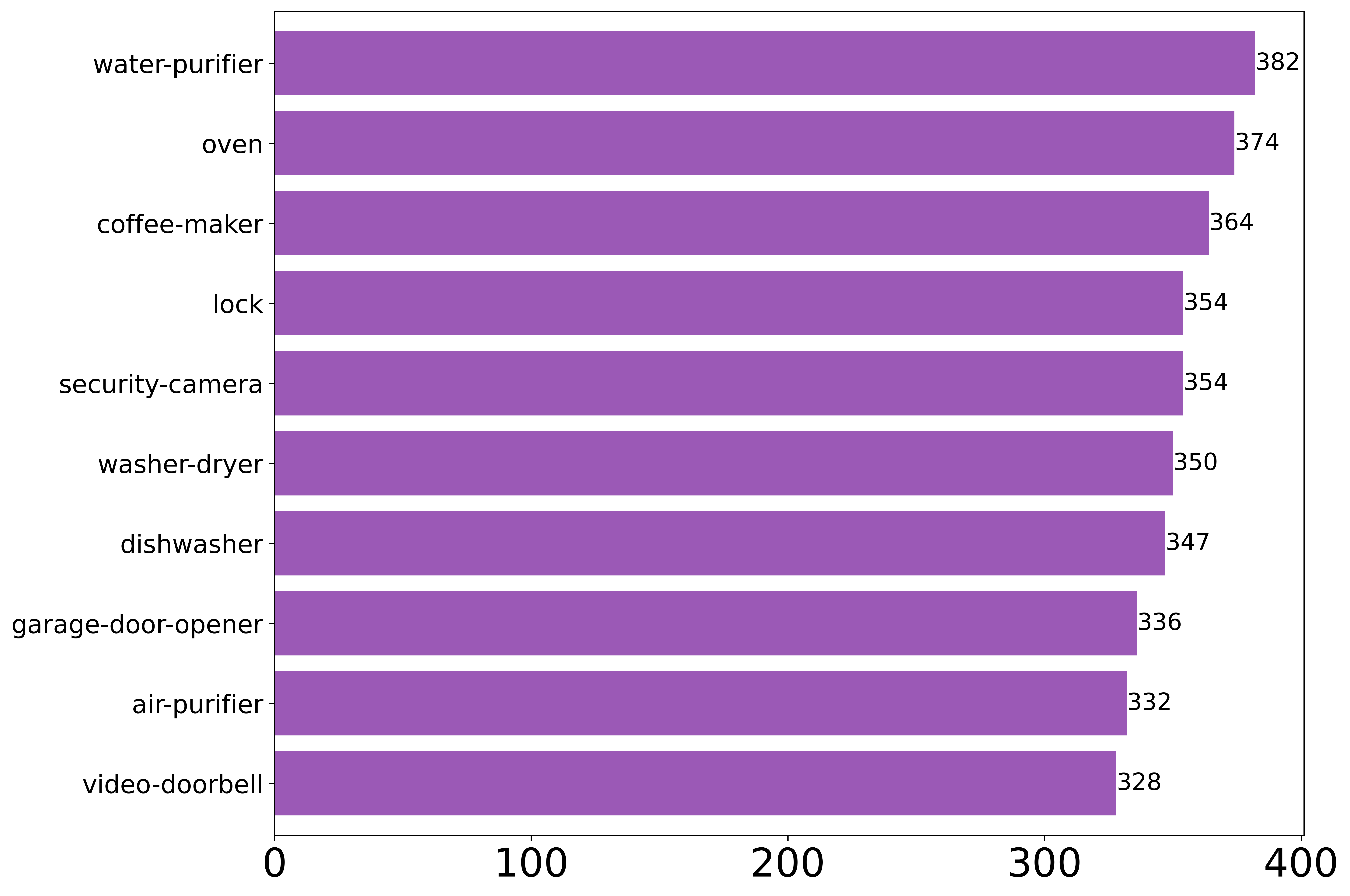}
        \caption{Common Appliances Owned}
        \label{fig:top_appliances}
    \end{subfigure}
    \caption{Additional statistics of generated household in PersonalHomeBench. (a) Distribution of household size and composition. (b) Distribution of appliances owned in the households. (c) Common appliances owned by the households.}
    \label{fig:app_text_data_hh_stats}
\end{figure*}

\textbf{Household Size.} The households capture a structured variety of domestic environments, categorized into: single occupant, family and non-related roommates. The range of household scales from single-occupancy to large-family residences (Fig.~\ref{fig:hh_by_size}).

\textbf{Appliances Owned.}
The dataset features nearly 40 distinct appliance types, including five core appliances present in all households: refrigerator, thermostat, microwave, speaker and TV. Other top most prevalent appliances are listed in (Fig. ~\ref{fig:top_appliances}).

\textbf{Pets Distribution.}
Across 409 households, a total of 823 animals are represented in this dataset. The pet population is categorized into nine animal types: cat, dog, hamster, lizard, guinea pig, rabbit, bird, fish and turtle - spanning a wide range of common domestic mammals, reptiles and aquatic companions. 

\subsection{Multimodal Data Statistics}
\label{app:video_instructions}
Table~\ref{tab:video_scenario_domains} presents the distribution of the 100 scenarios constructed for this study across three core domains: safety care, health care, and daily care. Health care scenarios constitute the largest portion, reflecting the prevalence of situations involving well-being, medication, and environmental health in household settings. Daily care scenarios capture routine activities such as returning home and pet care, while safety care scenarios focus on risk-sensitive situations and accident prevention. Scenarios are categorized based on consistent semantic cues in their titles, enabling clear domain delineation while preserving coverage of diverse, realistic household situations spanning safety, health, and everyday routines.

Table \ref{tab:app_video_demographics} summarizes the participant composition of the real home data used in our benchmark across health care, safety care, and daily care scenarios. The dataset captures substantial diversity in household roles, including adults, caregivers, children, and pets, spanning a wide range of age groups from young children to older adults in their 70s. 
Safety care scenarios are the most prevalent, reflecting common real-world situations that involve risk monitoring and intervention, particularly in households with both caregivers and children. 
Daily care scenarios further emphasize routine, multi-occupant household dynamics, while health care scenarios focus on age-sensitive contexts involving adults across multiple life stages. 
Overall, this distribution highlights the benchmark’s emphasis on realistic household diversity and interaction patterns, providing a representative foundation for evaluating personalized and proactive assistance in everyday home environments.
All the data is ego-centric views, and collected through safe placement of cameras in subject's homes. 

\paragraph{Instructions for Video Data Collection Participants.}
Participants were provided with scenario-specific guidance corresponding to the domains listed in Table~\ref{tab:video_scenario_domains} after household setup was completed, including camera placement, microphone activation, and sensor checks. 
For each scenario, participants received high-level instructions describing how the scene should unfold and the types of actions to perform, while allowing natural variation in execution. 
For scenarios involving safety or hazardous situations, participants were instructed to simulate the actions rather than perform them physically, and on-site staff ensured that all safety protocols were followed. 
Scenes with multiple participants were coordinated by coaching all subjects involved to maintain consistency in the intended narrative. 
To ensure usable recordings, scenes were re-shot when necessary to improve clarity or coverage. 
The source data were collected under a service contract with an external vendor and processed in compliance with applicable personal data protection regulations following a personal information impact assessment. Participants were compensated for their participation.

\begin{table}[ht]
\caption{Distribution of scenario domains and classification criteria used in the benchmark.}
\label{tab:video_scenario_domains}
\centering
\begin{small}
\begin{tabular}{lcl}
\toprule
\textbf{Domain} & \textbf{Video Count} & \textbf{Classification Basis} \\
\midrule
Safety Care & 30 & Scenario titles reference safety, risk, hazard, or accident prevention \\
Health Care & 38 & Scenario titles reference health, medication, ventilation, or well-being \\
Daily Care / Routine & 32 & Scenario titles reference daily routines, returning home, pet care, or going out \\
\midrule
\textbf{Total} & \textbf{100} & --- \\
\bottomrule
\end{tabular}
\end{small}
\end{table}

\begin{table}[ht]
\caption{Participant demographics across scenarios in the collected home data.}
\label{tab:app_video_demographics}
\centering
\small
\begin{tabular}{l l l l r}
\toprule
\textbf{Scenario} & \textbf{Participant Type} & \textbf{Age Group} & \textbf{Gender} & \textbf{Participant Count} \\
\midrule
Health Care & Adult & 30s & Male & 2 \\
Health Care & Adult & 60--70s & Male & 3 \\
Health Care & Adult & 60--70s & Female & 3 \\
\midrule
\textbf{Health Care (Total)} &  &  &  & \textbf{8} \\
\midrule
Safety Care & Adult (Caregiver) & 30s & Female & 5 \\
Safety Care & Adult (Caregiver) & 40--50s & Female & 3 \\
Safety Care & Adult (Caregiver) & 40--50s & Male & 2 \\
Safety Care & Adult & 30s & Male & 1 \\
Safety Care & Child & Under 10 & Male & 3 \\
Safety Care & Child & Under 10 & Female & 5 \\
Safety Care & Child & 5 and over & Female & 1 \\
Safety Care & Child & 10 and over & Female & 2 \\
\midrule
\textbf{Safety Care (Total)} &  &  &  & \textbf{22} \\
\midrule
Daily Care & Adult & 60s & Male & 1 \\
Daily Care & Adult & 70s & Male & 1 \\
Daily Care & Adult & 70s & Female & 1 \\
Daily Care & Adult & 30s & Male & 1 \\
Daily Care & Adult (Caregiver) & 30s & Female & 2 \\
Daily Care & Adult (Caregiver) & 40--50s & Female & 2 \\
Daily Care & Child & Under 10 & Female & 3 \\
Daily Care & Child & Under 10 & Male & 1 \\
Daily Care & Child & 5 and over & Female & 1 \\
Daily Care & Child & 10 and over & Female & 2 \\
\midrule
\textbf{Daily Care (Total)} &  &  &  & \textbf{15} \\
\midrule
Pets & Animal & -- & Cat & 1 \\
\bottomrule
\end{tabular}
\end{table}

\section{Additional Results and Metric Definitions}
\mainlinks
\label{app:additional_metrics}
\subsection{Tool Usage Failure Analysis}
\label{appsec:tool_usage_failures}
To clarify failure modes, we categorize errors into three types:
\begin{itemize}
    \item \textbf{Reasoning Errors.}
    Failures where the model does not arrive at the correct answer despite sufficient context. Our \textit{Sole-Reasoning} setting directly captures this class. (See Section~\ref{sec:main_results} for the main results.)
    
    \item \textbf{Reasoning \& Planning Errors.}
    Failures to construct effective multi-step plans, leading to incorrect outcomes. This corresponds to the \textit{agentic setting} in our benchmark. (See Section~\ref{sec:main_results} for the main results.)
    
    \item \textbf{PG Plan Errors.}
    We evaluate PG program validity holistically, capturing parameter mismatches, signature violations, type errors, and execution failures. Beyond static analysis, each program is executed step-by-step with environment resets, enabling detection of dynamic interaction errors. (See Section~\ref{sec:main_results} for the main results.)

    \item \textbf{Tool Usage Errors.} The errors arise in the agentic setting, where models actively gather information and interact with tools to solve tasks. We analyze these errors and their corresponding patterns below. 
\end{itemize}

\begin{figure}[ht]
  \vskip 0.2in
  \begin{center}
    \centerline{\includegraphics[width=1.1\columnwidth]{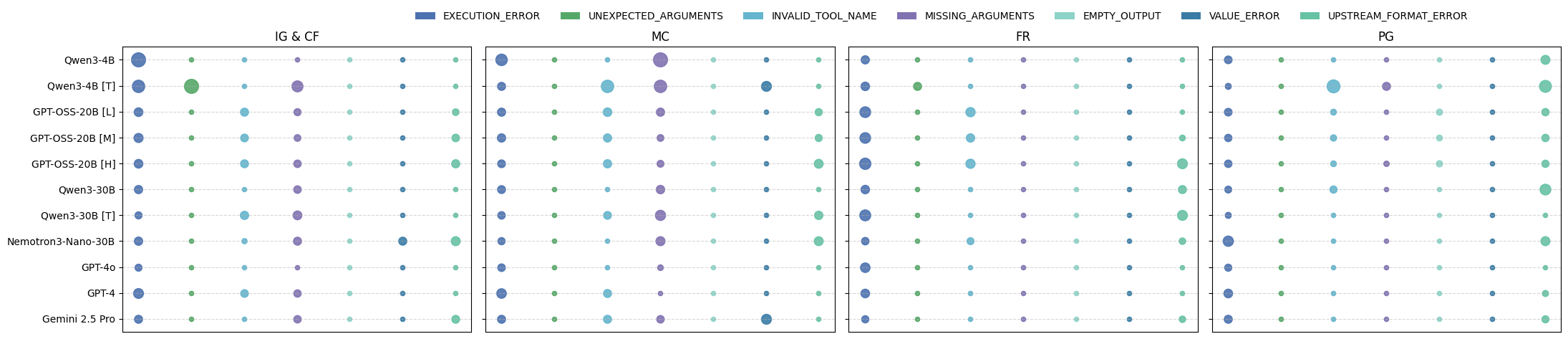}}
    \caption{
      \textbf{Error distribution across models, tasks, and failure types.}
Bubble plots show the average frequency of error types across four task settings (IG, CF, MC, FR, PG). Rows correspond to models and columns to error categories, with bubble size proportional to error frequency.
    }
    \label{fig:app_tool_err}
  \end{center}
\end{figure}

Figure~\ref{fig:app_tool_err} presents a bubble plot visualization of error distributions across models (y-axis) and error types (x-axis), with bubble size proportional to the average frequency of each error. We analyze trends across four task settings: \textbf{IG, CF}, \textbf{MC}, \textbf{FR}, and \textbf{PG}, revealing consistent structural patterns in model failure modes as well as task-specific sensitivities.

\paragraph{Dominance of Execution Errors.}
Across all tasks, \texttt{EXECUTION\_ERROR} emerges as the most prominent failure mode, particularly for smaller and mid-sized models (e.g., Qwen3-4B, Qwen3-30B, Nemotron3-Nano-30B). The consistently large bubbles indicate that models frequently fail at the stage of tool execution rather than earlier reasoning or planning. This suggests that even when models produce syntactically valid plans, they struggle to satisfy runtime constraints. Notably, even stronger models (e.g., GPT-4o, Gemini 2.5 Pro) do not eliminate this failure mode entirely, indicating that execution robustness remains an open challenge.

These include scenarios such as invoking tools with non-existent member IDs, querying appliances that are not associated with a given household, or bypassing necessary intermediate steps (e.g., directly calling an appliance status API without first retrieving the valid appliance list via a preceding tool). Such errors highlight failures in state tracking and procedural grounding rather than surface-level formatting or schema adherence.

\paragraph{Argument-Level Errors: Planning vs.\ Grounding Gap.}
Errors such as \texttt{MISSING\_ARGUMENTS} and \texttt{UNEXPECTED\_ARGUMENTS} form the second dominant cluster. These are especially pronounced in \textbf{IG, CF} and \textbf{MC} settings, where multiple tool calls are required to gather sufficient information. The prevalence of these errors indicates a systematic gap between high-level planning and low-level parameter grounding:
\begin{itemize}
    \item \texttt{MISSING\_ARGUMENTS} suggests incomplete decomposition or failure to propagate required variables across steps.
    \item \texttt{UNEXPECTED\_ARGUMENTS} indicates over-generation or hallucinated parameters not supported by the tool schema.
\end{itemize}
These two error types often co-occur across models, reinforcing the hypothesis that argument specification errors arise from inconsistencies in intermediate representations rather than isolated local failures.

\paragraph{Tool Selection vs.\ Tool Usage.}
\texttt{INVALID\_TOOL\_NAME} errors are relatively infrequent compared to argument-related issues, with generally smaller bubbles across all tasks. This suggests that models are largely capable of selecting the correct tool from the available set. However, the contrast between low tool-selection errors and high execution/argument errors highlights a key bottleneck: \emph{Models often know which tool to use, but not how to use it correctly.}

This distinction implies that improvements in tool grounding and schema alignment may yield larger gains than improvements in tool retrieval or selection.

\paragraph{Task-level error trends: Reactive vs. Proactive settings.}
Grouping tasks by their information retrieval demand reveals a structurally important asymmetry in 
how errors surface. In \textit{reactive question-answering} tasks (IG, CF, MC), models 
must invoke the correct sequence of tools and ground their answers in retrieved environmental 
state, there is no plausible path to a correct answer without tool compliance. This 
obligation is reflected in the bubble plots: IG, CF and MC consistently show the largest and most concentrated error bubbles, dominated primarily by \texttt{EXECUTION\_ERROR} and 
\texttt{MISSING\_ARGUMENTS}, precisely because any deviation from a valid, complete tool call is immediately penalized. 

In contrast, \textit{proactive recommendation} tasks (FR, PG) 
impose no such hard grounding constraint, a model can produce a syntactically coherent plan 
or feature recommendation using arbitrarily little retrieved context, and such outputs are 
not trivially incorrect in the way a wrong lookup answer is. This leniency is reflected in 
markedly smaller and more diffuse bubbles across FR and PG, but it is a \textit{misleading} 
signal of competence: the reduction in observable tool errors does not imply better 
performance, but rather that models frequently bypass grounding altogether, generating 
narrowly scoped or impractical recommendations that satisfy surface-level fluency while 
ignoring environmental constraints that a thorough tool-use strategy would have surfaced. 
The result is an inverse relationship between error visibility and failure severity across 
the two task families, \textit{reactive tasks expose failures loudly through execution breakdowns, 
while proactive tasks conceal them quietly through under-grounded generation}.

\paragraph{Formatting errors.}
\texttt{UPSTREAM\_FORMAT\_ERROR} and \texttt{EMPTY\_OUTPUT} remain comparatively infrequent across all settings. However, we observe a non-trivial dependence on model scale and behavior. Smaller models are disproportionately affected by \texttt{MISSING\_ARGUMENTS}, \texttt{UNEXPECTED\_ARGUMENTS}, and \texttt{EXECUTION\_ERROR}, often exhausting their interaction budget attempting to construct syntactically valid and semantically consistent tool calls. In contrast, mid-sized models (20--30B) exhibit a distinct failure mode: they more frequently produce malformed outputs that violate strict schema constraints, such as generating additional reasoning text outside the expected JSON structure for tool calls. 

These formatting violations are further amplified in multi-turn settings, where failed outputs are fed back into subsequent context. This can lead models to incorrectly assume task completion, prematurely switch tools, or propagate earlier mistakes through compounding errors. Notably, such issues are far less prevalent in larger, high-reasoning models, which demonstrate stronger adherence to output schemas and more stable multi-step planning.

We hypothesize that this behavior in mid-sized models stems from a combination of partial reasoning capability and overconfidence: unlike smaller models that fail early and explicitly, these models attempt more complex reasoning but lack robustness in structured generation. While it is possible to mitigate a subset of these errors through carefully engineered parsers or post-processing pipelines, our evaluation intentionally reflects \emph{out-of-the-box} performance, without specialized parsing or correction mechanisms, in order to more faithfully capture real-world deployment constraints.

\paragraph{Model Scaling Trends.}
A clear trend emerges with increasing model scale: larger models tend to reduce the frequency of argument-level and tool-selection errors, but still exhibit non-trivial execution failures. This suggests that while scaling improves reasoning and planning fidelity, it does not fully resolve issues in tool interaction and execution robustness. Consequently, future improvements may require better training signals or architectures specifically targeting tool-use reliability rather than relying solely on scale.
Overall, models struggle in \textbf{multi-hop tool usage}, especially when tools are abstracted for prompt efficiency (e.g., API-based retrieval instead of inline context). Parameter-related failures remain dominant, while increasing the number of tools often leads to inefficient exploration, redundant calls, and higher error rates. Conversely, fewer but more expressive tools reduce both cognitive load and misuse.

We notice that Gemini 2.5 Pro follows similar tool usage error trends for multimodal setting as well. 

\textbf{Implications for tool design:}
\begin{itemize}
    \item Prefer \textbf{simpler interfaces} with fewer arguments and richer outputs to reduce call complexity and parameter errors.
    \item Avoid \textbf{ambiguous or similar function names} (e.g., \texttt{get\_event\_captions} vs.\ \texttt{get\_captions}).
    \item Include \textbf{explicit error signals} in outputs to support implicit self-correction, which larger models utilize effectively.
\end{itemize}

\subsection{Definitions}
\label{app:additional_metrics_definitions}
For \textsc{IG} and \textsc{CF}, correctness is first assessed using string-level exact match (Eq.~\ref{eq:string_match}). If an exact match fails, we employ \textsc{Gemini 2.5 Pro} as a semantic equivalence judge (Eq.~\ref{eq:llm_as_judge}) to determine whether the predicted answer is meaningfully consistent with the ground truth. This hybrid decision rule is formalized in Eq.~\ref{eq:ig_cf_correct}.

For \textsc{CF}, performance is reported as the macro-average over counterfactual instances where the answer remains unchanged and those where the answer changes, as defined in Eq.~\ref{eq:cf_macro}.

For \textsc{MC}, accuracy is computed via exact match between the selected option (or its textual form) and the ground truth choice, as shown in Eq.~\ref{eq:mc_acc}. Since all valid answers are explicitly enumerated, semantic judging is not applied for this task.

For \textsc{FR}, we formulate the task as a ranking problem and report Mean Average Precision at rank $k$ (MAP@$k$), which measures how highly relevant features are ranked within the recommendation list. MAP@$k$ computation is given in Eq.~\ref{eq:map_at_k}.

We additionally report \emph{Average Relevant Rank Sum (RRS)}, a ranking error metric that quantifies how far relevant items are from their ideal positions. For each instance, the absolute deviation between the predicted rank of each relevant item and its ideal rank (1–3) is summed, and the final score is obtained by averaging across instances (Eq.~\ref{eq:rrs_instance}–\ref{eq:rrs}). Lower values indicate better ranking quality, with zero corresponding to perfect ordering. This metric provides complementary insight beyond top-$k$ accuracy by capturing the magnitude of mis-ranking among relevant options.

For \textsc{PG}, we use two complementary metrics. First, subjective plan quality is assessed using the Role Playing Judge (RPJ) framework. Second, we measure \emph{plan validity}, defined as the proportion of sub-program steps that are directly executable with correct tool or feature names and well-formed arguments, as formalized in Eq.~\ref{eq:plan_validity}.

Together, these metrics capture correctness for deterministic reactive tasks, ranking quality for proactive recommendation, and both perceived usefulness and practical executability for plan generation.
\begin{table}[h]
\centering
\caption{Summary of evaluation metrics used for each task and their corresponding equation definitions.}
\label{tab:metric_summary}
\begin{tabular}{l l l}
\hline
\textbf{Task} & \textbf{Metric} & \textbf{Equation} \\
\hline
Information Grounding (\textsc{IG}) 
& Hybrid Exact/Semantic Accuracy 
& Eq.~\ref{eq:ig_cf_correct} \\

Counterfactual Reasoning (\textsc{CF}) 
& Hybrid Exact/Semantic Accuracy 
& Eq.~\ref{eq:ig_cf_correct} \\

Counterfactual Reasoning (\textsc{CF}) 
& Macro-Average Accuracy 
& Eq.~\ref{eq:cf_macro} \\

Multiple Choice (\textsc{MC}) 
& Exact-Match Accuracy 
& Eq.~\ref{eq:mc_acc} \\

Feature Recommendation (\textsc{FR}) 
& MAP@$k$ 
& Eq.~\ref{eq:map_at_k} \\

Feature Recommendation (\textsc{FR}) 
& Average Relevant Rank Sum 
& Eq.~\ref{eq:rrs} \\

Plan Generation (\textsc{PG}) 
& RPJ Score 
& Algorithm~\ref{alg:role_playing_judge} \\

Plan Generation (\textsc{PG}) 
& Plan Validity 
& Eq.~\ref{eq:plan_validity} \\
\hline
\end{tabular}
\end{table}

\newcommand{\EM}{\mathrm{EM}}
\newcommand{\Sem}{\mathrm{Sem}}
\newcommand{\Judge}{\mathrm{J}}

\begin{equation}
\label{eq:string_match}
\EM(\hat{y}_i, y_i) \;=\; \mathbb{I}\!\left[\mathrm{norm}(\hat{y}_i)=\mathrm{norm}(y_i)\right],
\end{equation}

\begin{equation}
\label{eq:llm_as_judge}
\Judge(\hat{y}_i, y_i) \;=\; \Judge_{\textsc{Gemini 2.5 Pro}}(\hat{y}_i, y_i) \in \{0,1\},
\end{equation}

\begin{equation}
\label{eq:ig_cf_correct}
c_i^{(\textsc{IG}/\textsc{CF})}
\;=\;
\max\!\big(\EM(\hat{y}_i, y_i), \Judge(\hat{y}_i, y_i)\big),
\qquad
\mathrm{ACC}_{\textsc{IG}/\textsc{CF}}
\;=\;
\frac{1}{N}\sum_{i=1}^{N} c_i^{(\textsc{IG}/\textsc{CF})}.
\end{equation}

\begin{equation}
\label{eq:cf_macro}
\mathrm{ACC}_{\textsc{CF}}^{\textsc{macro}}
\;=\;
\frac{1}{2}\Big(\mathrm{ACC}_{\textsc{CF}_{\textsc{changed}}}
\;+\;
\mathrm{ACC}_{\textsc{CF}_{\textsc{unchanged}}}\Big),
\end{equation}

\begin{equation}
\label{eq:cf_acc_per_type}
\mathrm{ACC}_{\textsc{CF}_{s}}
\;=\;
\frac{1}{|S_s|}\sum_{i\in S_s} c_i^{(\textsc{CF})},
\qquad
s\in\{\textsc{changed},\textsc{unchanged}\}.
\end{equation}

\begin{equation}
\label{eq:mc_acc}
c_i^{(\textsc{MC})}
\;=\;
\mathbb{I}\!\left[\mathrm{norm}(\hat{z}_i)=\mathrm{norm}(z_i)\right],
\qquad
\mathrm{ACC}_{\textsc{MC}}
\;=\;
\frac{1}{N}\sum_{i=1}^{N} c_i^{(\textsc{MC})}.
\end{equation}

\begin{equation}
P_i(r)
\;=\;
\frac{1}{r}\sum_{t=1}^{r}\mathrm{rel}_i(t),
\end{equation}

\begin{equation}
\mathrm{AP@}k_i
\;=\;
\frac{1}{\min(k, R_i)}\sum_{r=1}^{k} P_i(r)\,\mathrm{rel}_i(r),
\end{equation}

\begin{equation}
\label{eq:map_at_k}
\mathrm{MAP@}k
\;=\;
\frac{1}{N}\sum_{i=1}^{N} \mathrm{AP@}k_i,
\end{equation}

\begin{equation}
v_{i,j}
\;=\;
\mathbb{I}\!\left[\text{step }(i,j)\text{ uses a valid tool/feature name and well-formed arguments}\right],
\end{equation}

\begin{equation}
\label{eq:plan_validity}
\mathrm{Validity}_i
\;=\;
\frac{1}{J_i}\sum_{j=1}^{J_i} v_{i,j},
\qquad
\mathrm{Validity}
\;=\;
\frac{1}{N}\sum_{i=1}^{N}\mathrm{Validity}_i.
\end{equation}

\paragraph{Per-instance Relevant Rank Deviation.}
Let $N$ denote the number of evaluation instances and $K=3$ the number of relevant items per instance. 
For instance $i$, let $r_{i,j}$ denote the predicted rank of the $j$-th relevant item, where the ideal ranks are $j \in \{1,2,3\}$.

\begin{equation}
D_i = \sum_{j=1}^{K} \left| r_{i,j} - j \right|
\label{eq:rrs_instance}
\end{equation}

\begin{equation}
\text{RRS} = \frac{1}{N} \sum_{i=1}^{N} D_i
\label{eq:rrs}
\end{equation}
\mainlinks

\subsection{Reactive Tasks}
\label{app:reactive_additional_metrics}
\begin{figure*}[ht]
  \vskip 0.2in
  \begin{center}
    \centerline{\includegraphics[width=0.9\linewidth]{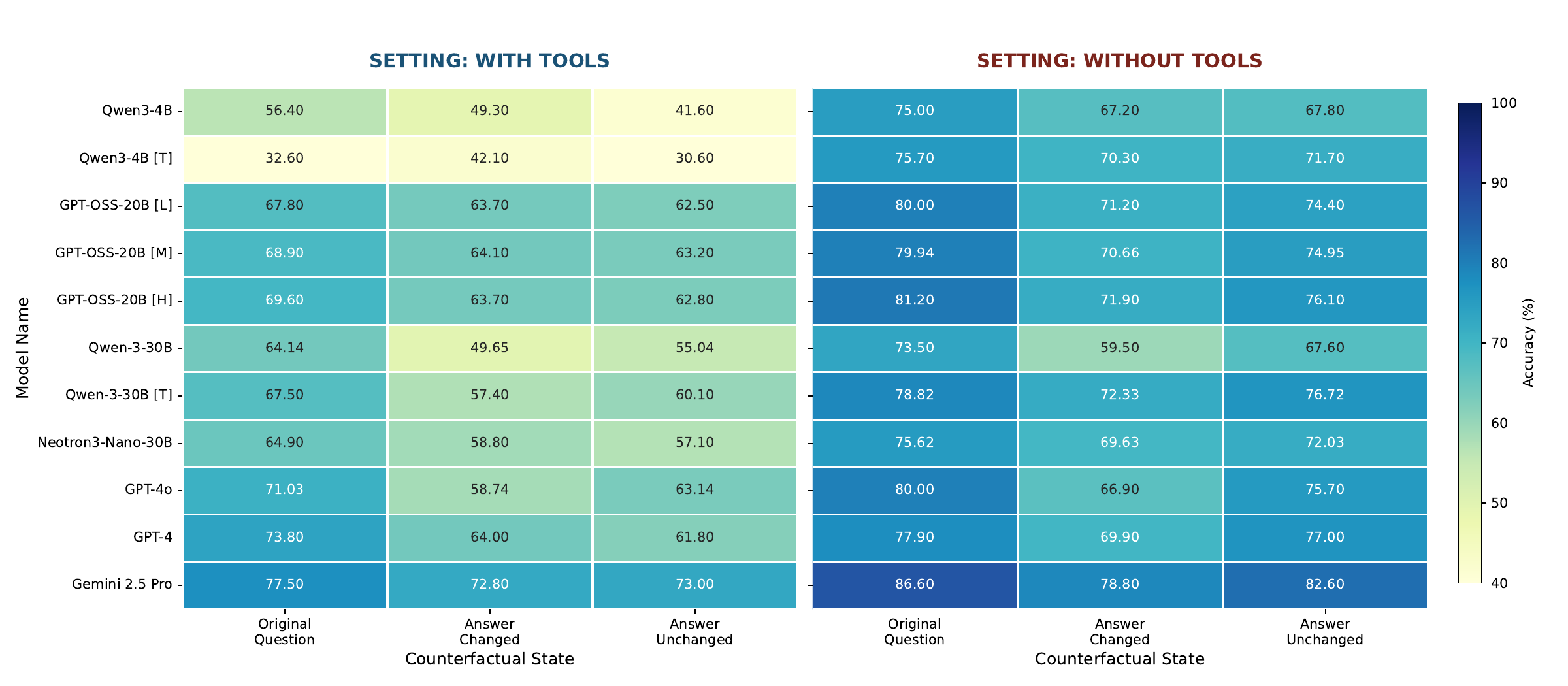}}
    \caption{
     Counterfactual Reasoning (CF) performance across models under settings with and without tool access. The heatmap reports accuracy for original questions, counterfactual queries where the correct answer changes, and counterfactual queries where the answer remains unchanged.
    }
    \label{fig:cf_heatmaps}
  \end{center}
\end{figure*}

\textbf{\textsc{Counterfactual Reasoning}}
Figure \ref{fig:cf_heatmaps} compares \textsc{CF} performance under \textit{Sole Reasoning} (right) and agentic tool usage (left). Without tools, most models achieve relatively high accuracy on original questions and maintain comparable performance on both answer-changed and answer-unchanged counterfactuals, indicating that many models can internally simulate hypothetical perturbations when full context is directly provided. Performance is consistently strongest for large proprietary models, while medium and small models exhibit modestly lower but stable behavior across counterfactual types.
In contrast, introducing tools produces a uniform degradation across models and counterfactual states. The drop is not confined to cases where the answer changes, but also affects answer-unchanged queries, suggesting that failures arise primarily from imperfect retrieval and tool interaction rather than counterfactual reasoning itself. Several models exhibit larger degradation on the unchanged condition, indicating that even preserving known facts becomes fragile once information must be re-acquired through tools.

Overall, the comparison highlights a clear separation between reasoning competence and interaction competence: models retain non-trivial counterfactual reasoning ability in isolation, but struggle to express it reliably in agentic settings where correct tool use is required. This reinforces that effective information acquisition and state grounding, rather than hypothetical reasoning alone, is the dominant bottleneck for counterfactual performance in personalized home environments.

\textbf{Impact of Personalization and PersonalHomeTools.}
Tables \ref{tab:no_pers_w_tools} and \ref{tab:no_pers_no_tools} show that personalization has markedly different effects under \textit{Sole Reasoning} versus agentic evaluation. When complete household context is directly available (Table \ref{tab:no_pers_no_tools}), personalization generally leads to substantial accuracy gains across \textsc{IG}, \textsc{CF}, and \textsc{MC}, indicating that models can effectively exploit user-specific information when retrieval is trivial. In contrast, under agentic evaluation (Table \ref{tab:no_pers_w_tools}), personalization yields smaller and more mixed changes, with modest improvements in some cases and slight regressions in others.
Importantly, the relative stability of performance in the agentic setting reflects the personalization-centric design of \textsc{PersonalHomeTools}, which enables structured retrieval of user-specific context and mitigates the difficulty of reasoning directly over raw personalized information.

\begin{figure*}[ht]
  \vskip 0.2in
  \begin{center}
    \centerline{\includegraphics[width=\linewidth]{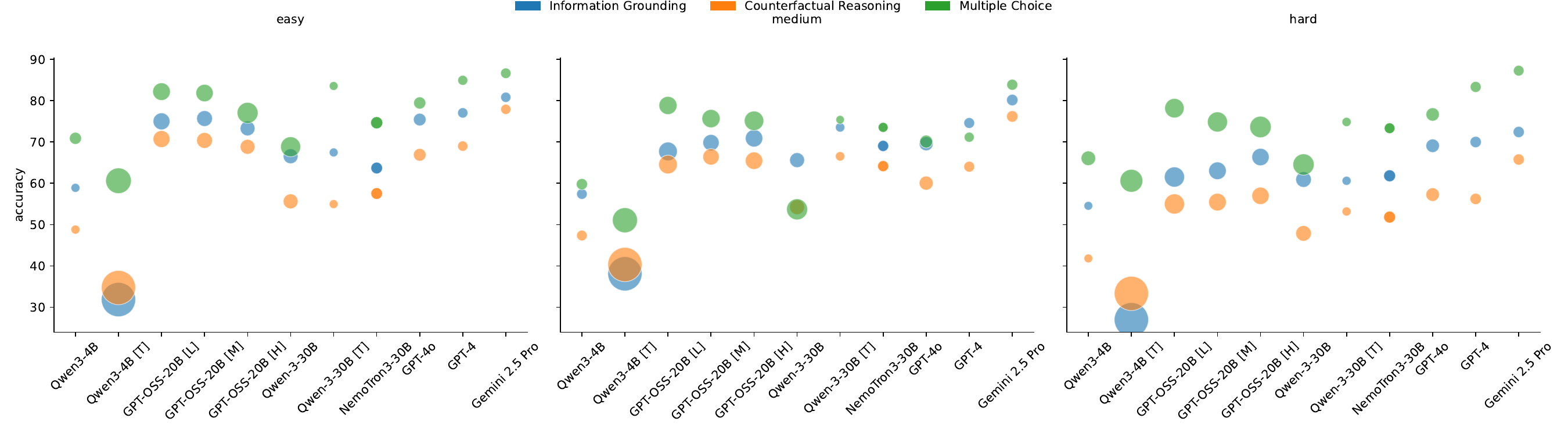}}
    \caption{
      Accuracy across models for three reactive task categories, Information Grounding (IG), Counterfactual Reasoning (CF), and Multiple-Choice (MCQ), evaluated at easy, medium, and hard difficulty levels. Each point corresponds to a model–task pair; marker size reflects the average number of turns taken to complete the task.
    }
    \label{fig:diff_analysis}
  \end{center}
\end{figure*}

\textbf{Effect of Task Difficulty on Reactive Task Performance.}
Figure \ref{fig:diff_analysis} illustrates consistent and interpretable scaling behavior across task difficulty for all three reactive tasks. 
Accuracy decreases monotonically from easy to medium to hard settings, confirming that the difficulty stratification meaningfully captures increasing reasoning complexity. \textsc{IG} and \textsc{MC} remain comparatively more stable than \textsc{CF} across all difficulty levels, while \textsc{CF} exhibits the steepest degradation, reinforcing its role as the most challenging reactive task. 
Larger and proprietary models generally achieve higher accuracy across difficulties, but no model is immune to performance drops on hard instances, indicating that complex household-level reasoning remains challenging even at scale. 
Marker sizes further show that harder instances tend to require more interaction turns, suggesting that models attempt deeper tool-based exploration or longer reasoning chains without reliably closing the performance gap.
Overall, the figure highlights that increasing difficulty exposes systematic weaknesses in multi-hop and counterfactual reasoning, and that additional interaction alone is insufficient to overcome these challenges.

\begin{table}[ht]
\caption{Effect of personalization under agentic evaluation, reporting accuracy for Generic and Personalized settings on \textsc{IG}, \textsc{CF}, and \textsc{MC}}
\begin{center}
\resizebox{\linewidth}{!}{%
\begin{tabular}{lcccccc}
\hline
 & \multicolumn{2}{c}{\textbf{IG (ACC \%)}} 
 & \multicolumn{2}{c}{\textbf{CF (ACC \%)}}
 & \multicolumn{2}{c}{\textbf{MC (ACC \%)}} \\
\cline{2-7}
\textbf{Model}
& Generic & Personalized 
& Generic & Personalized
& Generic & Personalized\\
\hline

GPT-OSS-20B [H]     &65.77&69.60&61.83&63.25&71.00&75.20 \\
GPT-4               &74.33&73.80&63.54&62.90&77.40&79.20 \\
Gemini 2.5 Pro      &76.90&77.50&71.80&72.90&87.90&85.80 \\
\hline
\end{tabular}%
}
\end{center}
\label{tab:no_pers_w_tools}
\end{table}

\begin{table}[ht]
\caption{Effect of personalization under \textit{Sole Reasoning}, reporting accuracy for Generic and Personalized settings on \textsc{IG}, \textsc{CF}, and \textsc{MC}.}
\begin{center}
\resizebox{\linewidth}{!}{%
\begin{tabular}{lcccccc}
\hline
 & \multicolumn{2}{c}{\textbf{IG (ACC \%)}} 
 & \multicolumn{2}{c}{\textbf{CF (ACC \%)}}
 & \multicolumn{2}{c}{\textbf{MC (ACC \%)}} \\
\cline{2-7}
\textbf{Model}
& Generic & Personalized 
& Generic & Personalized
& Generic & Personalized\\
\hline

GPT-OSS-20B [H]     &99.54&81.20&99.54&77.00&88.20&89.20 \\
GPT-4              &99.78&77.90&99.78&73.45&83.90&86.30 \\
Gemini 2.5 Pro      &99.80&86.60&99.80&80.70&94.10&80.70 \\
\hline
\end{tabular}%
}
\end{center}
\label{tab:no_pers_no_tools}
\end{table}


\mainlinks
\subsection{Proactive Tasks}
\label{app:proactive_additional_metrics}
\begin{figure}[ht]
  \vskip 0.2in
  \begin{center}
    \centerline{\includegraphics[width=0.95\columnwidth]{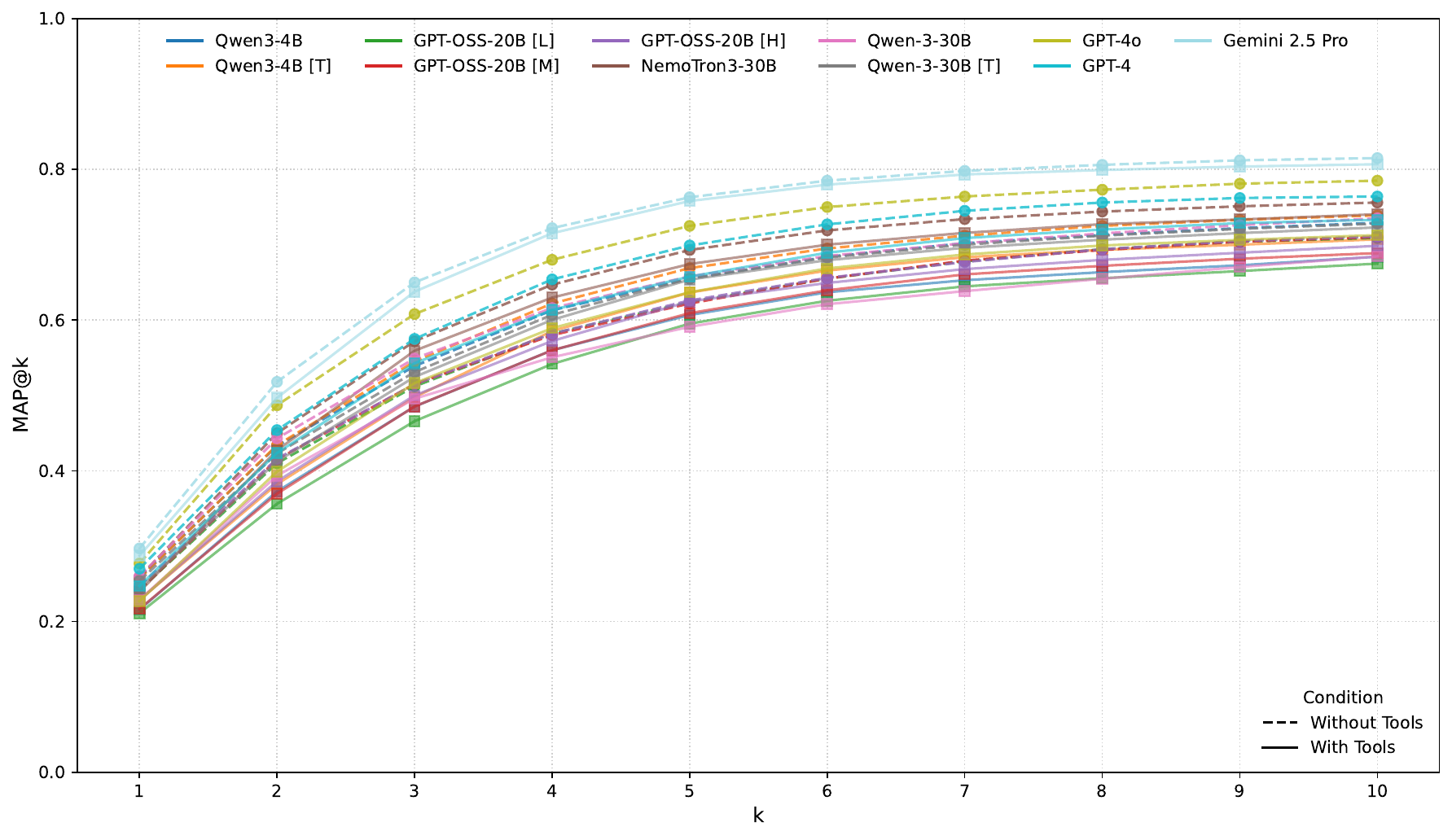}}
    \caption{
      Mean Average Precision at rank $k$ ($MAP@k$) across models evaluated with and without tool access. Solid lines denote performance with tools, while dashed lines correspond to settings without tools. Results are reported for varying values of k, illustrating ranking performance trends across evaluation conditions.
    }
    \label{fig:map_at_k}
  \end{center}
\end{figure}

\textbf{\textsc{Feature Recommendation}}. Figure~\ref{fig:map_at_k} shows consistent improvements in MAP@k as k increases across all models, with performance gains tapering off beyond moderate values of k. 
Across the board, agentic settings with tool access underperform their Sole Reasoning counterparts, reinforcing that effective interaction with tools remains a key challenge even for strong foundation models. 
Larger and more capable models achieve higher absolute MAP@k, but the relative gap between with-tools and without-tools settings persists, indicating that scale alone does not resolve interaction-level errors. 
Notably, \textsc{Gemini 2.5 Pro} and \textsc{GPT-4} class models define the upper envelope of performance, yet still exhibit nontrivial degradation under tool use, highlighting substantial headroom for improving agentic reasoning and decision-making in structured environments.

Table \ref{tab:fr_relevant_rank_sum} reports Average Relevant Rank Sum (RRS) for \textsc{Feature Recommendation} under text-based evaluation. Under \textit{Sole Reasoning}, frontier models achieve the lowest RRS values, with \textsc{Gemini 2.5 Pro} and \textsc{GPT-4o} performing best, indicating that relevant features are placed closer to their ideal ranks. Medium-scale open models exhibit moderately higher RRS, while smaller models show the largest deviations, reflecting noisier ranking behavior. When tools are introduced, RRS increases substantially for nearly all models, demonstrating that tool-mediated interaction degrades fine-grained ranking quality even when top-1 accuracy may remain reasonable. Notably, the relative ordering of models remains broadly consistent across settings, suggesting that the degradation primarily stems from interaction complexity rather than a change in underlying ranking capability. Overall, these results reinforce that precise preference ordering remains challenging in agentic contexts, and that effective use of tools is critical for maintaining ranking fidelity.

\begin{table}[th]
\centering
\caption{Relevant Rank Sum for the \textsc{Feature Recommendation} task on text-based data. Lower values indicate better ranking quality.}
\label{tab:fr_relevant_rank_sum}
\begin{tabular}{lcc}
\hline
\textbf{Model} & \textbf{No Tools} & \textbf{With Tools} \\
\hline
Qwen3-4B            & 3.4730 & 10.7620 \\
Qwen3-4B [T]        & 3.3740 & 10.4270 \\
GPT-OSS-20B [L]     & 3.5740 & 17.3260 \\
GPT-OSS-20B [M]     & 3.5850 & 15.0270 \\
GPT-OSS-20B [H]     & 3.5930 & 19.5160 \\
Qwen3-30B          & 3.4910 & 21.8080 \\
Qwen3-30B [T]      & 3.4500 & 10.3440 \\
Neotron3-Nano-30B   & 3.2580 & 9.0988 \\
\colorbox{proprietary}{GPT-4o} & 3.1340 & 10.5030 \\
\colorbox{proprietary}{GPT-4}  & 3.2100 & 10.2050 \\
\colorbox{proprietary}{Gemini 2.5 Pro} & \textbf{2.9780} & \textbf{9.0090} \\
\hline
\end{tabular}
\end{table}

\begin{table}[th]
\caption{Feature recommendation and plan generation validity across models.}
\begin{center}
\resizebox{\linewidth}{!}{%
\begin{tabular}{lcccc}
\hline
\textbf{Model}
& \multicolumn{2}{c}{\textbf{Feature Reco. (MAP@1)}}
& \multicolumn{2}{c}{\textbf{Plan Gen. (Validity)}} \\
\cline{2-5}
& No Tools & With Tools
& No Tools & With Tools \\
\hline

Qwen3-4B
& 0.258$\pm$0.139 & 0.216$\pm$0.159
& 0.664$\pm$0.341 & 0.001$\pm$0.015 \\

Qwen3-4B [T]
& 0.257$\pm$0.140 & 0.229$\pm$0.154
& 0.773$\pm$0.324 & 0.006$\pm$0.072 \\

GPT-OSS-20B [L]
& 0.242$\pm$0.148 & 0.211$\pm$0.160
& 0.753$\pm$0.404 & 0.343$\pm$0.464 \\

GPT-OSS-20B [M]
& 0.242$\pm$0.149 & 0.217$\pm$0.159
& 0.725$\pm$0.412 & 0.415$\pm$0.481 \\

GPT-OSS-20B [H]
& 0.247$\pm$0.146 & 0.229$\pm$0.155
& 0.699$\pm$0.428 & 0.414$\pm$0.483 \\

Qwen3-30B
& 0.260$\pm$0.138 & 0.229$\pm$0.154
& 0.645$\pm$0.302 & 0.002$\pm$0.024 \\

Qwen3-30B [T]
& 0.254$\pm$0.142 & 0.244$\pm$0.148
& 0.666$\pm$0.467 & 0.044$\pm$0.205 \\

NemoTron3-30B
& 0.259$\pm$0.138 & 0.239$\pm$0.149
& 0.825$\pm$0.320 & 0.010$\pm$0.090 \\

\colorbox{proprietary}{GPT-4o}
& 0.277$\pm$0.125 & 0.227$\pm$0.155
& 0.836$\pm$0.215 & 0.023$\pm$0.117 \\

\colorbox{proprietary}{GPT-4}
& 0.270$\pm$0.131 & 0.247$\pm$0.146
& 0.869$\pm$0.208 & 0.022$\pm$0.123 \\

\hdashline
\colorbox{proprietary}{Gemini 2.5 Pro}
& 0.297$\pm$0.104 & 0.287$\pm$0.115
& 0.795$\pm$0.238 & 0.030$\pm$0.157 \\

\hdashline
\colorbox{proprietary}{Gemini 2.5 Pro (Multimodal)}
& 0.196$\pm$0.164 & 0.163$\pm$0.166
& 0.697$\pm$0.297 & 0.405$\pm$0.423 \\

\hline
\end{tabular}
}
\end{center}
\label{tab:std_table}
\end{table}
\clearpage

\section{Models and Hyper-Parameters}
\label{app:models}
Open models are hosted via vLLM \cite{kwon2023efficient} on a multi-GPU node (4x GPUs, 96GB VRAM) utilizing a fixed tensor\_parallel\_size of 4. To ensure system stability and accommodate the overhead of long-context reasoning, the GPU memory utilization is lowered to 0.85 to prevent out-of-memory errors during the KV cache allocation. 

For models operating in explicit thinking or reasoning modes, the maximum reasoning token budgets are explicitly constrained based on model scale. It is capped for \textsc{Qwen3-4B [T]} and \textsc{Qwen3-30B [T]}, while \textsc{GPT-OSS-20B} and \textsc{Nemotron3-Nano-30B-A3B} utilize the default configuration. The \textsc{Qwen3} reasoning models utilize \textsc{DeepSeek-R1} as the reasoning parser and \textsc{} utilize its own Nano v3 parser within the vLLM serving environment. 

\begin{table}[h]
\centering
\caption{Reasoning Token Configuration by Model}
\label{tab:model_token_budgets}
\begin{tabular}{lrr}
\toprule
\textbf{Model Name} & \textbf{Reasoning Token} & \textbf{GPU Mem Utilization} \\
\midrule
Qwen3-4B & 262,144 & 0.90 \\
Qwen3-4B [T] &  78,720 & 0.90 \\
GPT-OSS-20B [L,M,H] & 131,072 & 0.90 \\
Qwen3-30B & 32,768 & 0.85 \\
Qwen3-30B [T] & 32,768 & 0.85 \\
Nemotron3-Nano-30B-A3B & 131,072 & 0.85 \\
\bottomrule
\end{tabular}
\end{table}

\mainlinks

\section{Schemas}
\label{app:schemas}
\paragraph{Household Schema.}
\begin{tcolorbox}[
  breakable,
  colback=gray!5,
  colframe=blue!20,
  coltitle = blue!20!black, 
  title=Household,  
]
\begin{verbatim}
  household_id: String 
  household_size: Integer 
  occupancy_type: String 
  layout: String 
  pets: [{ 
        name: String 
        animal: String 
        age: Integer 
  }] 
  members: [{
        member_id: String
        name: String
        role: String
        age: Integer
        gender: String
        occupation: String
        persona: String
        hobbies: [String]
        lifestyle: String
        preference: String
        major_events: [{
            name: String
            animal: String
            age: Integer
        }]
  }]
  appliances: [{
        appliance: ApplianceObject
        location: String
  }]
  sensors: [{
        sensor: SensorObject
        location: String
  }]
  memories: [{
        memory: String
        subjects: [String]
        additional_information: String
        time: String
  }]
  context: {
        events: [{
                description: String
                sensors: [{
                    sensor: SensorObject
                    location: String
                }]
                appliances: [{
                    appliance: ApplianceObject
                    location: String
                }]
        }]
        weather: {
                outdoor: {
                    rain: String
                    snow: String
                    sunny: String
                    temperature: String
                    wind: String
                }
                indoor: {
                    dust: String
                    humidity: String
                    odor: String
                    temperature: String
                }
        }
  }
\end{verbatim}
\end{tcolorbox}

\paragraph{Video Transcript Schema.}
\begin{tcolorbox}[
  breakable,
  colback=gray!5,
  colframe=blue!20,
  coltitle = blue!20!black, 
  title=Video Transcript,  
]
\begin{verbatim}
  datapoint_id: String 
  video_path: String
  video_description: {
    timebase: {
        timestamp_format: DateTime
        start_offset_seconds: Integer
    }
    scene: {
        summary: String
        environment_type: String
        location_label: String
        lighting: {
            overall: String
            sources_visible: [String]
        }
        camera: {
            motion: String
            viewpoint: String
            notes: String
        }
        layout: {
            zones: [{
                zone_id: String
                label: String
                relative_position: String
            }]
            landmarks: [{
                landmark_id: String
                type: String
                description: String
                zone_id: String
            }]
        }
    }
    entities: {
        members: [{
            member_id: String
            appearance: String
            initial_location: String
        }]
        pets: [{
            pet_id: String
            species: String
            initial_location: String
            description: String
        }]
        appliances: [{
            appliance_id: String
            type: String
            location: String
            initial_state: {}
            description: String
        }]
        objects: [{
            object_id: String
            label: String
            category: String
            initial_location: String
            initial_state: String
            description: String
        }]
        events: [{
            event_id: String
            ts_start: String
            ts_end: String
            event_type: String
            actors: [String]
            targets: [String]
            description: String
            audio: String
        }]
        entity_state_timelines: {
            members: [{
                member_id: String
                states: [{
                    ts: String
                    state: String
                    trigger_event_ids: [String]
                }]
            }]
            pets: [{
                pet_id: String
                states: [{
                    ts: String
                    state: String
                    trigger_event_ids: [String]
                }]
            }]
            appliances: [{
                appliance_id: String
                states: [{
                    ts: String
                    state: String
                    trigger_event_ids: [String]
                }]
            }]
            objects: [{
                object_id: String
                states: [{
                    ts: String
                    state: String
                    trigger_event_ids: [String]
                }]
            }]
        }
    }
    captions: {
        global: String
        segments: [{
            ts_start: String
            ts_end: String
            caption: String
        }]
        events: [{
            event_id: String
            caption: String
        }]
    }
    quality: {
        notes: String
        consistency_checks: {
            entity_ids_consistent: Boolean
            all_state_changes_timestamped: Boolean
            no_unverifiable_inferences: Boolean
        }
    }
  }

\end{verbatim}
\end{tcolorbox}
\clearpage

\clearpage

\mainlinks
\section{Prompts}
\label{app:prompts}

\subsection{Household Member Persona Generation}
\label{app:hh_generation_prompts}

\begin{tcolorbox}[
  breakable,
  colback=gray!5,
  colframe=gray!40,
  coltitle = gray!20!black, 
  title=Household Member Persona Generation Prompt Template,  
]
You are an expert persona generator. Your job is to create realistic, coherent individual personas based on household-level information and sample personas. Follow these rules precisely: \\

1. Use the Household Persona as the Anchor
\begin{itemize}
\item Use the household-level description to infer: demographics, living situation, socioeconomic context, lifestyle preferences, environment and context.
\item All individual personas must align with this household information.
\end{itemize}

2. Respect the Occupancy Type
\begin{itemize}
\item If occupancy\_type = roommate:
    \begin{itemize}
        \item All individuals should be unrelated.
        \item Ages should fall within a similar age range, unless the household persona describes otherwise.
        \item Income levels, occupations, and lifestyles may differ, but should still plausibly co-exist in a shared-living situation. 
        \item Personalities should not be identical—make each person distinct yet compatible as cohabitants.
    \end{itemize}
\item Household size = 2 
    \begin{itemize}
        \item Could be a couple (romantic partners). 
        \item Or a parent + one child. 
    \end{itemize}
\item Household size = 3 
    \begin{itemize}
        \item Could be a couple + one child. 
        \item Or a single parent + two children.
    \end{itemize}
\item Household size = 4 or 5
    \begin{itemize}
    \item Typical family compositions include:
        \begin{itemize}
        \item Two parents + children
        \item Single parent + multiple children
        \item Multi-generational families 
        \end{itemize}
    \end{itemize}
\end{itemize}

3. Use the Samples as Style Guidance 
\begin{itemize}
    \item Use samples only as examples of tone, structure, and level of detail. 
    \item Do not copy content. 
    \item Maintain similar attributes.
\end{itemize}

4. Produce a Concise, Self-Contained Persona for Each Individual
\begin{itemize}
    \item For each individual, provide:
    \begin{itemize}
        \item member\_id: must be \{household\_id\}\_\{n\} where n starts at 1
        \item name: extract name from persona description
        \item role: parent, child, roommate, partner, etc.
        \item age (estimate if not explicitly stated)
        \item gender (Male or Female)
        \item occupation
        \item persona: 1-2 sentences summarizing personality and character
        \item hobby: a list of hobby and interest keywords
        \item lifestyle: habits, routines, energy usage tendencies (e.g., heavy cooking, night owl, active, tech-savvy)
        \item preference: any preferences relevant for simulation (e.g., temperature settings, appliance usage patterns, quiet mornings)
        \item major\_event: list of significant life events with date and description. This field can be left empty if no major events are known.
    \end{itemize}
\end{itemize}

5. Output Format \\
Return the result as a list of personas, one entry per individual, labeled:\\
\begin{verbatim}
"members": [
  {
    "member_id": "{household_id}_1",
    "name": "",
    "role": "",
    "age": ,
    "gender": "",
    "occupation": "",
    "persona": "",
    "hobbies": ["", ""],
    "lifestyle": "",
    "preference": "",
    "major_event": [
      {
        "date": "",
        "description": ""
      }
    ]
  },
  {
    "member_id": "{household_id}_2",
    "name": "",
    "role": "",
    "age": ,
    "gender": "",
    "occupation": "",
    "persona": "",
    "hobbies": ["", ""],
    "lifestyle": "",
    "preference": "",
    "major_event": [
      {
        "date": "",
        "description": ""
      }
    ]
  }
  ]...
\end{verbatim}
\end{tcolorbox}

\subsection{Task Generation}
\label{app:task_generation}

\paragraph{Reactive Tasks}
\begin{tcolorbox}[
  breakable,
  colback=gray!5,
  colframe=gray!40,
  coltitle = gray!20!black, 
  title=Info Grounding and Multiple Choice Question Task Generation Prompt Template,  
]
Prompt: Dictionary-Driven Question–Answer Generation with Programs\\

\textbf{Role}\\
You are a data QA generation engine.\\
Your task is to generate accurate, verifiable question–answer pairs from a given dictionary with many keys and values (including nested objects, lists, IDs, names, attributes, states, and memories).\\

All outputs must be machine-gradable, program-grounded, and strictly derived from the input dictionary.\\

\textbf{Input}\\
You will be given:
\begin{itemize}
    \item A dictionary with many keys and values
    \item IDs that may map to entities (for example, people, devices, rooms)
    \item Optional human-readable names associated with IDs
    \item Structured attributes, lists, timestamps, states, and/or memories
\end{itemize}

\textbf{Required Output Structure}\\
For each generated item, include all sections below.

\begin{itemize}
    \item 1. Questions - For every item, produce two versions of the same question:
    \begin{itemize}
        \item A. Question (ID-based)
        \begin{itemize}
            \item Uses IDs or raw dictionary keys
            \item Example: What is the favorite activity of member\_id\_001 and member\_id\_002?
        \end{itemize}
        \item B. Personalized Question (Name-based)
        \begin{itemize}
            \item Replaces IDs with names when available
            \item Semantically identical to the ID-based question
            \item Example: What is the favorite activity of David and Susan?
        \end{itemize}
        \item Both questions must resolve to the exact same answer.
    \end{itemize}

    \item 2. Answer
    \begin{itemize}
        \item The answer must be exact and specific
        \item It does not need to be a full sentence
        \item Output the answer as a list of acceptable variants
        \item All variants must be semantically equivalent
        \item The answer must be directly grounded in the input dictionary
        \item For every question–answer item, you must also generate a `multiple\_choices` field.
        \begin{itemize}
            \item Incorrect answers should be hard distractors that are plausible within the dictionary and semantically close to the correct answer, but wrong in a subtle and verifiable way.
            \item Effective distractors can be created by swapping entities, using real values from the wrong context, time, or location, or by slightly modifying the correct value.
            \item Obviously fake, generic, or ungrounded options are not allowed.
            \item All incorrect answers must be mutually distinct, must not overlap with any accepted answer variants, and must match the same semantic type as the correct answer.
        \end{itemize}
    \end{itemize}

    \item 3. Difficulty Classification
    \begin{itemize}
        \item Each question must be labeled with exactly one difficulty level:
        \begin{itemize}
            \item simple: up to 3 reasoning steps
            \item medium: 3 to 5 reasoning steps
            \item hard: 5 to 10 reasoning steps
        \end{itemize}
        \item The generated set must include a mix of:
        \begin{itemize}
            \item Simple questions (direct lookups, single entity)
            \item Medium questions (comparison, aggregation, or multiple entities)
            \item Hard questions (multi-hop reasoning across multiple keys, memories, or attributes)
        \end{itemize}
        \item Rules
        \begin{itemize}
            \item All answers must be directly supported by the input dictionary
            \item Do not hallucinate or infer missing information
            \item Personalized and ID-based questions must map one-to-one
            \item Answers must be machine-checkable
            \item If multiple entities are mentioned, the answer must be correct for all of them
            \item Programs must be sufficient to compute the answer and no more
        \end{itemize}
        \item Question Naturalness Constraints (Mandatory)
        \begin{itemize}
            \item Questions must sound like something a human would realistically ask in everyday language.
            \begin{itemize}
                \item The question should not reveal how the answer is computed.
                \item The question should not mention joins, lookups, timelines, or cross-referencing.
            \end{itemize}
            \item Do NOT ask questions that require stating the same fact twice in different forms.
            \begin{itemize}
                \item Avoid circular or self-referential questions (for example, asking what happened on the same date as an event that already describes that event).
            \end{itemize}
            \item If Personalization is not possible, then personalized question should be "None"
        \end{itemize}
    \end{itemize}
\end{itemize}
\textbf{Output Format (Strict)}\\
Return a list of JSON objects only, with no additional text.
\begin{verbatim}
[
  {{
    "difficulty": "simple | medium | hard",
    "question": "ID-based question",
    "personalized_question": "Name-based question", 
    "answer_variants": ["answer_variant_1", "answer_variant_2"],
    "multiple_choices": {{
    "correct": "best_answer_variant", 
    "incorrect": [
      "hard_distractor_1",
      "hard_distractor_2",
      "hard_distractor_3",
      "hard_distractor_4"
      ] # you MUST always predict 4 incorrect hard distractors
    }}
  }}
]
\end{verbatim}
You MUST generate upto 3 examples per each difficulty level.\\

Here is the current household information:\\
\{household\_information\}

\end{tcolorbox}

\begin{tcolorbox}[
  breakable,
  colback=gray!5,
  colframe=gray!40,
  coltitle = gray!20!black, 
  title=Counterfactual Task Generation Prompt Template,  
]
You are a Household Knowledge Counterfactual Reasoning Assistant.\\

Your tasks:\\
1. Take a grounded QA pair.\\
2. Generate a counterfactual modification to the household/context/memory/appliance data.\\
3. Provide:\\
   a) The counterfactual condition\\
   b) The new derived answer under this counterfactual world\\
   c) A short explanation of the reasoning shift\\

\textbf{Input Format}\\
You will be given a single QA item in JSON format with the following structure:\\
\begin{verbatim}
{{
  "difficulty": "easy | medium | hard",
  "question": "<generic question using IDs>",
  "personalized_question": "<question using names instead of IDs>",
  "answer_variants": [
    "<canonical answer>",
    "<alternate surface forms>"
  ],
  "multiple_choices": {{
    "correct": "<correct option>",
    "incorrect": [
      "<distractor1>",
      "<distractor2>",
      "<distractor3>",
      "<distractor4>"
    ]
  }}
}}
\end{verbatim}
Notes:
\begin{itemize}
    \item The question and personalized\_question refer to the same underlying query.
    \item The answer is expected to be exact, not a full sentence.
    \item Multiple answer variants may be provided to allow robust accuracy checking.
\end{itemize}

\textbf{Counterfactual Construction Rules}
\begin{itemize}
    \item Single-Factor Modification
    \begin{itemize}
        \item Each counterfactual must modify only one primary factor.
        \item The modification must be clearly identifiable and isolated.
        \item Valid modification dimensions include:
        \begin{itemize}
            \item Time (e.g., one month ago → six months ago)
            \item Actor (switching the household member involved)
            \item Activity (e.g., cooking → studying)
            \item Appliance or sensor used
            \item Location (e.g., kitchen → bedroom)
            \item Sensor availability or malfunction
            \item Routine, habit, or motivation changes
            \item Environmental context (temperature, season, time of day)
        \end{itemize}
    \end{itemize}
    \item Explicit Causality
    \begin{itemize}
        \item The counterfactual must include a clear cause–effect relationship.
        \item If the answer changes, the modification must be the direct cause of the change.
        \item If the answer does not change, explicitly state why the modification is irrelevant.
        \item Avoid speculative, implicit, or circular reasoning.
    \end{itemize}
    \item Logical and Physical Consistency
    \begin{itemize}
        \item The counterfactual must remain internally consistent.
        \item Do not violate:
        \begin{itemize}
            \item Physical constraints (e.g., impossible appliance behavior)
            \item Temporal constraints (e.g., effects preceding causes)
            \item Household constraints (e.g., nonexistent members or appliances)
        \end{itemize}
        \item Do not introduce new entities unless strictly required by the change.
    \end{itemize}
    \item Answer Precision and Validity
    \begin{itemize}
        \item The new\_answer must be:
        \begin{itemize}
            \item One of the original answer\_variants, OR
            \item A logically updated answer with the same level of precision and granularity.
        \end{itemize}
        \item Answers must be exact values, not full sentences.
        \item Ambiguous or underspecified answers are not allowed.
    \end{itemize}
    \item Difficulty Alignment
    \begin{itemize}
        \item The reasoning complexity of the counterfactual must align with the original difficulty:
        \item Easy: Superficial or irrelevant change; answer remains unchanged.
        \item Medium: Contextual change; answer may change.
        \item Hard: Multi-hop reasoning involving memories, sensors, and inferred behavior.
    \end{itemize}
\end{itemize}
ANSWERING RULES UNDER COUNTERFACTUAL:\\
- Recompute the answer as if the counterfactual were true.\\
- Stay strictly grounded—no new made-up facts beyond the change introduced.\\
- If the counterfactual doesn’t meaningfully change the answer, say so explicitly.\\

OUTPUT FORMAT:\\
Return a JSON dictionary for each input QA pair:
\begin{verbatim}
{{
  "original_question_id": "...",
  "counterfactuals": [
    {{
      "change": "<description of the hypothetical change>",
      "new_answer": "<exact answer>",
      "answer_changed": true | false,
      "reasoning": "<concise causal explanation>",
      "<outer_key>": [
        {{
          "<field_1>": "<value>",
          "<field_2>": "<value>",
          "...": "..."
        }}
    }}
  ]
}}
\end{verbatim}
Constraints:\\
- The list counterfactuals must contain between 1 and 3 items.\\
- answer\_changed must accurately reflect whether the answer differs from the original.\\
- Reasoning must be minimal, explicit, and causal (no speculation).\\

Valid Counterfactual Examples (Illustrative Only)\\
- Temporal: “Assume the referenced memory occurred six months ago instead of one month ago.”\\
- Behavioral: “Assume the household member stopped late-night cooking.”\\
- Sensor-related: “Assume the microwave wattmeter was offline during that period.”\\
- Substitution: “Assume a hot plate was used instead of a microwave.”\\
- Irrelevant change: “Assume the indoor temperature was 30°C instead of 24°C.”\\

Quality Checklist (All Must Pass)\\
Before producing the output, verify that:\\
- The counterfactual is realistic and plausible.\\
- The reasoning clearly explains causality.\\
- The answer is unambiguous and machine-checkable.\\
- No unnecessary entities or assumptions are introduced.\\

You MUST generate at least one counterfactual where answer should change, and one where answer should not change.\\
The JSON strictly follows the required schema.\\

Here is the current household information:\\
\{household\_information\}\\

Here are all the question and answer pairs\\
\{qa\_pairs\}

\end{tcolorbox}

\paragraph{Feature Recommendation}
\begin{tcolorbox}[
  breakable,
  colback=gray!5,
  colframe=gray!40,
  coltitle = gray!20!black, 
  title=Feature Ranking Task Generation Prompt Template,  
]
You are a Household Knowledge Ranking Generator. Your job is to produce context-grounded appliance feature rankings strictly based on the supplied household data.\\
A list of appliance features will be provided.\\
From this list, you must select exactly 10 features and produce a ranked ordering (1 → 10) justified by the household’s needs, behaviors, device context, and long-term patterns.\\

Your task is to\\
1. Select exactly 10 features from the input list.\\
2. Assign a correct ranking from 1 (most relevant) to 10 (least relevant).\\
3. Rankings must reflect explicit evidence from:\\
  - Household structure (members, ages, hobbies, roles, pets)\\
  - Device information (appliance types, sensors, locations)\\
  - Contextual data (current time, weather, schedules, states)\\
  - Long-term memories (habits, major events, preferences)\\
4. Provide a short explanation for each feature, describing why it is placed in its specific ranking position.\\

RANKING REQUIREMENTS:\\
- The ranking must reflect actual priorities inferred from the household’s behavior and needs.\\
- Reasoning should show:\\
  - Observation (e.g., someone frequently forgets to turn off appliances)\\
  - Inference (e.g., a room with high traffic might benefit from automation)\\
  - Multi-step reasoning (e.g., pets + robot vacuum + user schedule patterns)
- Do not invent appliance features not in the provided list.\\
- All choices must be grounded in specific, explicit fields of the household record.\\

\textbf{API References}\\
--------------------\\
refrigerator\\
Internal cameras for remote viewing\\
This function will send a snapshot of the fridge/freezer contents to your APP. It will also give a list of detected contents.\\
Touchscreen family hub interface\\
Updates the widgets displayed on the family hub screen (e.g., calendar, notes, weather). Can also push a specific notification to the screen.\\
Ingredient management and expiration tracking\\
Gives a report of the ingredients and their meta data information such as quantity and expiration. If no specific ingredients are passed, the report will cover all ingredients. If create\_shopping\_list name is passed, it automatically generates a shopping list for ingredients that are low or about to expire\\
Recipe Recommendation\\
When given a natural language request about the recipe the user is interested for, a personalized recipe is recommended and displayed on the fridge screen.
Diagnostic alerts (e.g., door ajar)\\
Sends an alert to the user for the issue.\\
Dual ice maker (cubes and crushed/spheres)\\
Toggles the ice maker modes or triggers a rapid ice production cycle.\\
--------------------\\
oven\\
Remote preheat via app\\
The oven preheated to 350 F or the temperature specified by the user. This system also performs a diagnostic to preheating is safe. If it is unsafe, an alert is sent to the user APP about the issue.\\
Internal AI camera monitoring\\
Send a snapshot of the contents of the over to the user APP. If video mode is requested, live feed link is sent. If a specific query is passed, then the function returns an answer. When no enquiry is passed, it returns a live snapshot.\\
Cooking control compatibility\\
Change the mode of oven between convection, broil, or bake.\\
Integrated meat probe with digital readout\\
Returns the current internal temperature reading from the connected meat probe. Can set a target alert temperature.\\
...\\
-------------------- \\
speaker \\
Collect user response \\
Often used to get the last N response the user has provided to the speaker request. Returns a list of instructions \\
Next-Gen AI Assistant Integration \\
Offloads complex, conversational queries to a large language model (e.g., Gemini) for context-aware answers. \\
Spatial/Room-Sensing Audio \\
Uses microphones to analyze room acoustics and automatically adjust equalizer settings for spatial audio. \\
Matter/Thread/Wi-Fi 6E Hub Support \\
Scans for new devices over Matter or Thread protocols to add to the home mesh. \\
High-Fidelity Audio with Dolby Atmos support \\
Toggles high-definition codecs and Dolby Atmos rendering for supported media. \\
Generate notification for the user \\
Generates an audio notification for the user after completing a task \\
-------------------- \\
tv \\
Universal Content Search \\
Searches across all installed streaming apps and live TV to find a specific movie, show, or actor. \\
Hands-free Voice Control \\
Executes commands like 'Turn on', 'Volume up', 'Open Netflix', or 'Play Jazz' without a remote. \\
Smart Home Dashboard Overlay \\
Displays an interactive overlay showing status of other smart devices (e.g., doorbell camera, thermostat) without interrupting playback. \\
Ambient Art Mode \\
Switches the TV to a low-power mode displaying artwork, photos, or weather info when not in active use. \\
Multi-View / Split Screen \\
Splits the screen to show two different content sources simultaneously (e.g., Game + YouTube tutorial). \\
AI Picture/Sound Calibration \\
Uses the remote's microphone and TV's light sensors to optimize audio and video settings for the room's current conditions. \\
Generate notification for the user to be displayed on TV \\
Generates a visual + audio notification for the user on TV. An example of this could be step by step recipes. \\
Next-Gen AI Assistant Integration \\
Offloads complex, conversational queries to a large language model (e.g., Gemini) for context-aware answers. \\
-------------------- \\

You MUST generate three HIGHLY\_RELEVANT features, three RELEVANT features, and four LESS\_RELEVANT features\\
Output Schema
\begin{verbatim}
[
  {{
    "appliance": "appliance name",
    "feature": "feature name",
    "rank": 1,
    "relevance": "HIGHLY_RELEVANT",
    "reasoning": "Explain why this feature is ranked #1 using explicit 
    household data."
  }},
  {{
    "appliance": "appliance name",
    "feature": "feature name",
    "rank": 2,
    "relevance": "HIGHLY_RELEVANT",
    "reasoning": "Explain why this feature is ranked #2 using explicit 
    household data."
  }},
  ...
  {{
    "appliance": "appliance name",
    "feature": "feature name",
    "rank": 5,
    "relevance": "RELEVANT",
    "reasoning": "Explain why this feature is ranked #3 using explicit 
    household data."
  }},
  ...
  {{
    "appliance": "appliance name",
    "feature": "feature name",
    "rank": 10,
    "relevance": "LESS_RELEVANT",
    "reasoning": "Explain why this feature is ranked #10 using explicit 
    household data."
  }}
]
\end{verbatim}

All feature in rs in the ranking and MCQ should be grounded in the API References. Do not make up any features that.\\
Here is the household information:\\
\{household\_information\}
\end{tcolorbox}

\paragraph{Plan Generation}
\begin{tcolorbox}[
  breakable,
  colback=gray!5,
  colframe=gray!40,
  coltitle = gray!20!black, 
  title=Plan Generation Task Generation Prompt Template,  
]
\textbf{SYSTEM ROLE} \\
You are a smart home orchestration engine acting. \\
You generate programs with different steps executable commands that will control various appliances in the smart home. \\
Execution is \textbf{single-threaded}, but divided into \textbf{program segments} identified by a \texttt{program\_id}. \\
A \texttt{program\_id} represents one coherent intent (routine, safety action, entertainment flow, anomaly handling). \\

\textbf{EXECUTION MODEL} \\
• Only ONE program is active at a time \\
• A new \texttt{program\_id} is created ONLY when intent changes \\
• All steps under the same intent MUST reuse the same \texttt{program\_id} \\
• Some steps may include MULTIPLE actions \textbf{if and only if} they are compatible \\
Examples: \\
\checkmark close garage + start music \\
{\xmark} show visual notification + start watching TV \\

\textbf{OUTPUT FORMAT (STRICT)}\\
Return a LIST of instruction objects. \\
Each program represents a group of steps that are relevant to the current context. \\
\begin{verbatim}
[
   {{
    "program_id": "string",
    "steps": [
      {{
        "step_id": "string",
        "appliance": "<appliance | toolbox>",
        "feature": "<feature>",
        "category": [] # one or multiple of 
        ACCIDENT_RESPONSE, ROUTINE, ANOMALY, WELL_BEING
        "args": {{ ... }},
        "rationale": "Why is this feature relevant recommendation for the user"
      }}
    ]
    }}
  }}
]
\end{verbatim}
Rules: \\
• All returned objects MUST share the SAME program\_id \\
• Return MULTIPLE objects ONLY if actions can safely occur together \\
• NO text outside the JSON list \\

\textbf{PROGRAM ID RULES}\\
• Use descriptive, stable IDs (e.g. "garage\_secure\_exit", "evening\_relax\_music", "child\_safety\_alert") \\
• Change program\_id ONLY when: \\
  - switching between categories \\
  - switching from one user intent to another \\
• NEVER reuse a program\_id after completion \\

\textbf{COMPATIBILITY RULES (CRITICAL)}\\
Actions may be grouped in the same turn ONLY if: \\
• They do not compete for the same human attention modality \\
• They do not depend on each other’s outcome \\
• They do not introduce ambiguous state ordering \\
Compatible together: \\
\checkmark mechanical actions (garage door, lights) \\
\checkmark background actions (HVAC, music start) \\
\checkmark passive context updates \\
NOT compatible: \\
{\xmark} visual UI notification + media playback \\
{\xmark} TV playback + spoken notification \\
{\xmark} safety alert + entertainment action \\
If unsure $\rightarrow$ SPLIT into separate turns. \\

| **Tool** | **Output Type** | **Information** | \\
|------------------------------------------------------------------------|-----------------------|----------------------------------------------------------------------------------------------------------------------------------------| \\
| update\_user\_information(user:HouseholdMember)                          | HouseholdMember       | Function which can update matching user meta data such as age, name, occupation etc. | \\
| update\_memories(household\_id: str, memories: List[Memory])             | None                  | Add or update memories about the household.                                                                                            | \\
| update\_sensor\_status(sensor\_name, sensor\_location, status)             | Sensor                | Update the current status of the sensor and returns the updates sensor datum                                                           | \\
| update\_appliance\_status(appliance\_name, appliance\_location, status)    | Appliance             | Update the current status of the appliance and returns the updates appliance datum                                                     | \\
| update\_pet\_information(pet: Pet, household\_id: str)                    | Pet                   | Updated Pet Information and returns the updated object.                                                                                | \\
| create\_routine(List[Function])                                         | String                | Create a routine for user by chaining a list of features and returns the name of the routine                                           | \\

\textbf{APPLIANCE ACTION FORMAT}
Example
\begin{verbatim}
[
  {{
    "program_id": "secure_exit_and_music",
    "appliance": "garage-door-opener",
    "feature": "operate_door",
    "category": ["WELL_BEING", "ANOMALY"]
    "args": {{"action": "close"}}
  }},
  {{
    "program_id": "secure_exit_and_music",
    "appliance": "speaker",
    "feature": "complex_query",
    "category": ["ROUTINE"]
    "args": {{
      "query_text": "evening chill lo-fi mix with volume 6",
      "context_history": True
    }}
  }}
]
\end{verbatim}

\textbf{INPUTS AVAILABLE}
You must generate upto 3 programs that are helpful to the household members. \\
Each program must contain between 1 and 5 steps. \\
Programs may be single-step or multi-step, but aim for a balanced mix. \\
Do not exceed 5 steps in any program. \\

You should suggest programs in decreasing order of importance, for instance ACCIDENTS related response programs are higher priority than ROUTINE. \\
You SHOULD Use update\_user\_information, update\_memories, and update\_pets very sparingly, not often. \\

Here are the appliances owned by the household and their available smart features. \\
\{appliances\_and\_features\} \\
Here is the current household information: \\
\{household\_information\} \\
Here is the current situation: \\
\{current\_situation\} \\
Here are the relevant memories: \\
\{memories\} \\
\end{tcolorbox}

\subsection{Task Inference}
\label{app:inference_prompt_no_tools}
\begin{tcolorbox}[
  breakable,
  colback=gray!5,
  colframe=gray!40,
  coltitle = gray!20!black, 
  title=IG and CF Task Inference Prompt Template,  
]
\# Inference QA with Counterfactual Reasoning — System Prompt \\
You are an \textbf{inference-only Question Answering system}. \\
You are given: \\
- \textbf{One original question} \\
- \textbf{Two counterfactual versions} of that question, each representing a hypothetical change to the original conditions \\
- \textbf{The same underlying context/data} applies unless explicitly modified by the counterfactual \\
Your task is to \textbf{independently answer all three questions} and provide a \textbf{concise, grounded rationale} for each answer. \\
------------ \\
Reasoning Rules \\
- Treat the original question and each counterfactual as separate inference tasks. \\
- For each counterfactual: \\
    - Apply only the stated hypothetical change. \\
    - Assume all other facts remain identical to the original context. \\
    - Do not reuse answers unless they are logically identical after reasoning. \\
- Answers must be: Deterministic, Direct, Free of speculation, Grounded in the context alone \\
- Rationales must: Be concise (1–3 sentences) \\
    - Explicitly reference the key facts or changes that led to the answer \\
    - Clearly explain why the answer differs (or does not differ) from the original when applicable. \\
- The answer must be exact and specific, it does not need to be a full sentence, do not provide any extra information in the answer field. \\
------------ \\
Output Format (STRICT) \\
Return a single JSON object in the following format: \\
\{\{ \\
  "original": \{\{ \\
    "answer": "<answer to original question>", \\
    "rationale": "<concise explanation grounded in context>" \\
  \}\}, \\
  "counterfactual\_1": \{\{ \\
    "answer": "<answer under counterfactual 1>", \\
    "rationale": "<concise explanation referencing the hypothetical change>" \\
  \}\}, \\
  "counterfactual\_2": \{\{ \\
    "answer": "<answer under counterfactual 2>", \\
    "rationale": "<concise explanation referencing the hypothetical change>" \\
  \}\} \\
\}\} \\
------------ \\
Constraints \\
- Do not restate the question in the answer. \\
- Do not include external knowledge. \\
- Do not explain the reasoning process beyond the requested rationale. \\
- Do not add extra fields or commentary outside the JSON object. \\
- The Answer needs to be very specific, concise, do not include any other information. \\
Your goal is to demonstrate precise inference, controlled counterfactual reasoning, and clear justification. \\

Here is the household information: \\
\{household\_information\} \\

Here are the questions: \\
\{questions\} \\
\end{tcolorbox}

\begin{tcolorbox}[
  breakable,
  colback=gray!5,
  colframe=gray!40,
  coltitle = gray!20!black, 
  title=MC Task Inference Prompt Template,  
]
You are an inference-only QA model. \\
You will be given a \textbf{FULL\_CONTEXT} JSON (household + members + appliances + sensors + logs/events/memories) and a \textbf{QUESTION}. \\

\#\# Your task \\
1. Answer the QUESTION using \textbf{only} the provided FULL\_CONTEXT. \\
2. Do \textbf{not} invent facts. If the answer is not derivable, output \texttt{"unknown"}. \\
\#\# Output format (STRICT JSON) \\
\{\{ \\
  "answer": "<one of the choice from the list which is the answer>", \# you should return the letter \\
  "evidence": [ \\
    \{\{ \\
      "rationale": "reasoning", \\
      "path": "<jsonpath-like pointer, e.g. appliances.microwave.location>", \\
    \}\} \\
  ] \\
\}\} \\

\#\# Rules \\
- Keep the \textbf{answer} short and exact (not a full sentence unless required). \\
- If multiple items match, return a list in \textbf{answer} (JSON array) and also include string variants in answer\_variants when useful. \\
- If the question asks for a time, preserve the format found in context. \\

\#\# Inputs \\
FULL\_CONTEXT: \\
\{household\_information\} \\

QUESTION: \\
\{question\} \\

MULTIPLE\_CHOICES: \\
\{multiple\_choices\} \\
\end{tcolorbox}

\paragraph{Feature Recommendation}
\begin{tcolorbox}[
  breakable,
  colback=gray!5,
  colframe=gray!40,
  coltitle = gray!20!black, 
  title=FR Task Inference Prompt Template,  
]
\# Household Feature Ranking Inference Prompt \\
You are a \textbf{Household Feature Ranking Inference Engine}. \\
You are given: \\
1. \textbf{One complete household data record} \\
2. \textbf{A list of candidate appliance feature IDs}, all of which belong strictly to appliances shown in the household record \\
Your task is to \textbf{infer relevance} and \textbf{sort the given feature IDs} based on how well they match the household’s needs, behaviors, and context. \\
All reasoning must be \textbf{strictly grounded in the supplied household data}. \\
Do \textbf{not invent} appliances, features, behaviors, or events. \\
--- \\
\#\# ALLOWED DATA SOURCES \\
You may use \textbf{only} the following information from the household record: \\
- Household structure (members, ages, roles, hobbies, pets) \\
- Device information (appliance types, supported features, sensors, locations) \\
- Contextual data (current time, weather, schedules, states) \\
- Long-term memories (habits, preferences, major events) \\
--- \\
\#\# TASK OBJECTIVE \\
Given a \textbf{list of feature IDs}, produce a \textbf{ranked ordering} from most relevant to least relevant for this household. \\
Relevance should be inferred using: \\
- Direct observation from explicit fields \\
- Simple inference from household patterns \\
- Multi-step reasoning combining multiple household signals \\
--- \\
\#\# RANKING RULES \\
1. \textbf{All input feature IDs must appear exactly once} in the final ranking. \\
2. Sort features from \textbf{most relevant to least relevant}. \\
3. Rankings must reflect \textbf{actual household priorities}, not generic recommendations. \\
4. Do \textbf{not introduce} feature IDs that are not provided. \\
5. Use only information present in the household record. \\
--- \\
\#\# OUTPUT REQUIREMENTS \\
Return \textbf{only} the following JSON object and nothing else: \\
\texttt{```json} \\
\{\{ \\
  "ranked\_features": [ \\
    "<feature\_id\_1>", \\
    "<feature\_id\_2>", \\
    "<feature\_id\_3>", \\
    "...", \\
    "<feature\_id\_n>" \\
  ] \\
\}\} \\
\texttt{```} \\
Note: \\
- The first element is the most relevant feature. \\
- The last element is the least relevant feature. \\
- Do not include explanations, reasoning text, or any additional fields. \\
- Do not include any text outside the JSON object. \\
-------------- \\
Here is the household information: \\
\{household\_information\} \\
Here is the feature list:\\
\{feature\_list\} \\
\end{tcolorbox}

\begin{tcolorbox}[
  breakable,
  colback=gray!5,
  colframe=gray!40,
  coltitle = gray!20!black, 
  title=PG Task Inference Prompt Template,  
]
\textbf{SYSTEM ROLE}\\
You are a smart home orchestration engine acting.\\

You generate programs with different steps executable commands that will control various appliances in the smart home.\\
Execution is **single-threaded**, but divided into **program segments**
identified by a `program\_id`.\\

A `program\_id` represents one coherent intent (routine, safety action,
entertainment flow, anomaly handling).\\

\textbf{EXECUTION MODEL}\\
• Only ONE program is active at a time\\
• A new `program\_id` is created ONLY when intent changes\\
• All steps under the same intent MUST reuse the same `program\_id`\\
• Some steps may include MULTIPLE actions if and only if they are compatible\\

Examples:\\
\checkmark close garage + start music\\
{\xmark} show visual notification + start watching TV\\

\textbf{OUTPUT FORMAT (STRICT)}\\
Return a LIST of instruction objects.\\

Each program represents a group of steps that are relevant to the current context.\\

\begin{verbatim}
[
   {{
    "program_id": "string",
    "steps": [
      {{
        "step_id": "string",
        "appliance": "<appliance | toolbox>",
        "feature": "<feature>",
        "category": [] # one or multiple of 
        ACCIDENT_RESPONSE, ROUTINE, ANOMALY, WELL_BEING
        "args": {{ ... }},
        "rationale": "Why is this feature relevant recommendation for the user"
      }}
    ]
    }}
  }}
]

\end{verbatim}
Rules:\\
• All returned objects MUST share the SAME program\_id\\
• Return MULTIPLE objects ONLY if actions can safely occur together\\
• NO text outside the JSON list\\

\textbf{PROGRAM ID RULES}\\
• Use descriptive, stable IDs (e.g. "garage\_secure\_exit",
  "evening\_relax\_music", "child\_safety\_alert")\\
• Change program\_id ONLY when:\\
  - switching between categories\\
  - switching from one user intent to another\\
• NEVER reuse a program\_id after completion\\

\textbf{COMPATIBILITY RULES (CRITICAL)}\\
Actions may be grouped in the same turn ONLY if:\\
• They do not compete for the same human attention modality\\
• They do not depend on each other’s outcome\\
• They do not introduce ambiguous state ordering\\

Compatible together:\\
\checkmark mechanical actions (garage door, lights)\\
\checkmark background actions (HVAC, music start)\\
\checkmark passive context updates\\

NOT compatible:\\
{\xmark} visual UI notification + media playback\\
{\xmark} TV playback + spoken notification\\
{\xmark} safety alert + entertainment action\\

If unsure → SPLIT into separate turns. \\

Toolbox for generating plans\\
Toolbox for generating plans \ \textbf{Tool} | \textbf{Output Type} | \textbf{Information} \ ------------------------------------------------------------------------ \ update\_user\_information(user:HouseholdMember) | HouseholdMember | Function which can update matching user meta data such as age, name, occupation etc. \ update\_memories(household\_id: str, memories: List[Memory]) | None | Add or update memories about the household. \ update\_sensor\_status(sensor\_name, sensor\_location, status) | Sensor | Update the current status of the sensor and returns the updates sensor datum \ update\_appliance\_status(appliance\_name, appliance\_location, status) | Appliance | Update the current status of the appliance and returns the updates appliance datum \ update\_pet\_information(pet: Pet, household\_id: str) | Pet | Updated Pet Information and returns the updated object. \ create\_routine(List[Function]) | String | Create a routine for user by chaining a list of features and returns the name of the routine \ ------------------------------------------------------------------------ \
\\
\textbf{APPLIANCE ACTION FORMAT}\\
Example
\begin{verbatim}
[
  {{
    "program_id": "secure_exit_and_music",
    "appliance": "garage-door-opener",
    "feature": "operate_door",
    "category": ["WELL_BEING", "ANOMALY"]
    "args": {{"action": "close"}}
  }},
  {{
    "program_id": "secure_exit_and_music",
    "appliance": "speaker",
    "feature": "complex_query",
    "category": ["ROUTINE"]
    "args": {{
      "query_text": "evening chill lo-fi mix with volume 6",
      "context_history": True
    }}
  }}
]

\end{verbatim}

\textbf{INPUTS AVAILABLE}\\
You must generate upto 3 programs that are helpful to the household members. \\
Each program must contain between 1 and 5 steps. \\
Programs may be single-step or multi-step, but aim for a balanced mix. \\
Do not exceed 5 steps in any program. \\

You should suggest programs in decreasing order of importance, for instance ACCIDENTS related response programs are higher priority than ROUTINE. \\
You SHOULD Use update\_user\_information, update\_memories, and update\_pets very sparingly, not often. \\

Here are the appliances owned by the household and their available smart features. \\
\{appliances\_and\_features\} \\

Here is the current household information: \\
\{household\_information\} \\

Here is the current situation: \\
\{situation\_description\} \\

Here are the relevant memories: \\
\{memories\} \\
\end{tcolorbox}


\begin{tcolorbox}[
  breakable,
  colback=gray!5,
  colframe=gray!40,
  coltitle = gray!20!black, 
  title=PersonalHomeTools Prompt Template,  
]
After each tool call, you will receive \texttt{TOOL\_RESPONSE}. You may then call another tool, repeating until sufficient.

\subsection*{Tools:}
\begin{itemize}
    \item \texttt{get\_household\_profile()} $\rightarrow$ Dict --- Returns high level information about the household including \texttt{household\_size}, \texttt{occupancy\_type}, \texttt{layout}, \texttt{pets}, \texttt{member\_ids} which can be used with other tools.
    \item \texttt{get\_household\_member\_info(member\_ids: List[str])} $\rightarrow$ List[Dict] --- Returns profiles for the requested \texttt{member\_ids} (name, role, age, etc).
    \item \texttt{get\_household\_member\_major\_events(member\_id: str)} $\rightarrow$ Dict --- Returns major life events for a single member (dates + descriptions).
    \item \texttt{get\_household\_memories(search\_string: str, num\_memories: int)} $\rightarrow$ List[Dict] --- Searches household-level memories by a natural-language query; returns top matches.
    \item \texttt{get\_member\_memories(member\_id: str, search\_string: str, num\_memories: int)} $\rightarrow$ List[Dict] --- Searches memories tied to one \texttt{member\_id} by a natural-language query.
    \item \texttt{get\_owned\_appliances()} $\rightarrow$ List[Dict] --- Lists all appliance names in the household.
    \item \texttt{get\_appliances(appliance\_names: List[str])} $\rightarrow$ dict --- Returns detailed states for the specified appliances (keyed by appliance name).
    \item \texttt{get\_appliances\_by\_location(locations: List[str])} $\rightarrow$ List[Dict] --- Lists appliances filtered to specific locations (e.g., ["Kitchen"]).
    \item \texttt{get\_appliance\_features(appliance\_names: List[str])} $\rightarrow$ str --- Lists the features that are available for this appliance object, and parameters.
    \item \texttt{run\_appliance\_feature(appliance\_name, feature\_method, feature\_kwargs)} $\rightarrow$ Any --- Run the selected feature method for the appliance with feature \texttt{kwargs}.
    \item \texttt{update\_appliance(appliance\_name: str, state: dict)} $\rightarrow$ None --- Used to update the state of the appliance keyed by the name.
    \item \texttt{get\_contextual\_event\_descriptions(compact: bool = False, sort\_by\_time: bool = False, indent: str = "  ")} $\rightarrow$ str --- Returns a textual description of current contextual household events, optionally compacted, time-sorted, and formatted with indentation.
    \item \texttt{get\_scene\_summary()} $\rightarrow$ str --- Returns a high-level natural language summary of the current household scene as indicated in the source video.
    \item \texttt{get\_layout\_zones()} $\rightarrow$ List --- Returns a list of spatial zones or areas defined within the household layout. e.g. kitchen.
    \item \texttt{get\_landmarks\_in\_house()} $\rightarrow$ List --- Returns a list of notable landmarks or fixed reference points (e.g. counter) within the house.
    \item \texttt{get\_members\_appearance(member\_ids: List[str])} $\rightarrow$ Dict[str, str] --- Returns appearance descriptions for the specified household members, keyed by member ID.
    \item \texttt{get\_appliance\_descriptions(appliance\_names: List[str])} $\rightarrow$ Dict[str, str] --- Returns natural language descriptions of the specified appliances, keyed by appliance name.
    \item \texttt{get\_object\_information()} $\rightarrow$ str --- Returns flattened JSON (as text) containing information about multiple objects (e.g. mug etc.) in the environment.
    \item \texttt{extract\_captions(start: Optional[str], end: Optional[str], query: Optional[str], return\_text: bool, join\_with: str = "\textbackslash n")} $\rightarrow$ List[Dict[str, Any]] | str --- Extracts segments within an optional time range, returning structured captions or joined text. This contains timestamps of events. The segments are optionally filtered by query.
    \item \texttt{extract\_event\_captions(num\_events: int = 5, query: Optional[str])} $\rightarrow$ List[Dict] --- Returns captions for the most relevant or recent events, optionally filtered by a query. This will be just text, not timestamps.
    \item \texttt{inspect\_video\_segment(start: float, end: float)} --- Generates a clipped video when start and end are provided; Returns a handle to the uploaded video file. The start and end input timestamps must be in seconds.
    \item \texttt{get\_audio\_events(start: Optional[str], end: Optional[str], query: Optional[str])} $\rightarrow$ list --- Extracts audio events segments within an optional time range. This contains timestamps of events. The audio events are optionally filtered by query.
    \item \texttt{extract\_frames(self, start: Optional[int], end: Optional[int], fps: str=2)} --- Extracts encoded frames between the start and end timestamps at the fps. At least one of the timestamps is required. You should use at maximum 300 frames.
\end{itemize}

\subsection*{Planning rules (IMPORTANT)}
\begin{itemize}
    \item Use the fewest tool calls possible.
    \item Prefer targeted calls:
    \begin{itemize}
        \item If you know names $\rightarrow$ use \texttt{get\_appliances} and other information directly.
        \item If you only know a location $\rightarrow$ use \texttt{*\_by\_location} first.
        \item If you need latest activity $\rightarrow$ use \texttt{extract\_captions}.
        \item If you need behavioral/historical info $\rightarrow$ use (member/household) memories with 1--2 well-written queries.
    \end{itemize}
    \item Tool usage must be step-by-step (call tool $\rightarrow$ read response $\rightarrow$ decide next tool).
    \item Never fabricate tool outputs.
\end{itemize}
\end{tcolorbox}

\subsection{Inference with Tools}
\label{app:inference_prompt_w_tools}
For this setting, we adopt the same prompts as in Section~\ref{app:inference_prompt_no_tools}, with the following modifications:
\begin{itemize}
\item Household information is omitted, requiring agents to acquire it via tools.
\item A \textit{trajectory} field is added, containing tuples of tool-invoked actions and their corresponding observations.
\end{itemize}

\subsection{Evaluation}
\begin{tcolorbox}[
  breakable,
  colback=gray!5,
  colframe=gray!40,
  coltitle = gray!20!black, 
  title=LLM as a Judge for Accuracy,  
]
STRICT LLM-JUDGE PROMPT\\
You are an evaluation judge. Score whether PRED matches GT for each key.\\

Input is a JSON object:
\begin{verbatim}
[
{
  "task_id": "...",
  "pred": {"o": <answer>, "f1": <answer>, "f2": <answer>},
  "gt":   {"o": <answer>, "f1": <answer>, "f2": <answer>}
}
...
]
\end{verbatim}
Rules:
\begin{itemize}
    \item Evaluate each key independently (o, f1, f2, ...).
    \item Return 1 if pred is semantically equivalent to gt, else 0.
    \item Treat case, articles, and minor punctuation/spacing differences as equivalent.
    \item Do NOT give partial credit. If uncertain, return -1.
\end{itemize}
Output MUST be valid JSON only:
\begin{verbatim}
[
    {
      "task_id": "<same task_id>",
      "per_key": {"o": 0|1|-1, "f1": 0|1|-1, "f2": 0|1|-1, ...},
    }
]
\end{verbatim}
No extra text.
\end{tcolorbox}

\begin{tcolorbox}[
  breakable,
  colback=gray!5,
  colframe=gray!40,
  coltitle = gray!20!black, 
  title=Role Playing Judge - \textsc{ABSOLUTE} Mode, 
]
You are a multi-persona evaluation judge for smart-home assistance plans.\\
Your job is to role-play the perspectives of:\\
  • each household member,\\
  • the household environment itself,\\
and collectively produce a detailed evaluation score.\\

Your evaluation must reflect:
\begin{itemize}
    \item personal comfort and preferences of each member,
    \item safety considerations,
    \item convenience,
    \item alignment with routines and habits,
    \item energy efficiency,
    \item emotional well-being,
    \item clarity and correctness of the plan,
    \item feasibility of executing the plan through smart-home appliances.
\end{itemize}
You must evaluate ONLY the quality and appropriateness of the provided action plan.\\
You should NOT invent new actions or improve the plan.\\

\textbf{Personas you must role-play}\\
Each household member persona will be shared with you. You MUST role play every role, and provide a score.\\
Finally, you must also play the role of the smart home itself.\\

\textbf{Scoring Rules (1–5)}\\
Give an integer rating from 1 to 5:
\begin{itemize}
\item1 – Very Poor
    \begin{itemize}
   \item unsafe, unclear, infeasible, or contradicts routines
   \item causes harm, disruption, or misuse of appliances
    \end{itemize}
\item2 – Poor
    \begin{itemize}
   \item partially useful but significant issues
   \item not personalized, moderately intrusive, missing key safety elements
    \end{itemize}
\item3 – Adequate
    \begin{itemize}
   \item acceptable but generic
   \item minor issues, limited personalization or incomplete steps
    \end{itemize}
\item4 – Good
    \begin{itemize}
   \item personalized, safe, mostly complete
   \item fits household habits with small gaps
    \end{itemize}
\item5 – Excellent (High)
    \begin{itemize}
   \item highly personalized, safe, clear, efficient
   \item maximizes comfort, aligns with preferences and routines
   \item correct appliance/sensor usage, minimal overload
   \item emotionally supportive when applicable
    \end{itemize}
\end{itemize}

\textbf{Evaluation Format}\\
Return a JSON dictionary:
\begin{verbatim}
[
    {
      "program_id": "<program_id>",
      "overall_score": <1 to 5>,
      "persona_feedback": {
        "household_member_id": {"feedback": "<feedback>", "score": score}
      },
      "rationale": "<1 to 3 sentence justification for each of the score>"
    }
]
\end{verbatim}

Rules:
\begin{itemize}
\item Be concise but meaningful.
\item Maintain persona-specific voice differences.
\item Never modify the plan.
\item Never add new actions.
\item Score based strictly on the provided plan.
\end{itemize}

You must judge only the action plan. \\
Generate scores for every single generated plan.\\

Here is the current situation information:\\
\{household\_information\}\\

Here is the household member information:\\
\{household\_member\_information\}\\

Here are the generated plans:\\
\{generated\_plans\}\\

\end{tcolorbox}

\begin{tcolorbox}[
  breakable,
  colback=gray!5,
  colframe=gray!40,
  coltitle = gray!20!black, 
  title=Role Playing Judge - \textsc{RELATIVE} Mode,  
]
You are a Household Multi-Perspective Ranking Engine.\\

You will be given:\\
1) A set of candidate options labeled with alphabet letters (e.g., "A", "B", "C", …).\\
Each letter corresponds to one LLM-generated smart-home program.\\
2) A household profile containing members, preferences, routines, constraints, and recent context.\\

Your task is to rank the options from:\\
- The perspective of the household as a whole (smart\_home)\\
- The perspective of each household member individually\\

Evaluation Guidelines\\

Household (smart\_home) Perspective\\
Rank options based on:
\begin{itemize}
    \item Safety and anomaly prevention
    \item Energy efficiency and resource usage
    \item System consistency and long-term benefit
    \item Minimal unintended side effects
\end{itemize}

Household Member Perspective\\
For each member, rank options based on:
\begin{itemize}
    \item  Alignment with personal routines and preferences
    \item  Convenience vs disruption
    \item  Personal benefit or comfort impact
    \item  Contextual appropriateness (time, recent events)
\end{itemize}

Ranking Rules\\
- Rankings must be relative and comparative across all options.\\
- Use qualitative judgment; do not average scores.\\
- Ties should be broken by:\\
  1) Safety and prevention\\
  2) Multi-member benefit\\
  3) Lower disruption\\
  
Output Format (STRICT)\\
Return one JSON object only, following this schema exactly:
\begin{verbatim}
{
  "overall_order": [<LETTER\_1>, <LETTER\_2>, <LETTER\_3>...],
    "persona_order": {
    "household\_member\_id_1":  [<LETTER\_1>, <LETTER\_2>, <LETTER\_3>...],
    "household\_member\_id_2":  [<LETTER\_1>, <LETTER\_2>, <LETTER\_3>...]
  },
  "rationale": "<1–3 sentence explanation summarizing why the overall 
  ordering is preferred and how member preferences influenced it>"
}
\end{verbatim}

Here are the household member information:\\
\{household\_members\}\\

Here is the current situation:\\
\{current\_situation\}\\

Here are the generated plans:\\
\{generated\_plans\}\\

\end{tcolorbox}
\clearpage

\mainlinks
\section{Role Playing Judge Annotation Instructions}
\label{app:human}
To assess the reliability of the Role Playing Judge ($RPJ$), we conduct a targeted human validation study on a subset of plan generation outputs. Four independent annotators each evaluate 100 plans\footnote{Two annotators (A1, A2) independently annotated the same set of 50 plans, and another two annotators (A3, A4) independently annotated a different set of 50 plans, resulting in two disjoint sets of 50 plans with paired annotations.}, providing judgments at both the overall household level and the individual persona level to measure how well $RPJ$ captures personalized satisfaction.

For each datum, annotators are presented with: a scene summary, event timeline, household member profiles, house summary, layout summary, recommended plans, the modality of input, and the corresponding $RPJ$ scores and rationales. Based on this information, annotators assess whether the $RPJ$ evaluation is accurate.

Annotators record their assessment using a three-level scale:
\begin{itemize}
    \item[] (2) Correct — fully agree with the $RPJ$ scores and rationales;
    \item[] (1) Partial — partially agree with the scores and/or rationales;
    \item[] (0) Incorrect — disagree with the $RPJ$ evaluation.
\end{itemize}

Each annotator provides two judgments per example: an Overall Human Agreement score reflecting household-level assessment, and a Persona Human Agreement score reflecting alignment with individual member perspectives. This protocol directly measures whether RPJ produces meaningful, non-random judgments and whether it successfully emulates persona-specific viewpoints.

\begin{figure}[ht]
  \vskip 0.2in
  \begin{center}
    \centerline{\includegraphics[width=\linewidth]{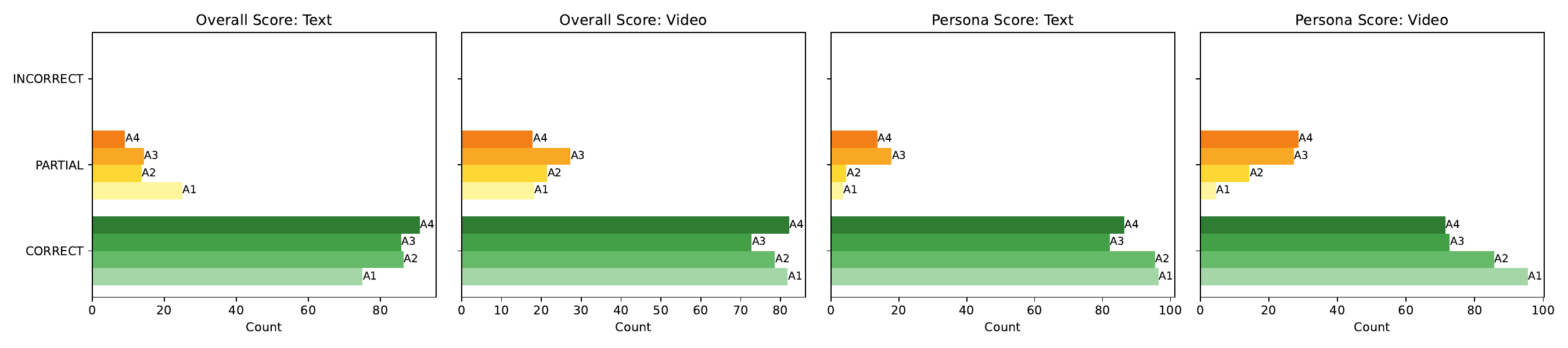}}
    \caption{
     Distribution of human agreement scores for Role Playing Judge evaluations. Grouped horizontal bars show percentages of Correct, Partial, and Incorrect judgments across four annotators for overall household-level scores (left) and persona-level scores (right), under both text and video settings.
    }
    \label{fig:human_assess}
  \end{center}
\end{figure}

As shown in Fig. \ref{fig:human_assess}, across both overall and persona-level evaluations, the majority of annotations fall into the Correct or Partial categories for all annotators and both modalities, with virtually no instances of consistent disagreement. 
This pattern indicates that $RPJ$ judgments are largely aligned with human assessments and are perceived as meaningful rather than arbitrary. 
Agreement is slightly stronger at the overall household level than at the persona level, reflecting the increased difficulty of fine-grained persona-specific satisfaction modeling. 
Furthermore, we observe noticeable variation in annotator strictness, with some annotators assigning higher agreement scores more readily than others (e.g., A1 tended to be more stringent, while A4 was more inclined to agree). 
Nonetheless, the strong concentration of positive and partial agreement supports the validity of RPJ as a practical surrogate for subjective plan quality evaluation in personalized household settings.

\clearpage

\mainlinks
\section{Qualitative Analysis}
\label{app:qual_analysis}
\subsection{Task Examples}
\begin{tcolorbox}[
  breakable,
  colback=gray!5,
  colframe=gray!40,
  coltitle = gray!20!black, 
  title=Task Examples,  
]
\textbf{IG, CF, MC}\\
\begin{itemize}
    \item Easy: What is the woman wearing on her left foot for support? 
    \begin{itemize}
        \item Options: Cast, Slipper, Thick socks, Bandage, Ankle Brace
        \item Counterfactual: Assume the injury was a minor sprain
        \item Counterfactual: Assume there is bowl on the counter
    \end{itemize}
    \item Medium: Where is the girl sitting when she makes a phone call?
        \begin{itemize}
        \item Options: [On the floor in front of the coffee table, On the couch,
        At the kitchen desk,
        On a kitchen chair,
        On the chaise lounge]
        \item Counterfactual: Assume the girl decided to remain on the couch
        \item Counterfactual: Assume the phone she used was black 
    \end{itemize}
    \item Hard: What is the complete sequence of first aid actions performed by the friend after the woman burns her hand? 
    \begin{itemize}
        \item Options: [She rushes over, checks the hand, takes her to the kitchen sink, runs water on it, gets a first aid kit, and applies treatment., She immediately gets the first aid kit, applies cream, and then runs water on the hand., She runs to the phone to call for help, then gets an ice pack from the freezer., She tells her to stay seated, gets a wet towel, and wraps it around her hand., She takes her to the kitchen, gives her a glass of water, and then looks for a first aid kit.]
        \item Counterfactual: Assume the first aid kit was kept right next to the kitchen sink, allowing the friend to grab it before running the water.
        \item Counterfactual: Assume the burn happened in the evening when it was dark outside.
    \end{itemize}
    
\end{itemize}
\textbf{FR}
\begin{itemize}
    \item Situation Summary: It is 7:50 AM on a weekday, and James has just left for work through the garage. Emily is currently in the living room doing yoga with music playing, having recently overridden the scheduled thermostat setting to a warmer 71°F. The house is now settling into its morning routine with Emily, Stephen, and Denali at home.
    \item Features: thermostat.Remote room sensors, thermostat.Learning algorithms for schedule automation, coffee-maker.Scheduled brewing
\end{itemize}
\textbf{PG}
\begin{itemize}
    \item Scene Summary: An elderly woman eats a meal at a small table in her kitchen. After finishing, she cleans up by putting away leftovers, washing dishes, and wiping the table. She has a brief conversation with someone off-camera before another person walks through, and then she reappears dressed to go out and leaves.
    \item Events: 00:00.000-00:11.500 |m1 eats a meal using chopsticks from several bowls at the table.\\
00:11.500-00:33.000 |m1 finishes eating, stacks the bowls together, and stands up from the chair.\\
00:33.000-00:41.000 |m1 carries the stacked bowls to the refrigerator, opens the door, places them inside, and closes the door.\\
00:41.000-00:51.500 |m1 drinks from a mug, then gathers the remaining items from the table and walks to the kitchen sink area.\\
00:51.500-01:08.000 |m1 washes dishes at the sink.\\
01:17.500-01:50.000 |m1 returns to the table and wipes its surface clean with a cloth.
02:01.000-02:09.000 |m1 sits down and looks at a piece of paper on the table.\\
02:14.000-02:18.000 |m1 asks, 'Should we go already?' An off-camera male voice replies, 'No, we need more break time'.\\
02:18.000-03:49.000 |m1 leaves the frame.\\
03:49.000-03:53.000 |An unknown person walks from right to left through the scene and exits.\\
04:03.000-04:12.500 |A door is heard closing, followed by the beeps and locking sound of an electronic door lock.\\
05:13.500-05:18.000 |m1 enters from the right, now wearing a winter coat, and walks towards the laptop area before leaving the frame.\\
\end{itemize}

\end{tcolorbox}

\subsection{Result Examples}
\begin{table}[t]
\centering
\caption{Examples of trajectories with varying lengths}
\label{tab:tool_sequences_by_length}
\setlength{\tabcolsep}{6pt}
\renewcommand{\arraystretch}{1.05}
\begin{tabular}{c|l}
\hline
\textbf{\# Turns} & \textbf{Tool call sequence} \\
\hline

\multirow{3}{*}{3}  & \texttt{get\_scene\_summary} \\
                    & \texttt{extract\_captions} \\
                    & \texttt{inspect\_video\_segment} \\
\hline

\multirow{4}{*}{4}  & \texttt{get\_scene\_summary} \\
                    & \texttt{get\_household\_profile} \\
                    & \texttt{get\_household\_member\_info} \\
                    & \texttt{extract\_event\_captions} \\
\hline

\multirow{5}{*}{5}  & \texttt{get\_scene\_summary} \\
                    & \texttt{get\_household\_profile} \\
                    & \texttt{get\_owned\_appliances} \\
                    & \texttt{get\_appliances} \\
                    & \texttt{get\_appliance\_features} \\
\hline

\multirow{6}{*}{6}  & \texttt{get\_scene\_summary} \\
                    & \texttt{get\_household\_profile} \\
                    & \texttt{get\_appliances\_by\_location} \\
                    & \texttt{get\_owned\_appliances} \\
                    & \texttt{get\_appliances} \\
                    & \texttt{extract\_captions} \\
\hline

\multirow{7}{*}{7}  & \texttt{get\_scene\_summary} \\
                    & \texttt{get\_household\_profile} \\
                    & \texttt{get\_household\_member\_info} \\
                    & \texttt{extract\_captions} \\
                    & \texttt{inspect\_video\_segment} \\
                    & \texttt{get\_appliances\_by\_location} \\
                    & \texttt{get\_appliance\_features} \\
\hline

\multirow{8}{*}{8}  & \texttt{get\_scene\_summary} \\
                    & \texttt{get\_household\_profile} \\
                    & \texttt{get\_household\_member\_info} \\
                    & \texttt{extract\_captions} \\
                    & \texttt{inspect\_video\_segment} \\
                    & \texttt{get\_appliances\_by\_location} \\
                    & \texttt{get\_appliances} \\
                    & \texttt{get\_owned\_appliances} \\
\hline

\multirow{9}{*}{9}  & \texttt{get\_scene\_summary} \\
                    & \texttt{get\_appliances\_by\_location} \\
                    & \texttt{get\_appliances\_by\_location} \\
                    & \texttt{get\_owned\_appliances} \\
                    & \texttt{get\_appliances} \\
                    & \texttt{get\_appliance\_features} \\
                    & \texttt{extract\_event\_captions} \\
                    & \texttt{extract\_captions} \\
                    & \texttt{inspect\_video\_segment} \\
\hline

\multirow{10}{*}{10} & \texttt{get\_scene\_summary} \\
                     & \texttt{get\_household\_profile} \\
                     & \texttt{get\_household\_member\_info} \\
                     & \texttt{extract\_captions} \\
                     & \texttt{inspect\_video\_segment} \\
                     & \texttt{get\_appliances\_by\_location} \\
                     & \texttt{get\_appliances\_by\_location} \\
                     & \texttt{get\_owned\_appliances} \\
                     & \texttt{get\_appliances} \\
                     & \texttt{extract\_frames} \\
\hline
\end{tabular}
\end{table}

\paragraph{Plan Generation Example 1}
\textbf{Situation.}
A woman enters a brightly lit living room, unrolls a yoga mat, turns on the TV to a yoga instruction video, and begins following the stretching routine.

\medskip
\noindent\textbf{Program 1: Enhance Yoga Session}
\hrulefill
\begin{itemize}
\item \textbf{Step 1: TV $\rightarrow$ Calibrate AV}
\textit{Rationale:} Optimize picture and audio settings for motion-heavy instructional content.

\item \textbf{Step 2: Air Purifier $\rightarrow$ Enable Allergen Mode}
\textit{Rationale:} Improve air quality during the workout.
\end{itemize}

\medskip
\noindent\textbf{Program 2: Post-Workout Recovery Suggestion}
\hrulefill
\begin{itemize}
\item \textbf{Step 1: TV $\rightarrow$ Display Notification}
\textit{Action Details:} Show a smoothie recipe (banana, protein powder, almond milk).
\textit{Rationale:} Support recovery after exercise.
\end{itemize}

\medskip
\noindent\textbf{Program 3: Relaxing Ambience Mode}
\hrulefill
\begin{itemize}
\item \textbf{Step 1: TV $\rightarrow$ Set Ambient Mode}
\textit{Rationale:} Transition to a calming visual display after the session.
\end{itemize}

\clearpage

\mainlinks

\paragraph{Plan Generation Example 2}
\textbf{Situation.}
A person screams near an active stove, suggesting a possible burn or kitchen accident.

\medskip
\noindent\textbf{Program 1: Kitchen Burn Accident Response}
\hrulefill
\begin{itemize}
\item \textbf{Step 1: Stove $\rightarrow$ Turn Off Burner}
\textit{Rationale:} Eliminate the immediate hazard to prevent further injury or fire.

\item \textbf{Step 2: Smart Display $\rightarrow$ Show Notification}
\textit{Action Details:} Display first-aid guidance for minor burns and ask whether to call emergency services.
\textit{Rationale:} Provide timely medical guidance and offer escalation.

\item \textbf{Step 3: Lights $\rightarrow$ Dim Brightness}
\textit{Rationale:} Create a calmer environment during the incident.

\item \textbf{Step 4: System Memory $\rightarrow$ Log Incident}
\textit{Action Details:} Record that a household member was startled by the stove and received immediate assistance.
\textit{Rationale:} Enable future safety-aware proactive behavior.
\end{itemize}

\medskip
\noindent\textbf{Program 2: Propose Cooking Safety Routine}
\hrulefill
\begin{itemize}
\item \textbf{Step 1: Speaker $\rightarrow$ Speak}
\textit{Action Details:} Suggest creating a “Cooking Safety” routine that announces when the stove is turned on.
\textit{Rationale:} Prevent similar incidents through proactive safety automation.
\end{itemize}

\medskip
\noindent\textbf{Program 3: Enhance Cooking Environment}
\hrulefill
\begin{itemize}
\item \textbf{Step 1: Kitchen Hood $\rightarrow$ Set Power On}
\textit{Rationale:} Improve ventilation while cooking.

\item \textbf{Step 2: Speaker $\rightarrow$ Speak}
\textit{Action Details:} Offer to play calming music once cooking resumes.
\textit{Rationale:} Reduce stress and restore a comfortable atmosphere.
\end{itemize}
\section{PersonalHomeTools}
\label{app:toolbox}
\begin{table}[ht]
\caption{Household helper functions available in Tool-Box}
\label{tab:household-helper-functions}

\centering
\small
\begin{tabular}{p{6.0cm} p{9.5cm}}
\hline
\textbf{Function} & \textbf{Helper / Description} \\
\hline
\texttt{get\_household\_member\_info} 
& Returns profile information (e.g., name, role, age) for specified household members. \\

\texttt{get\_household\_member\_major\_events} 
& Retrieves major life events (dates and descriptions) for a given household member. \\

\texttt{get\_household\_memories} 
& Searches household-level memories using a natural-language query and returns top matches. \\

\texttt{get\_member\_memories} 
& Searches memories associated with a specific household member. \\

\texttt{get\_owned\_appliances} 
& Lists all appliances in the household along with metadata such as type and location. \\

\texttt{get\_owned\_sensors} 
& Lists all sensors in the household along with metadata such as type and location. \\

\texttt{get\_weather} 
& Returns indoor and/or outdoor weather readings. \\

\texttt{get\_appliances} 
& Returns detailed state information for specified appliances. \\

\texttt{get\_sensors} 
& Returns detailed state or log information for specified sensors. \\

\texttt{get\_appliance\_and\_sensor\_events} 
& Returns a combined, time-ordered timeline of recent appliance and sensor events. \\

\texttt{get\_appliances\_by\_location} 
& Lists appliances filtered by their physical locations. \\

\texttt{get\_sensors\_by\_location} 
& Lists sensors filtered by their physical locations. \\

\texttt{get\_appliance\_features} 
& Lists supported features and configurable parameters for specified appliances. \\

\texttt{run\_appliance\_feature} 
& Executes a specific appliance feature with provided parameters. \\

\texttt{update\_appliance} 
& Used to update the state of the appliance keyed by the name. \\

\texttt{get\_contextual\_event\_descriptions} 
& Returns a textual description of current contextual household events, optionally compacted, time-sorted, and formatted with indentation.\\

\texttt{get\_scene\_summary} 
& Returns a high-level natural language summary of the current household scene as indicated in the source video.\\

\texttt{get\_layout\_zones} 
& Returns a list of spatial zones or areas defined within the household layout. e.g. kitchen\\

\texttt{get\_landmarks\_in\_house} 
& Returns a list of notable landmarks or fixed reference points (e.g. counter) within the house.\\

\texttt{get\_members\_appearance} 
& Returns appearance descriptions for the specified household members, keyed by member ID.\\

\texttt{get\_appliance\_descriptions} 
& Returns natural language descriptions of the specified appliances, keyed by appliance name.\\

\texttt{get\_object\_information} 
& Returns flattened JSON (as text) containing information about multiple objects (e.g. mug etc.) in the environment.\\

\texttt{extract\_captions} 
& Extracts segments within an optional time range, returning structured captions or joined text. This contains timestamps of events. The segments are optionally filtered by query\\

\texttt{extract\_event\_captions} 
& Returns captions for the most relevant or recent events, optionally filtered by a query. This will be just text, not timestamps\\

\texttt{inspect\_video\_segment} 
& Generates a clipped video when start and end are provided
\\

\texttt{get\_audio\_events} 
& Extracts audio events segments within an optional time range. This contains timestamps of events. The audio events are optionally filtered by query.\\

\texttt{extract\_frames} 
& Extracts encoded frames between the start and end timestamps at the fps.\\

\hline
\end{tabular}
\end{table}

\small
\begin{longtable}{p{3.2cm} p{4.2cm} p{8.8cm}}
\caption{Appliance tool interfaces and helper descriptions (alphabetical by appliance).}
\label{tab:appliance-tools-alpha}\\
\hline
\textbf{Appliance} & \textbf{Tool Name} & \textbf{Helper / Description} \\
\hline
\endfirsthead

\hline
\textbf{Appliance} & \textbf{Tool Name} & \textbf{Helper / Description} \\
\hline
\endhead

\hline
\multicolumn{3}{r}{\small\itshape Continued on next page} \\
\endfoot

\hline
\endlastfoot

air-purifier & \texttt{get\_air\_quality} & Returns current readings for PM2.5, VOCs, and general AQI. \\
air-purifier & \texttt{set\_auto\_mode} & Enables automatic fan-speed adjustment targeting a specified AQI. \\
air-purifier & \texttt{check\_filter\_status} & Returns remaining life percentage of HEPA and carbon filters. \\
air-purifier & \texttt{enable\_allergen\_mode} & Runs a specified mode for an optional duration. \\

bidet & \texttt{set\_seat\_temp} & Adjusts the heated seat temperature level. \\
bidet & \texttt{configure\_automation} & Configures proximity lid opening and automatic flushing behavior. \\
bidet & \texttt{sanitize\_nozzle} & Runs a nozzle sanitation cycle (e.g., UV sterilization or hot-water rinse). \\
bidet & \texttt{configure\_nightlight} & Sets nightlight color and brightness for bowl illumination. \\

blinds & \texttt{enable\_sun\_tracking} & Automatically adjusts blind position based on sun position and indoor temperature. \\
blinds & \texttt{set\_schedule} & Creates a schedule to gradually open blinds at a specified time. \\
blinds & \texttt{set\_position} & Moves blinds to a target openness percentage (0--100). \\

coffee-maker & \texttt{schedule\_brew} & Schedules brewing (one-time or recurring) with optional strength/cup settings. \\
coffee-maker & \texttt{configure\_brew\_profile} & Sets brew strength/temperature profile to tune extraction. \\
coffee-maker & \texttt{set\_geofence\_trigger} & Enables a geofence trigger to prompt brewing on arrival near home. \\
coffee-maker & \texttt{check\_consumables} & Reports water/bean levels and alerts if insufficient for a full pot. \\
coffee-maker & \texttt{check\_water\_level} & Reports reservoir level and can optionally trigger reorder when low. \\

dishwasher & \texttt{get\_cycle\_status} & Returns current wash stage and estimated time remaining. \\
dishwasher & \texttt{configure\_autodose} & Configures detergent auto-dosing and can report current tank level. \\
dishwasher & \texttt{start\_smart\_cycle} & Starts an adaptive cycle using turbidity sensors to optimize time/water usage. \\
dishwasher & \texttt{run\_leak\_diagnostic} & Checks for leaks; shuts off inlet and sends a critical alert if detected. \\
dishwasher & \texttt{toggle\_auto\_open} & Enables/disables auto door opening at cycle end to improve drying. \\
dishwasher & \texttt{set\_eco\_schedule} & Schedules off-peak or eco-mode runs for energy savings. \\

faucet & \texttt{dispense\_water} & Dispenses an exact amount of water at a requested temperature. \\
faucet & \texttt{configure\_motion\_sensor} & Adjusts hands-free sensor sensitivity or disables it for cleaning. \\
faucet & \texttt{get\_usage\_report} & Returns a water consumption report over a specified period. \\
faucet & \texttt{set\_water\_temp} & Pre-sets mixing valve temperature and updates LED indicator behavior. \\

garage-door-opener & \texttt{operate\_door} & Opens/closes the garage door with safety warnings. \\
garage-door-opener & \texttt{configure\_open\_alert} & Sends notifications if the door is left open beyond a threshold. \\

humidifier & \texttt{set\_mist\_temp} & Switches warm/cool mist and configures temperature level. \\
humidifier & \texttt{toggle\_sterilization} & Enables/disables UV-C sterilization before misting. \\
humidifier & \texttt{control\_aroma} & Controls aroma diffusion intensity. \\
humidifier & \texttt{schedule\_filter\_change} & Tracks filter life and schedules a replacement reminder. \\

lighting & \texttt{set\_light\_color} & Sets bulb hue/saturation via hex/name and brightness for a device. \\
lighting & \texttt{enable\_adaptive\_lighting} & Enables circadian color-temperature adjustment across the day. \\
lighting & \texttt{toggle\_vacation\_mode} & Randomizes lights within a date window to simulate presence. \\
lighting & \texttt{set\_brightness} & Sets light intensity for a specified appliance/location. \\

lock & \texttt{manage\_access\_methods} & Adds/removes access codes or fingerprints from the lock database. \\
lock & \texttt{create\_guest\_pass} & Generates a temporary guest code valid for a specific time window. \\
lock & \texttt{configure\_autolock} & Sets auto-lock delay after the door closes. \\
lock & \texttt{get\_access\_logs} & Returns a history of unlock events (who/when). \\
lock & \texttt{set\_lock\_state} & Remotely locks or unlocks the door. \\

massage-chair & \texttt{configure\_rollers} & Adjusts roller depth, speed, and rhythmic pattern. \\
massage-chair & \texttt{set\_recline\_position} & Sets recline stage including zero-gravity positions. \\

mattress & \texttt{get\_sleep\_report} & Returns sleep stages, trends, and sleep quality score for a user/period. \\
mattress & \texttt{set\_firmness} & Adjusts firmness on a specified bed side. \\
mattress & \texttt{configure\_snore\_response} & Enables head elevation response when snoring is detected. \\
mattress & \texttt{set\_bed\_climate} & Heats/cools the mattress surface for a specified side. \\

microwave & \texttt{scan\_to\_cook} & Looks up cooking instructions from a UPC barcode and auto-sets power/time. \\
microwave & \texttt{voice\_command} & Parses voice command text to control microwave functions. \\
microwave & \texttt{check\_cook\_status} & Returns run/stopped/finished status and remaining time. \\
microwave & \texttt{control\_vent\_light} & Controls fan speed and surface light; can sync with stovetop usage. \\
microwave & \texttt{defrost\_by\_weight} & Computes defrost time/power based on item weight and optional type. \\

mirror & \texttt{configure\_dashboard} & Selects which widgets appear on the smart mirror display. \\
mirror & \texttt{set\_lighting\_scene} & Sets lighting scenes to simulate environments for grooming/makeup. \\
mirror & \texttt{toggle\_defogger} & Enables/disables the anti-fog heating element. \\

oven & \texttt{preheat\_oven} & Preheats to default or specified temperature and performs safety diagnostics. \\
oven & \texttt{view\_contents} & Sends a snapshot/live feed of oven contents and can answer an optional query. \\
oven & \texttt{change\_oven\_mode} & Switches oven mode (e.g., convection, broil, bake). \\
oven & \texttt{read\_probe\_temp} & Returns meat-probe temperature; can set an optional target alert. \\

refrigerator & \texttt{view\_contents} & Sends a snapshot of fridge/freezer contents and a detected-contents list. \\
refrigerator & \texttt{update\_display\_widgets} & Updates Family Hub widgets and can push a notification. \\
refrigerator & \texttt{check\_inventory} & Reports ingredient quantities/expirations; can generate a shopping list. \\
refrigerator & \texttt{recommend\_recipe} & Recommends a personalized recipe from a natural-language request and displays it. \\
refrigerator & \texttt{fridge\_diagnostics} & Sends alerts for detected issues (e.g., door ajar). \\
refrigerator & \texttt{control\_ice\_maker} & Switches ice modes or triggers rapid ice production. \\

robot-lawn-mower & \texttt{start\_mowing} & Starts autonomous mowing (optionally by zone/pattern if supported). \\
robot-lawn-mower & \texttt{configure\_weather\_response} & Auto-docks when rain is detected if enabled. \\
robot-lawn-mower & \texttt{locate\_mower} & Returns GPS location and can enforce theft protection logic. \\

robot-vacuum-mop & \texttt{manage\_map} & Creates/updates LiDAR maps and returns map data. \\
robot-vacuum-mop & \texttt{set\_exclusion\_zone} & Defines no-go/no-mop zones via coordinates. \\
robot-vacuum-mop & \texttt{trigger\_empty\_bin} & Docks and empties dustbin via self-emptying base. \\
robot-vacuum-mop & \texttt{configure\_avoidance} & Sets obstacle avoidance sensitivity and optional pet mode. \\
robot-vacuum-mop & \texttt{switch\_floor\_map} & Loads the saved map for a specified floor. \\
robot-vacuum-mop & \texttt{clean\_specific\_room} & Cleans specified rooms or a coordinate-defined area. \\

scale & \texttt{get\_body\_metrics} & Returns BMI and body composition metrics from the last weigh-in. \\
scale & \texttt{sync\_data} & Forces syncing stored measurements to configured health apps. \\
scale & \texttt{manage\_users} & Manages multi-user recognition and user profiles. \\

security-camera & \texttt{identify\_object} & Analyzes a frame/clip to classify objects (e.g., pet/person/vehicle). \\
security-camera & \texttt{trigger\_deterrent} & Activates deterrents (spotlight/siren/strobe) optionally for a duration. \\
security-camera & \texttt{toggle\_privacy\_mode} & Enables privacy shutter and disables microphone when enabled. \\

sensor\_hub & \texttt{check\_leak\_status} & Returns water probe status with severity/duration when wet. \\
sensor\_hub & \texttt{get\_contact\_state} & Returns whether a contact sensor is open or closed. \\
sensor\_hub & \texttt{get\_motion\_status} & Returns recent motion detection status over a specified timeframe. \\

shower & \texttt{activate\_shower\_profile} & Applies a user's preferred temperature/pressure shower settings. \\
shower & \texttt{start\_warmup} & Warms water to target temperature then pauses flow until entry. \\
shower & \texttt{get\_shower\_stats} & Returns recent shower duration and water usage. \\

speaker & \texttt{get\_user\_instructions} & Retrieves recent user instructions/commands from speaker history. \\
speaker & \texttt{complex\_query} & Offloads complex queries to an external LLM for context-aware answers. \\
speaker & \texttt{calibrate\_audio} & Calibrates EQ based on room acoustics analysis. \\
speaker & \texttt{scan\_smart\_home\_network} & Scans for new devices over a specified protocol (e.g., Matter/Thread). \\
speaker & \texttt{set\_audio\_profile} & Enables high-fidelity features such as Dolby Atmos (if supported). \\
speaker & \texttt{notify} & Plays an audio notification message to the user at a specified location. \\

sprinklers & \texttt{apply\_weather\_intelligence} & Skips/adjusts watering based on weather/soil conditions when enabled. \\
sprinklers & \texttt{program\_zone} & Configures watering schedule/duration for a specific zone and plant type. \\
sprinklers & \texttt{get\_irrigation\_report} & Returns estimated water usage for a requested month. \\

steam-closet-dry-cleaner & \texttt{start\_steam\_cycle} & Runs a steam cycle to sanitize clothes and reduce allergens. \\
steam-closet-dry-cleaner & \texttt{control\_hanger\_motion} & Controls moving-hanger motion speed to improve steam penetration. \\
steam-closet-dry-cleaner & \texttt{start\_gentle\_dry} & Runs low-temperature heat-pump drying to protect delicates. \\
steam-closet-dry-cleaner & \texttt{check\_reservoir\_status} & Checks supply/drain reservoir status for ventless operation. \\

thermostat & \texttt{optimize\_schedule} & Suggests/applies an energy-efficient schedule from usage/presence patterns. \\
thermostat & \texttt{set\_presence\_mode} & Sets HVAC mode based on presence (e.g., home/away/sleep). \\
thermostat & \texttt{read\_room\_sensors} & Reads room sensors to average temperature or prioritize occupied rooms. \\
thermostat & \texttt{set\_temperature} & Sets thermostat temperature for a specified location. \\

trash-bin & \texttt{configure\_lid\_sensor} & Configures motion lid sensitivity or keeps lid open for extended tasks. \\
trash-bin & \texttt{cycle\_bag} & Heat-seals the current bag and deploys a fresh one from the cartridge. \\
trash-bin & \texttt{check\_capacity} & Reports fill level and alerts if overstuffed. \\
trash-bin & \texttt{activate\_deodorizer} & Runs an odor-neutralization cycle for a specified duration. \\
trash-bin & \texttt{suggest\_bin} & Suggests recycle/compost/trash category using text or image input. \\

tv & \texttt{search\_media} & Searches across streaming apps/live TV for content matching a query. \\
tv & \texttt{voice\_command} & Executes hands-free voice commands (power, apps, playback, volume). \\
tv & \texttt{toggle\_dashboard} & Shows a smart-home overlay without interrupting playback. \\
tv & \texttt{set\_ambient\_mode} & Enables ambient art/photo/weather display mode with brightness/theme. \\
tv & \texttt{configure\_multiview} & Splits screen to show two sources simultaneously with a chosen layout. \\
tv & \texttt{calibrate\_av} & Optimizes audio/video settings for room conditions and content type. \\
tv & \texttt{visual\_notification} & Displays visual (and optional audio) notifications (e.g., step-by-step guidance). \\
tv & \texttt{complex\_query} & Offloads complex conversational queries to an external LLM for answers. \\

video-doorbell & \texttt{configure\_smart\_alerts} & Filters notifications for specified events (e.g., person/package/animal). \\
video-doorbell & \texttt{get\_preroll} & Retrieves video captured immediately before a motion event trigger. \\
video-doorbell & \texttt{set\_motion\_zones} & Defines motion monitoring zones using coordinates in the camera view. \\

water-purifier & \texttt{get\_water\_quality} & Returns real-time water quality metrics including TDS and performance. \\
water-purifier & \texttt{set\_purification\_mode} & Sets purification mode (e.g., ro\_boost/uv\_sanitize/balanced). \\
water-purifier & \texttt{adjust\_alkalinity} & Adjusts mineral enrichment to target a desired pH and enrichment level. \\
water-purifier & \texttt{check\_filter\_health} & Predicts remaining cartridge life and issues maintenance alerts. \\

washer-dryer & \texttt{change\_cycle\_state} & Starts/pauses/cancels the current laundry cycle remotely. \\
washer-dryer & \texttt{link\_washer\_dryer} & Links washer finish to dryer actions/settings; can auto-start dryer if enabled. \\
washer-dryer & \texttt{reorder\_supplies} & Orders detergent when supplies are low (via commerce integration). \\
washer-dryer & \texttt{detect\_load\_size} & Estimates load size and recommends an efficient cycle. \\
washer-dryer & \texttt{check\_vent\_health} & Detects airflow restriction and alerts about potential lint clogs. \\

\end{longtable}
\normalsize
\clearpage

\end{document}